# A guided journey through non-interactive automatic story generation

**Luís Miguel Botelho**[1,*]

[1] Instituto de Telecomunicações (IT-IUL) and Department of Information Science and Technology of Instituto Universitário de Lisboa (Iscte-IUL); Luis.Botelho@iscte-iul.pt

**\*** Correspondence: Luis.Botelho@iscte-iul.pt; Tel.: +351-963-405-048

## Abstract

We present a literature survey on non-interactive computational story generation. The article starts with the presentation of requirements for creative systems, three types of models of creativity (computational, socio-cultural, and individual), and models of human creative writing. Then it reviews each class of story generation approach depending on the used technology: story-schemas, analogy, rules, planning, evolutionary algorithms, implicit knowledge learning, and explicit knowledge learning. Before the concluding section, the article analyses the contributions of the reviewed work to improve the quality of the generated stories. This analysis addresses the description of the story characters, the use of narrative knowledge including about character believability, and the possible lack of more comprehensive or more detailed knowledge or creativity models. Finally, the article presents concluding remarks in the form of suggestions of research topics that might have a significant impact on the advancement of the state of the art on autonomous non-interactive story generation systems. The article concludes that the autonomous generation and adoption of the main idea to be conveyed and the autonomous design of the creativity ensuring criteria are possibly two of most important topics for future research.

## 1 Introduction

Before actually starting, we would like to address the reader and other researchers. In spite the fact that this is a survey, we have not reviewed the whole literature in the subject area. We apologize to all researchers whose work we have not reviewed in this overview, and we suggest the reader to search additional work beyond the referred one. Besides, the descriptions and comments we present about the reviewed literature result of our personal interpretations of the considered work; they are neither comprehensive nor objective. We strongly encourage the reader to consult the original sources.

The present overview is exclusively about the automatic generation of non-interactive stories. It does not address interactive storytelling per se. However, it may present methods or algorithms originally put forth, by their authors, in the context of interactive storytelling, if those methods or algorithms are relevant in non-interactive contexts.

This article does not evaluate or compare the results achieved by reviewed proposals mainly because not all of them present the generated results and because each of the proposals was conducted and/or evaluated under different conditions. It does also not address methodologies used to evaluate stories by people [1]. However, it considers the type of evaluation performed by the system during or at the end of the generation process with the purpose of deciding if the generated story can be presented as output, if it is worth further steps of the creative process, or if it should be discarded.

This article mainly addresses comprehensive methods or algorithms for the global story generation process. Thus, we tried to present work on more specific components of the generation process only within the scope of a comprehensive algorithm or method.

The focus of this article is plot generation. It does not focus on the generation of texts, especially the generation of literary prose.

In addition to being useful per se, the capability to automatically writing stories and other types of texts has several applications. Chapelle, Cotos and Lee [2] use automatic writing in the context of writing evaluation. Roscoe and colleagues [3] use automatic writing capabilities in an intelligent tutoring system for improving writing skills. Wright [4] presents an automated author of journal news. Harrison and Riedl [5] propose an approach that uses stories to teach believable behaviors to agents through a q learning algorithm. And, in the sixties of the last century, the linguist Joseph Grimes created possibly the first ever system for automatic story generation [6, 7] to help him explore and document the structure of the rare and unwritten Huichol language (Wixárika), which is spoken by indigenous people living in the Sierra Madre Occidental, Mexico. In spite of the obvious interest of applications of story generation, this article focuses solely on the story generation process, not on applications other than the generated stories.

We try to compensate for some lack of coverage, by providing richer details about the reviewed approaches.

The organization of the research on story generation in classes may follow different criteria. We briefly mention a few possibilities. Kybartas and Bidarra [8] categorize existing research according to the degree of autonomy of the system (ranging from manual to fully automatic). Ontañón, Zhu and Plaza [9] propose organizing criteria based on the main problem the system tries to tackle, namely the author cognitive process, the structure of the story, or the behavior of its characters. Margaret Boden [10-12] defines different classes of creative processes (i.e., combinational, exploratory, and transformational). Most often, though, researchers present their proposals framed by the main used technological approach, such as story schemas, analogy, rule-based systems, planning algorithms, evolutionary algorithms, and learning. None of these criteria is ideal in the sense that, whatever it might be, many existing systems would be categorized in more than one category. Rather than following one of the mentioned approaches, this article presents a hybrid organization that, above all, is designed to better help the reader and the new researcher understanding selected issues that we believe may be some of the key topics in non-interactive story generation.

The categorization of the existing literature according to the degree of autonomy of the reviewed systems is the underlying principle of the recent overview of Kybartas and Bidarra [8]. It would make no sense to present another survey, following the same criteria, especially because of its close temporal proximity with the referred overview. However, to the extent that autonomy is a close sequent of intentionality, the current review devotes one of its initial sections to a related topic: the intentionality requirement (e.g., [13-15]), according to which, only intentional systems can be considered truly creative (see section 2.1).

In this survey, we have also addressed the issues underlying the organization criteria proposed by Ontañón, Zhu and Plaza [9], namely the cognitive process of the author, the structure of the story, and the behavior of its characters. Although it has not constituted the main bulk of work on computational creativity, there is research regarding cognitive and computational models of the creative process (e.g., [16-20]), which are covered in sections 2.2, 2.3 and 2.4.

The concerns about narrative structure and about the behavior of the story characters is reflected in a comparatively larger body of work mostly aimed at creating stories with the right balance between the narrative goals intended by the author, either the system or its user (e.g., [21, 22]), the properties of good narratives (e.g., [23, 24]), and the behavioral coherence and believability of its characters (e.g., [25-27]). Given that these concerns are present in a large variety of story generation systems, we have decided not to use them as major classifying principles for the organization of this survey. Instead, when describing the reviewed work, we have explicitly noticed and discussed these concepts (e.g., narrative principles, narrative goals, coherence and believability) whenever appropriate. Additionally, we included a section about the efforts made by several researchers to improve the quality of the generated stories that further delves into narrative structure and behavioral coherence and believability of the story characters (section 9).

Instead of creating a section for each class of system following their type of creativity, the classification of automatic story generation approaches according to whether they exhibit combinational, exploratory, or transformational creativity [11, 12] is presented when analyzing the reviewed approaches. The rational for this decision is twofold. First, it is easier to follow a classification based on the used technology because it corresponds to the way researchers frame their work. Second, the different kinds of creativity are closely associated with specific technological choices, although with exceptions. Story generation systems based on search algorithms such as planning-based approaches (e.g., [28-31]) are mostly associated with exploratory creativity. Case-based reasoning especially when adaptation combines several cases (e.g., [9, 32), and evolutionary algorithms (e.g., [33, 34]) are the kind of technological choices adequate for combinational creativity. Finally, following Boden [11], evolutionary programming is the technological approach best suited to transformational creativity. However, evolutionary programming has not been used for automatic story generation. The comparatively less frequent transformational creativity occurs in approaches in which the creative process transforms the original search space (e.g., [35, 36]) or the rules used to search it (e.g., [37], as discussed in [38]). Transformational creativity is not associated to a specific group of story generation technologies therefore it is noted within the analysis of specific proposals.

This overview starts by presenting the intentionality requirement for creativity, general approaches to computational creativity, models of human creativity, and models of human creative writing.

Then, it presents the technological approaches most used in story generation systems. A variety of technological approaches have been used in story generation. Possibly, the approach that has deserved the most part of the attention of computational creativity researchers is the use of planning algorithms to generate the sequence of events that will constitute the contents of a story (e.g., [21, 24, 29, 31, 39-43]).

Aside from AI planning, several other approaches have been widely used. Story schemas or templates (e.g., [44-48]) specify the sequence of optional and mandatory story events, or the grammar of valid stories, or even smaller event patterns that may be gathered to form a larger story.

Analogical reasoning (e.g., [32, 37, 40, 49, 50]) includes two classes of approach. One, case-based reasoning (CBR), instantiates, modifies and possibly combines concrete stories, story schemas, or story fragments, stored in the system's case base to create the desired story. The other, sometimes referred to as computational analogy, relies on algorithms that identify the analogy between given source and a target domains, and then apply the created analogy to a story (or part of it) in the object domain, generating a story (or part of it) in the target domain.

Rule-based reasoning (e.g., [25, 51, 52]) uses a set of rules to control the story characters, including their actions, their beliefs and their goals, and/or to make decisions involving narrative knowledge, for instance changing the frequency with which other rules are evaluated.

Although to a lesser extent, the computational creativity community has proposed evolutionary algorithms (e.g., [33, 34]) to create stories by repeatedly modifying or combining existing ones.

Learning-based approaches have also been used. It is possible to distinguish two kinds of learning-baes approaches: those that learn explicit knowledge representations (e.g., story schemas or scripts) that may then be used to generate the stories and those that learn to immediately generate the story without the intermediate learning of explicitly represented narrative knowledge. To the former group, we call explicit knowledge learning (e.g., [53-57]). And we call implicit knowledge learning algorithms to the latter group (e.g., [58-62]).

In spite of its diversity, this large number of technological approaches is often not enough to organize all research on story generation. The work of particular authors may not fit neatly in any of these categories, or may mix several of them. It happens often that case-based reasoning systems integrate schemas as a way to represent story abstract specifications in the cases (e.g., [32, 63]). Other examples of hybrid approaches abound, for example analogy and planning (e.g., [39, 40, 64]), story schemas and planning ([31, 42]), and learning and story schemas, as explained about the explicit knowledge learning algorithms.

Some researchers on automatic story generation classify their approaches (e.g., [25, 65, 66]) as simulated-based story generation (or emergent story generation), which consists of generating stories from the interactions among the set of autonomous agents corresponding to the story characters and possibly other agents with a global view of the story (e.g., the story director, in [25]). Although presented as an approach to automatic story generation, the systems following it use very different technological approaches. For instance, the Virtual Storyteller [25, 67] uses rule-based and case-based reasoning, FearNot! [65] uses rules and planning, and the system proposed in [66] uses planning. Therefore, this survey does not include a section on simulation-based story generation. Rather, in each reviewed class of story generation approaches (e.g., analogy, rules, and planning), whenever appropriate, we devote some space to corresponding distributed approaches.

The last section of the survey before the concluding remarks discusses possible strategies to improve the quality of the generated stories: use rich characters, use better narrative knowledge, extend the space of possible stories, and empower the systems with better models or better knowledge that enables them to avoid making unprincipled random decisions.

In the concluding remarks, we suggest several research areas that could potentially contribute to significantly improve the generated stories. The suggested areas relate to a wide range of story generation systems' concerns, from the autonomous generation of the main idea do be conveyed to the autonomous design of the creativity criteria the system might want its generated outcomes to fulfil.

## 2   Intentionality requirement and models of creativity

Before addressing the most important requirements and the models of creativity, it is important to discuss if there is a differentiated process responsible for the generation of creative ideas and artefacts, and if so, what that process might be.

Hodson [68] argues that there is no special process that generates creative outcomes. Creative outcomes result of the same processes used in general problem solving. According to Hodson, being creative is a quality of some outcomes, a property that results of an evaluation *a posteriori* of the generated artefacts, not of the processes. In fact, diverse models of creativity could be considered models of problem solving in general. For example, the model of creative thinking of Newell, Shaw and Simon [16], although presented as a model of creativity, could be seen as a model of general problem solving. Even processes often associated with creativity such as conceptual blending (e.g., [69]), have been presented by the authors that originally studied them as pervasive processes of human cognition.

As evidence to his claim, Hodson [68] showed that a simple process, involving a simple reasoning infrastructure, is capable of generating undisputable creative ideas, such as "*Every resurrection destroys some dead person*" and "*No time lasts longer than forever*".

In his view, the same processes that sometimes generate creative artefacts, often (if not usually) generate artefacts not deemed to be creative. Hodson suggests that creativity emerges at the intersection of many sub-systems, the repurposing of cognitive pathways, the available information and the particular circumstances. Creativity emerges of a rich, practically unbounded set of combinations of generative behaviors. Since there are no creative processes, every creative artefact can conceivably have originated from a different mixture of underlying cognitive processes.

In spite of Hodson's belief that there is no special purpose process that could be taken to produce creative outcomes, there is a significant body of work proposing more computationally oriented or more human centered processes that would generate creative outcomes. As Pérez y Pérez puts it [70], computational creativity is an activity that can be situated anywhere in a continuum between two possibly complementary extremes: the engineering mathematical approach and the cognitive social approach.

Curiously, diverse general descriptions of creative processes agree with respect to some aspects. The most important of these communalities consists of seeing the creative process as a repetitive process in which the evaluation of the output generated at a certain stage leads to further improvements until a sufficiently good outcome is produced (e.g., [16, 17, 71-73]).

Instead of starting with a well-defined goal or problem, as it is the case in general problem solving, creative processes start with ill-defined problems sometimes specified by a set of constraints (e.g., [16, 17]), which opens up the possibility for serendipity (e.g., [72, 74, 75]).

Another communality is the belief that, during the generate-and-test process, the intermediate results produced at a given stage should be externalized (e.g., [17, 20, 76, 77]).

This section analysis the challenging requirement that creativity must be intentional. Then, it describes some general approaches that may be used in computer systems to generate creative outcomes, in principle, in any creative domain. It then moves to models of human creativity, either at an individual level or at a social level. Finally, the section gives an account of existing models of human creativity in the particular domain of story writing.

## 2.1 The intentionality requirement

Possibly, the most challenging goal of computer creativity is designing and implementing computer agents capable of intentional creativity. This philosophical problem involves several other problems of artificial intelligence that remain without widely accepted answers, such as the problem of computer intentionality [78] and computer first person perspective [79], and the question of whether a computer could ever be accepted as part of the human moral community [11].

Guckelsberger, Salge and Colton [13] believe that people will not consider a computational system to be creative if it is incapable of providing an account for its own intentional creativity. In addition, Ventura [14] and Bay, Bodily and Ventura [15] recognize that saying that a system is intentional just because it has goals that might have been given to it by the designer is not entirely satisfactory. For Bay, Bodily and Ventura, systems will be considered more creative if their intentions are mutable and result of some form of inspiration.

This subsection presents diverse proposals regarding the necessary conditions for a creative system to be intentional. Curiously, most of authors reviewed in this section emphasize the role change plays in intentional systems (e.g., [13, 77, 80, 81]). Although with very different rationales, these authors relate intentionality with the capability of both reacting to changes and changing themselves. Some authors (e.g., [14, 81, 82]) see intentionality in systems whose internal process correlates with their products and that are capable of explaining them at the light of such correlation. Finally, researchers also discuss possible roles and targets of intention in creative processes [15].

Guckelsberger, Salge and Colton [13] propose that the intentional creativity of a system must be grounded in the maintenance of the system's identity. According to these authors [13], autopoietic creativity of an organizationally closed system is the system's active modification of its structure to ensure its continuous existence. Nevertheless, given that the system is closed (the external environment cannot modify the system's structure), it does not have to generate non-trivial modifications of its structure. Therefore, organizationally closed systems may be only minimally creative. However, systems situated in their environment through their physical embodiment will always be subject to entropic forces (e.g., perturbations and energetic dependencies) that may lead to their disorganization. Thus, adaptive embodied systems must be capable of undergoing a variety of significant structural changes aimed at maintaining their own existence and identity. This means that only adaptive systems situated in their environment through their embodiment must be capable of intentional creativity, i.e., autopoietic creative activity grounded on the maintenance of their own existence and identity. Although in a paper not specifically related with computational creativity, Botelho, Nunes, Ribeiro and Lopes [79] do not think intentionality must result of the system's striving to preserve its identity. Rather, they argue that programs may become intentional agents if they discover the goals that would have caused their behavior (as if they were intentional in the first place) and then adopt those goals.

According to Guckelsberger, Salge and Colton [13], the usefulness of the systems' autopoietic creativity is its contribution to the maintenance of their continuous existence and identity. They exhibit novel behavior when they are either responding to a familiar perturbation in a different way, or when they are responding to perturbation that they have never encountered before.

Given that the system's specific embodiment (i.e., the way it is connected to its environment) determines the types of autopoietic creativity required from and allowed to them, Guckelsberger, Salge and Colton conclude that the autopoietic creativity of adaptive situated systems can only be assessed taking their embodiment in consideration, not ours.

Linkola, Kantosalo, Männistö and Toivonen [80] claim that intentional creativity (creative autonomy or intrinsic motivation, in their own words) may be exhibited by meta-creative systems, that is, systems with the capability to reflect on their own creative processes and to adjust them so that they can create artifacts outside the control of the programmer. Ventura [14] also considers the benefits of meta-level creativity. Ventura argues that building a hierarchical system that incorporates meta-level creativity capable of improving base level creativity would increase the overall system's autonomy, distancing it from the original designer.

Linkola, Kantosalo, Männistö and Toivonen [80] propose that self-reflection and self-control (two complementary aspects of self-awareness) allow a (creative) system to make justified decisions about its own behavior, in particular to monitor and adapt its creative processes to the current circumstances. These authors [80] define several types of self-awareness, including but not limited to artifact-awareness, generation awareness, goal awareness and meta-self-awareness. In a meta-creative system, all these forms of awareness, comprising reflection and control, are exerted by manager components over target components. In all cases, reflection and control may be weak (black box access to information of the target component) or strong (access to information and control of the inner parts of the target). If the system possesses apt connections between reflection and control, it may evolve meaningfully during its lifespan. These connections may be exogenous (provided by the programmer) or endogenous (learned by the system). Finally, the paper proposes six kinds of creative systems, depending on the presence of the mentioned properties (the several types of self-awareness, the types of reflection and control, the types of connections between reflection and control). These include but are not limited to creative (as opposed to mere generative), self-transforming and autonomously creative systems. From these ideas, the authors propose that an intentional creative system must have several components (target), some of which are monitored by other components (managers) capable of reflection and control linked by apt connections.

Grace and Maher [77] emphasize the importance of unexpectedness in assessments of creativity. According to Grace and Maher, a creative system acts intentionally if, in response to unexpectedness associated with generated or observed concepts or artefacts, the system autonomously changes the space of acceptable concepts, the space traversing process, or both so that it may encounter (generate) more such unexpected concepts or artefacts. Such a system is an intentional transformative creative system.

Bodily and Ventura [81] propose that a creative system must be capable of autonomously selecting and changing its aesthetic in accordance with its meta-aesthetic, which they think should be explainability. That is, the systems' aesthetic must be explainable. An explainable aesthetic makes it possible for a creative system to communicate the value of its creations to others. An explainable aesthetic makes it possible for others to understand why an artifact is valued and possibly, then, to appreciate it (more) themselves.

Although Ventura [14, 82] agrees that a computationally creative system must exhibit intentional behavior, according to the conceptualization of intentionality proposed in [14, 82], the system does not have to invent its own goals. Ventura proposes two related conditions for intentional creativity: the correlation between the system's outcome and the system's process, and the capability of the system to explain its product. Following Ventura, the system may be said to exhibit intentionality if its product (the artefacts it produces) is correlated with its process. For example, through its aesthetic-based evaluation mechanisms, the system has the intention to generate only certain kinds of products. For Ventura [82], if a system generalizes a set of inspiring exemplars of valuable artefacts, proposes new artefacts, evaluates them according to some objective function and outputs only the best ones, then it may be argued that the system has some intentionality. It intends to generate artefacts that are at least as good and novel as specified by its objective function. Additionally, intentional systems must be capable of explaining why and how they generated the produced artefact (e.g., showing source material for musical inspiration) [14].

In addition to discussing the conditions under which a creative system is intentional, researchers also discuss the roles and targets of intentions in the creative process. For Bay, Bodily and Ventura [15], intentions may shape the creative process in isolation from each other or in combination. Intentions may direct the creation of an artefact, may be used to evaluate an artefact, and may be used to select or reject the entire artefact or parts of it. Irrespective of their role within the creative process (e.g., generation process, evaluation and selection), intentions may concern the purpose of the artefact (e.g., emotion), its sociocultural context (e.g., language), or its structural organization.

## 2.2 General approaches of computational creativity

The body of work reviewed in this section focuses on proposals with two distinguishing properties. First, they do not claim to model the human processes or conditions (either social or individual) often associated with creativity or, at least, they do not emphasize such a research objective. Second, they were presented as general approaches to creativity not specific of particular application domains.

It is worth noting that several of the proposals presented hereunder argue that creativity should generate novel, unexpected or surprising artefacts but not too novel, unexpected or surprising, so that the aversion system of the audience does not reject them. Most of the proposed models rely on the hypothesis that creative systems must have a generative and an evaluative component, which is responsible for the desired creativity of the generated artefacts. However, we will present an exception, in which the generation of artefacts tries to ensure their creativity without an explicit evaluative component.

Creativity is often associated with novelty, both in the mind of the layman and as considered by the computer creativity community in general. However, although recognizing that novelty is a fundamental property of creativity, several researchers (e.g., Saunders and Gero [83], Colton, Charnley and Pease [84], Elgammal, Liu, Elhoseiny and Mazzone [73], and Saunders [20]) adopt the point of view that the generated artefacts should not be too novel. Stimuli excessively novel may activate the aversion system of the audience.

In the *Artificial Creativity* model of social creativity (e.g., [20, 83]), the evaluation of artwork by the agents of the community reflects this point of view. An internal model of preference based on the Wundt curve, where similar but different perceptual experiences are preferred, defines the preferred range of novelty for an agent.

Drawing from an analogy between creativity and dynamical systems, Valitutti [85] suggests that creativity may be seen as reflecting search paths (in the space of concepts) that lead to new basins of attraction. Maybe this suggestion by Valitutti may be related with the before mentioned proposal that creative artefacts should be novel but not so novel that they trigger aversion from the audience. If one speculates that basins of attraction, in the space of possible concepts, reflect different comfort zones of a putative audience then one may hypothesize that novel concepts are accepted if they fall within one of the comfort zones of the audience. This would mean that a novel concept can be significantly different from the ones that served as inspiring set for its creation if only it can be perceived as sufficiently close to another region of accepted concepts.

Ventura [14] proposes that computationally creative systems must have generative and evaluative components (an *aesthetic*, in [14]). The generative components generate internal representations of proposed outcomes, which are evaluated according to criteria designed to operationally assess creativity (e.g., value and novelty). When the generated outcome satisfies the creativity evaluation, it may be rendered as an external artefact that is also subject to evaluation before being considered the system's outcome.

Elgammal, Liu, Elhoseiny and Mazzone [73] propose the Creative Adversarial Networks model of creativity (CAN). CAN extends the GAN model (Generative Adversarial Nets, [86]) to be capable of generating more innovative artefacts than those generated by GAN, but not too much innovative. The generative adversarial network consists of two neural networks, the generator and the discriminator. The generator network generates proposals while the discriminator classifies (or ranks) the generated proposals according to the learned criteria. The generator uses the output of the discriminator network to improve its generation strategy so that the discriminator evaluates more positively more of its outputs. According to Elgammal, Liu, Elhoseiny and Mazzone, the GAN model tends to an equilibrium in which the generative network generates outputs of exactly the same distribution used to train the discriminative network. That is, in equilibrium, the generated outcomes do not reflect any novelty relative to the dataset used for training. The proposed CAN model [73] allows to improve the novelty of the generated artwork while keeping it sufficiently low to avoid triggering aversion in the audience.

As GAN, CAN comprises a generator and a discriminator. However, the discriminator produces two outputs (instead of one): the classification of the received input as art (as in GAN); and the confidence of the network that the received art belongs to a given style. The generator learns to optimize two criteria: to decrease the confidence of the discriminator that the generated art belongs to a given style, and to improve the chances that the discriminator classifies the generated outputs as art. Minimization of the first criterion contributes to novelty. Maximization of the second one contributes to keep novelty within acceptable limits. CAN was demonstrated in the domain of painting generation but the same principles can be used in other domains.

Instead of using novelty and value as proxies of creativity, some researchers rely on other concepts such as unexpectedness. Yannakakis and Liapis [87] propose an evolutionary algorithm, aligned with the divergent search paradigm, named *Surprise Search*. As with the usual genetic algorithms, *Surprise Search* generates possible outcomes. When a generated outcome satisfies the evaluation criteria, it may be used as the system's outcome. The distinguishing characteristic of this proposal is that generated outcomes are evaluated according to their degree of unexpectedness. The algorithm uses a prediction function that produces possible expected outcomes from the already known examples. Then, it compares the outcomes generated by the evolutionary process with the expected outcomes. A generated outcome satisfies the evaluation criteria only if it is sufficiently different from the expected possibilities. The proposed algorithm does not make any commitment regarding the specific nature of prediction function or the comparison function, which may be more or less domain dependent. This approach can be used to generate creative artefacts, in principle, in any domain. Yannakakis and Liapis [87] demonstrated the proposed approach in a painting generation setting and in maze traversing problems.

Contrarily to the previously referred computational approaches that favor a balance between novelty and audience acceptance (not too much novelty), the *Surprise Search* algorithm is a merely divergent algorithm. Gravina, Liapis and Yannakakis [88] propose a new algorithm, also based on surprise, but that tries to reach an appropriate balance between divergence and convergence during search.

Guzdial and Riedl [89] proposed another general approach to generate creative outcomes, which complies with the more abstract generate-and-test pattern proposed by Ventura [14]. The paper presents a meta-search algorithm that combines different models learned by machine learning algorithms. This search process uses three combination operators: concept blending [69], amalgamation [90], and compositional adaption [91]. The search procedure of the proposed approach can be implemented as a diversity of existing search algorithms such as hill-climbing, simulated annealing, genetic algorithm, or reinforcement learning. The meta-objective function combines three widely accepted measures of creativity, which were proposed by Boden [10]: novelty, surprise, and value. According to the authors, combinatorial meta-search can discover new generative models for which data may never have existed and thus expand the space of possible creative artifacts that can be generated.

More recently, the same authors proposed a particularization of their initial idea for the special case of combining previously existing neural network models [92]. The new proposal can be used both to classify and generate new concepts that result of the combination of previously existing concepts for which neural network models have been trained.

Bodily, Bay and Ventura [93] propose a general probabilistic learning approach to computational creativity that does not match the generate-and-test blueprint put forth by Ventura [14] because, in their proposal, generation and validation are not separate processes. The system generates artefacts that satisfy the creativity criteria implicitly present in the dataset used for learning (but contrast with [94]). The generative process relies on the Hierarchical Bayesian Program Learning framework (HBPL, [95]), which is especially adequate for generating novel and surprising artefacts, particularly when used in classes of concepts with a rich compositional structure. HBPL represents concepts as simple probabilistic programs, i.e., probabilistic generative models expressed as structured procedures in an abstract description language. HBPL learns concepts by constructing the probabilistic programs that better explain the observations. HBPL builds these programs compositionally from parts, subparts, and relations among them. HBPL can generate new types of concepts by combining parts and subparts in new ways.

In addition to proposing the HBPL framework for computational creativity, Bodily, Bay and Ventura [93], recognizing the importance of intentionality in creativity, propose inspiration as the source of intention and use the HBPL framework to capture the notion of creative artefacts generated to satisfy an intention that emerged of some inspiration.

Bodily, Bay and Ventura [93] present their proposal as a general approach to computational creativity and demonstrate it in the domain of lyrical composition. Spendlove, Zabriskie and Ventura [94] used the same approach for six word story generation.

## 2.3  Models of human creativity

According to Saunders [20], there are two broad perspectives on human creativity: creativity as a mental phenomenon and creativity as a social construction. This section presents three models of human creativity. The first, due to Newell, Shaw and Simon [16], addresses the individual cognitive processes involved in creativity. The second, which is being worked on and used by several authors (e.g., Cardoso and colleagues [18] and Wiggins [19]), is based on a distributed cognitive theory aimed at modelling human consciousness, which was originally proposed by Bernard Baars [96, 97]. Finally, the third, due to Saunders and colleagues [20, 83, 98], is a computational model of the sociocultural processes of creativity.

According to Newell, Shaw and Simon [16], creative activity appears simply to be a special class of problem solving activity. Problem solving is deemed creative if its product has novelty and value (either for the thinker or for his culture); if it is unconventional, in the sense that it requires modification or rejection of previously accepted ideas; if it requires high motivation and persistence, taking place either over a considerable span of time (continuously or intermittently) or at high intensity; or if the problem, as initially posed, is vague and ill-defined, so that part of the task consists of formulating the problem itself.

Newell, Shaw and Simon [16] propose a general generate-and-test model relying on the use of imagery (which is a special case of alternative representation), heuristics to guide the search process, and learning new heuristics when necessary. The testing component of the proposed model should ensure the novelty and value of the generated outcomes. The use of imagery enables the creative process to learn and infer features of the concepts being processed that it would not notice if an alternative representation had not been used. These unforeseen aspects may lead the system to the often-called "illumination", which consists of suddenly coming up with a solution for the faced problem. The capability to replace the current subset of active heuristics with another adequate subset is responsible for unconventional thinking and for speeding up the process. Motivation and persistence, and the better and more rigorous definition of the ill-defined problem seem related with the "incubation" phase, which characterizes creative processes. Yet, the proposed model does not address "incubation".

Visual representations (imagery) are associated with specific processes, for which otherwise implicit information becomes explicit. For instance, when some concept is represented as an arrow, the order by which attention is allocated to the items connected by the arrow is immediately determined. This illustrates that the use of imagery enables the system to perceive information that otherwise would stay unnoticed. This may contribute essentially to novelty and processing speed (given that the newly acquired information may contribute to rapidly solve the problem).

In spite of the mentioned advantages, Newell, Shaw and Simon note that representing abstract constructs as visual imagery introduces the danger of inadequate inferences. For example, if the system represents some entity as a line, it can erroneously attribute it the property of continuity. One may thus conclude that a creative system should have the means to take advantage of imagery and yet to avoid inappropriate inferences.

While Newell, Shaw and Simon [16] emphasize the important role of imagery (a particular case of an alternative representation), Sharples [17] emphasizes the role of using different knowledge areas. According to Sharples, a writer works with multiple overlapping areas of knowledge, draws analogies between them, and solves encountered problems in a different space.

Sometimes finding an appropriate solution requires getting out of stereotyped thinking, that is, it requires unconventional thinking. Unconventional thinking can result from substituting a subset of heuristics for another subset of the problem solver's repertoire, causing the search to move in a different direction. It is then possible to conclude that creative systems must have rich subsets of heuristics, along with the means for selecting more suitable ones, where appropriate. Sharples [17] proposes that creative writers are capable of unconventionality by transforming mental structures to which they are accustomed using such transformations as "consider the negative", "remove a particular constraint", "substitute one structure for another", and "reconsider from a different viewpoint".

If the creative system does not possess an alternative adequate subset of heuristics to rely on, unconventional thinking will mean to go back to brute force generate and trial search in usually very large spaces. According to Sharples [17], even an accomplished writer will sometimes need to use general methods if faced with a new type of rhetorical demand. As noted by Newell, Shaw and Simon [16], to avoid the large costs of relying on brute-force reasoning in large search spaces, it is advantageous for the creative system to learn new heuristics.

Learning new heuristics may take different forms, from simply remembering a solution to a given problem to the most demanding case of learning the relation between successful and failing branches in the search space.

Remembering the path to a solution will contribute to speed up the search process of the creative system because it will be capable of using remembered paths to solve the same problems [16]. According to Sharples [17], expert writers can call on a large stock of remembered plans and schemas, built up through a long apprenticeship in the craft of writing. However, Newell, Shaw and Simon argue that if too many such paths are stored, performance will deteriorate. In such cases, it is a good balance to store the solved problems instead of the paths to the solution. If the system possesses more than one problem solving method, the remembered knowledge (solved problems or solution paths) must be associated to the methods for which it was useful.

Learning a new adequate heuristic may seem at first as a good idea. However, the space of possible heuristics may be even larger than the problem space. Therefore, the creative system should use knowledge of the problem being solved to guide the search for an adequate heuristic. Newell, Shaw and Simon [16] propose that the system learns the relation between the failing branches of the problem space and the successful ones. This learned relation may be useful in problems of the same class.

Cardoso and colleagues [18], and Wiggins and Forth [19, 99] present a distributed computational architecture that is based on a cognitive theory, originally proposed by Bernard Baars (e.g., [96, 97]), aimed at modeling human consciousness. The architecture comprises a Global Workspace containing the only information the creative system is conscious of, a number of Generators (associated to all the system's sensory modalities) that compete to have access to it, and an Associative Memory accessible to all generators. The associations between the representations of two tokens in this memory is the probability that one of the input tokens follows the other. This information is used by generators to predict the next input.

The generators make predictions regarding the next input token and generate content. In [18], these two functions correspond to different types of generators: predictors and concept generators. At each time, a threshold based selection mechanism chooses the generator that gains access to the global workspace. The selection mechanism will only select generators that produce significantly more information than the others. Information content is computed as in Shannon's information theory [100]. According to Wiggins and Forth [99], generators that are not selected for a large period are forgotten. This means they must empty their output buffers and start anew. Whenever a generator gains access to the global workspace, the contents of the global workspace are stored in the associative memory, and the information generated by the selected generator is copied to the global workspace.

Generators continuously sense the environment, possibly acquiring perceptual input. They use the probability information in the shared associative memory to predict the next input. Whenever the predicted input matches the actual input of the generator, it processes the input possibly producing new content (e.g., resulting of inferences made on the received input). The predicted input and the possibly generated new content are stored in the generator's output buffer.

The most noticeable form of creativity arises when there is no input. When there is no input, the generators produce outcomes conditioned only by the shared associative memory and by their internal state. When a generator gains access to the global workspace, the current contents of the global workspace are flushed to associative memory, updating the encoded probabilities, and the contents of the output buffer of the selected generator are copied to the global workspace. This process is responsible for the spontaneous pop up of ideas in consciousness as well as for their subsequent storage in memory for future use. According to Cardoso and colleagues [18], this mechanism simulates human spontaneous creativity. It is interesting to note that, due to the updating of the probabilities encoded in the associative memory, this spontaneously generated contents will also condition the subsequent parsing of input stimuli. That is, creativity is not reflected only through a conscious manifestation; it influences future processing.

For Wiggins and Forth [99], predictive parsing of input is both efficient and robust. Prediction allows the listener to get ahead of the speaker, and to reconstruct obscured or unclear parts of the input stream. Wiggins and Forth consider that these two characteristics (getting ahead of the speaker and reconstructing input) are restricted forms of creative behavior, which enables the creative system to permanently jumping to conclusions, correct or otherwise.

The described model reflects the principle of information efficiency, which has been adopted because information processing is an expensive resource [19]. This general principle entails three more specific operational principles. First, a creative system should be predictive: it should use its information not just to plan its future actions but also to anticipate future events. Second, a creative system should continuously process input from its environment to construct the most efficient possible model of that environment. Third, a creative system should pay attention only to those structures that contain sufficient information. The model provides not only for a continuous learning of new facts, but also for the adaption of its representations to account for new patterns in the data to which it is exposed. Since memory directly drives behavior, creative systems implemented according to the proposed model will adapt their behavior. According to the described model, attention is focused only on the memory regions with more information content [19].

The described model may be used in diverse application domains. Cardoso and colleagues [18] developed two concept generators. One of them uses concept blending to produce novel creatures (represented as concept maps) from a set of existing ones, in the context of procedural content generation for games. The other generator uses an evolutionary algorithm to develop context-free design grammars, which consist of very compact representations of large image spaces.

The model discussed so far is more than a model of creativity. It is a general cognitive model that may be used in with different purposes. Wiggins and Forth [19, 99] explored its use in the context speech and musical prediction.

Saunders, Gero and Grace [20, 83, 98] propose a computational model of the emergent nature of creativity in societies of communicating agents, called *Artificial Creativity*, which is an adaptation of the dual generate-and-test model proposed by Liu [101]. The proposed computational model aims at acquiring a better understanding of socio cultural creativity in human societies.

According to the *Artificial Creativity* model, the creative process relies on two fundamental concepts: the *field* and the *domain*. The *field* is the set of all individuals and institutions in the society. In the reviewed papers, all members of the *field* are creative systems. However, the authors suggest that it would be a good idea to populate the society with different kinds of agents besides the producers of innovations, including consumers, distributors and critics.

The *domain* is the set of artefacts, which were generated by members of the *field* and have been considered creative by the society. Thus, the *domain* represents the cultural or symbolic component of the system.

In the creative process, agents in the society (i.e., *field* members) use the artefacts of the *domain* as the starting point for new, yet sufficiently familiar, artefacts. When an agent generates an artefact (e.g., evolving existing ones), if its own evaluation of the artefact is greater than a specified value, the agent communicates its new artefact with its evaluation to other agents in the *field*. A generated artefact will become part of the *domain* if at least one agent in the *field* (except the creator) evaluates it above a defined threshold.

In the proposed model, the artefacts of the *domain* are not managed in a central database. Each agent of the *field* holds and manages a subset of *domain* artefacts.

Emergence is an important feature of the model. No individual can dictate the collective evaluations of creativity. They can only try to influence other individuals by exposing them to their products and their personal evaluations. Thus, the socio-cultural evaluation of creativity is an emergent phenomena. Creativity at the macro level also emerges of the creativity and evaluations of the individuals in the society. Creativity at the individual level is also an emergent property of the whole system because each individual is affected by its socio-cultural context. Agents in society generate creative products that are different from the ones they would produce in isolation. On one hand, when an agent evaluates the artefacts of another agent, it changes the creative process of the other agent (if the other wants to comply with the evaluation). On the other hand, when an agent exposes another agent to its artefacts, it changes the other agent's creative process by changing its notion of novelty and/or its notion of novel but still understandable artefacts.

Hence, both in the social / historic sense (H-creativity [11]) and in the individual sense (P-creativity [11]), creativity is an emergent property of this model.

According to the authors, the main difference between this model and the dual generate-and-test model [101] is that the socio cultural creativity test, in the dual generate-and-test model, is performed centrally by a single entity, whereas in the *Artificial Creativity* model, the socio cultural creativity test is performed by individual agents of the *field*.

In [83], individual agents use evolutionary algorithms [102] to generate artefacts and self-organizing map networks (SOM, [103]) to assess novelty (novelty is the categorization error of the network). The communication of artefacts among agents relies on the symbolic representations used in the agent's evolutionary algorithms. This way, the receiving agent can further evolve the received artefact without the need for reverse engineering.

Besides providing a computational account of socio-cultural creativity, the described model allowed researchers to study concomitant socio-cultural phenomena. Saunders, Gero and Grace [20, 83, 98] observed the dynamic formation of social structures (e.g., cliques) and social behavior (e.g., the ostracization of those who do not generate innovative work), the evolution of language and its acquisition by members of the *field*, and the evolution of *domains*.

## 2.4 Models of human creative writing

The engagement and reflection model of writing, originally proposed by Sharples [17], has been used and developed in diverse applications (e.g., the *Scéalextric Simulator* [104]), most notably in the automatic story generating system called MEXICA [37, 105]. The model was even used in creative domains other than writing, for instance software development [106]. The engagement and reflection model of writing was also rigorously formalized and analyzed by Alvarado and Wiggins [38].

Conceiving creative writing not just as a cognitive activity, but also as an open-ended design activity, situated in the writer's environment and mediated by tools and resources, Sharples [17] presents three properties of the writing process: the use of primary generators, the fusion of analysis and synthesis, and the use of tools and external representations.

Writers (as other designers) tend to adopt a relatively simple idea early in the design process, the primary generator. That idea limits the space of possible solutions and acts as a framework that guides the creative writing process. As the creative writing progresses, writers may replace or modify the primary generator.

There is not a clear division between analysis and synthesis, especially because the problems faced by creative thinking are ill-defined and the writer may change the primary generator during the creative process. Often, writers reason and learn about their writing through a series of trial solutions. Intermediate drafts act as generators of new ideas. Iteration is a fundamental aspect of the model: the synthetic process of translating ideas into prose can be interrupted by the analytic process of editing, which may in turn lead the writer into a new phase of writing.

A writer is a thinker situated in a material world, surrounded by books, files, notes, drafts and drafting tools, databases, reference books and colleagues, which together constitute a rich interoperative system.

In order to avoid cognitive overload, writers write all kinds of contents (e.g., ideas, intentions, schemas, annotations, reminders) on paper (or some other external medium) in the form of external representations that stand for mental structures. These external representation act as an external memory that supports or facilitates diverse activities of the creative process such as structuring and transforming the content, checking if the text is progressing in the desired way, rapid identification of parts that are incomplete or need revision, and the generation of compact descriptions of complex contents. The media used for the external representation conditions the use the writer makes of it. For example, a small notebook and a scrollable computer screen are used differently and facilitate different processes. Differently integrated external resources (e.g., online dictionary, colleague writers and domain experts) shape the whole creative process in diverse ways.

In the model proposed by Sharples [17], during the writing process, writers impose constraints to guide and control the generation of new material and the inadequate proliferation of story events, ideally ensuring that the output of the writing process is creative (novel and adequate). These constraints come from diverse sources, including the given writing task, external resources, and the writer's knowledge and experience.

Creative writing involves a repetition of two processes - engagement and reflection. During engagement (or engaged knowledge telling), the writer generates written material for consideration, conditioned only by the set of currently active constraints. During reflection, the writer reviews the generated material interpreting it as a potential source of new ideas to be explored and transformed, and creates new constraints or guidelines that will drive the next engagement period.

It is worth noting the remarkable resemblance of the engagement and reflection model of creative writing [17] and current neuropsychological accounts of the creative process as described, for example, by Fox and Beaty [107] and by Beaty, Seli and Schacter [108].

According to Fox and Beaty [107], the creative process is a dual stage process comprising unintentional content generation and intentional evaluation. In this neuropsychological description, the unintentional production of content corresponds to engagement, and its intentional evaluation corresponds to reflection.

Beaty, Seli and Schacter [108] provide a neurological understanding of the creative process, in which ideas are generated by the default mode network (a network of interacting brain regions that supports self-referential and spontaneous thought processes, e.g., memory retrieval), whereas the executive control network (a network of interacting brain regions responsible for the cognitive control of behavior, e.g., goal maintenance and attention control) is responsible for guiding, constraining, and modifying the idea generation processes to meet creative task goals. Establishing a parallel between this neurological account of creativity and the engagement and reflection model of creative writing, it is not difficult to conclude that the default mode brain network might be responsible for engagement whereas the executive control brain network might be responsible for reflection.

According to Sharples [17], writing may be creative if it is novel and appropriate to the task and the audience. The constraints writers impose to their creative process (e.g., selected knowledge schemas and rhetoric structures) aim to ensure the appropriateness of the writing. The generative power of language combined with the imposed constraints implicitly define the space of acceptable concepts, which experienced writers limit themselves to use.

A writing episode starts not with a single goal, but with a set of external and internal constraints. These come as some combination of a set task or genre (such as a college essay), a collection of resources (for example information on company performance to be pulled together into a business report), aspects of the writer's knowledge and experience, and a primary generator (or guiding idea). Experienced writers have the ability to create a generator that is manageable enough to be realized in the mind, yet sufficiently powerful to spawn the entire text.

Constraints guide the writing process. As the writing progresses, the writer may re-represent some of them in a more explicit form, as a conceptual space to be explored and transformed. The movement between engaged writing, guided by tacit constraint, and more deliberate reflection forms the cognitive engine of writing.

An engaged writer devotes full attention to creating a chain of associated ideas and turning them into text (as notes or fully fleshed-out prose). Working memory is entirely devoted to the task. While engaged in writing the writer thinks with the writing, not about the writing. The only other deliberative mental activity that the writer can perform while engaged in text creation is to speak the words out loud.

Reflection is an amalgam of mental processes. It interacts with engaged writing through reviewing, contemplation, and planning. Reviewing involves reading the written material, while carrying out minor edits. At the same time the writer explicitly represents the procedures used during engagement, including the constraints, making them available for analysis and transformation. Contemplation may involve more focused deliberate reasoning or the less focused exploration of associated ideas and analogies linked by some common emotion, theme or experience. Planning takes the results of contemplation and generates new sets of constraints and guidelines for the next period of engaged writing.

The proposed engagement and reflection process of creative writing is not used exactly in the same way by all writers. The duration of the engagement and reflection stages varies among writers. The way each stage is used by different writers is also different. For instance, some writers may want to plan everything ahead of writing, while others may want to immediately write almost everything and reflect only in the end. Sharples identifies five different writing profiles [17].

Alvarado and Wiggins [38] formally analyzed the engagement¬ reflection model in terms of the creative systems framework proposed by Wiggins [109] to describe, analyze and compare creative systems. The major conclusion of the analysis is that the engagement-reflection model is capable of transformational creativity. First, the conceptual space and the exploration strategy used during engagement, and those used during reflection are different. Therefore, the engagement-reflection model exhibits a simple form of transformational creativity resulting of the repeated switching between different conceptual spaces with different exploration strategies. Second, whenever reflection sets new constraints that guide the generation of further written material, it changes the exploration strategy to be used in engagement. This constitutes another form of transformational creativity.

The MEXICA story generation system (e.g., [37, 105], described in section 4) implements the engagement and reflection model of creative writing. Following the diverse nature and purpose of engagement and reflection, MEXICA uses two different technological approaches for these two stages of the creative process. Given that the purpose of engagement is the free generation of content (which is supposed to be as unconstrained as possible), MEXICA implements the engagement stage using a set of rule-like knowledge structures that suggest possible alternative events to add to the story when the story world matches specified patterns of emotion links and tensions between characters. Other more specific properties of the world (e.g., character locations) are not considered in this process. However, the choice of new events to add to the story during engagement is guided by active constraints. Thus, MEXICA uses a set of constraints aimed at ensuring desired properties of the story being generated (e.g., novelty and interestingness). Of the actions suggested by the rule like knowledge structures, MEXICA considers only those that satisfy the active constraints. Engagement ends when no event can be added to the story (an impasse) or when a certain (configurable) number of events have been added.

According to the Sharples' model, the main purposes of reflection include solving impasses, ensuring the coherence of the story events, and designing new more appropriate constraints for the next engagement stage. If the previous engagement stage ended with an impasse, in which it could not suggest a new event to add to the story, reflection uses case-based reasoning to find an action used in previous stories similar to the one being generated, in particular those containing the event before the impasse. To ensure the coherence of story events, MEXICA checks if the preconditions of all story events are satisfied. When it finds an unsatisfied precondition, it adds new events that can fulfill it. To create the appropriate constraints for the next engagement stage, MEXICA evaluates the story being generated with respect to the desired properties (e.g., pattern of temporal tension conveyed to reader, and story novelty). If the story being generated does not have the desired properties, MEXICA sets a new constraint that will condition the generation of appropriate content in the next engagement stage. The desired properties of the current story are evaluated against the case base of previous stories. The tension pattern of a story (either a previous story or the one being generated) is represented as a numeric vector encoding the values of the tension conveyed to the user along the story. The novelty of the current story is estimated by comparing its events with the events of the stories stored in MEXICA's case base.

Riegl and Veale [104] developed the *Scéalextric Simulator* which generates narratives consisting of diverse episodes, each of which is created by *Scéalextric* [47]. *Scéalextric Simulator* generates stories by simulating alternative solutions and choosing the best one. Each simulation implements the engagement-reflection model of creative writing, although with significant differences relative to MEXICA [37, 105]. Whereas MEXICA works by adding actions to the story in progress, *Scéalextric Simulator* adds full episodes. Contrarily to MEXICA, in which, each engagement stage ends after a few actions are added to the story in progress, the engagement stage of the *Scéalextric Simulator* ends only after the story being simulated is complete. Reflection, which occurs only in the end of each simulation, focuses on the impact of information transfer between characters on the global coherence of the story. The goal of the repeated simulation process is to find a story that maximizes global coherence amongst its episodic parts.

Gervás and León [72] propose the reflection construction model of literary creativity. The major idea of the model is that the feedback information resulting of the evaluation of the story being generated may lead to changes of not only the generated draft but also of the explicitly represented constraints that must be satisfied by the generated outputs. The proposed model is an extension of previous work by the same authors, in particular the ICTIVS (Invention, Composition, Transmission, Interpretation and Validation of Stories) model [76]. ICTIVS is more focused on the constructive approach to the creative process and does not explicitly considers the task of revising an already existing draft.

In more detail, computational creative literary writing is viewed by Gervás and León [72] as an iterative creative process alternating between two major stages: reflection and construction. The model applies the two stages to a representation called the *work in progress*, consisting of the draft of the ongoing literary output and the set of constraints that must be satisfied by the final result. In general, at each reflection step, the partial draft is analyzed in the light of the existing constraints, giving rise to a diagnostic. The current draft is partitioned into sections, according to the diagnostic. Each section is marked as either to be kept as it is, to be regenerated, to be revised or to be rejected. The creative process may also find it necessary to add new content to the *work in progress*. In that case, placeholders for the new content, marked as to be generated, are added to the partial draft. According to the proposed model, the reflection stage may involve the interpretation of the current draft, which consists of reverse engineering the set of its subjacent constraints. The diagnostic produced in the reflection stage results of the comparison of the current set of explicit constrains with the constraints extracted from the current draft.

At each construction step, the sections of the *work in progress* are subject to the four possible operations, as marked in the previous reflection stage: to be regenerated, to be revised, to be rejected, or to be generated.

There are four major differences between the reflection-construction model [72] and the engagement-reflection model [17]. First, the reflection construction process starts with reflection, whereas the engagement-reflection model provides support for different profiles of writers, varying from those that try to plan everything ahead of writing, to those that avoid planning altogether. Second, during construction steps of the reflection construction model, the set of constraints (not only the current draft) may also be changed, whereas in the engagement-reflection model, the constraints are modified only during reflection steps. Third, during reflection, the model reverse-engineers the constraints that could have led to the generation of the current draft and compares them with the set of constraints currently included in the *work in progress*. This process may find that some reverse engineered constraints should be added to the currently considered set of constraints, affecting the future creative process. This corresponds to the occurrence of serendipity: the constructive process leads to valuable features that did not originally exists but which may be noticed during the interpretation of the current draft. Finally, during reflection, the diverse sections of the draft being generated may be marked with five different levels of required revisions (to keep as it is, to be revised, to be regenerated, to be generated, or to be rejected), whereas the engagement and reflection model considers only changes that solve impasses and changes that fix the possible lack of coherence of the current event sequence.

As already noted by Alvarado and Wiggins [38], with respect to the engagement-reflection model, Gervás and León [72] emphasize that dynamically changing the system's constraints results in transformational creativity.

Gervás and León [72] establish an analogy between the reflection construction model of literary writing and the model of writing put forth by Flower and Hayes [71]. The model by Flower and Hayes comprises three basic process: planning, translating, and reviewing. These three processes are guided by a monitor that activates one or the other as needed. The planning process generates ideas and sets goals that can later be taken into account by any or both of the other processes. The translating process converts the generated ideas into words. The reviewing process evaluates the text being generated and revises it in accordance with the result of the evaluation.

According to Gervás and León [72], the ability to set goals for other processes as established during the planning process of Flower and Hayes's model [71] corresponds to the establishment of constraints, covered in the reflection construction model. The translating process of Flower and Hayes's model [71] corresponds, in the reflection construction model, to the generation of a new instance of a particular section of the draft (revision), according to the applicable set of constraints, as carried out during a construction stage.

# 3   Schema-based approaches

Schema-based approaches to story generation rely on the concept of script, which was proposed by Roger Shank and Robert Abelson [110] for representing sequences of typical events in both artificial intelligence and in cognitive science. The main assumption behind the majority of schema-based approaches is that a large class of different stories may be generated from a single story schema. This assumption arises of the observation that large collections of different stories share approximately the same structure and often content type as well. For instance, Vladimir Propp defined the schema shared by Russian folktales [111] and Benjamin Colby defined the schema of Eskimo folktales [112].

The first known story generation system, created by the linguist Joseph Grimes *circa* 1963 [6, 7], used a story schema-based on Propp's morphology of Russian folktales [111], which is a sequence of mandatory and optional narrative functions. Grimes never published a scientific account of his system. James Ryan [7] described the system from personal communications with the author and some non-scientific sources.

The system traverses the story schema and probabilistically chooses the optional narrative functions to include in the story. Each of these narrative functions is associated with the different concrete actions that may realize it. Grimes' system randomly chooses the actions to realize the selected narrative functions. The system generates natural language but it is not concerned with the generation of literary prose. Ryan [7] hypothesized that the actions available to the system were associated with templates for textual generation with slots for character filling, which the system randomly chose. However, the process of text generation is more sophisticated than merely filling the character slots with the specific character names. In fact, character references are not always the same across the text ("a lion" vs. "the lion") and the generated text expresses rhetoric relations between text segments (e.g., "thus"):

```
A LION HAS BEEN IN TROUBLE FOR A LONG TIME. A DOG STEALS
SOMETHING THAT BELONGS TO THE LION. THE HERO, LION, KILLS
THE VILLAIN, DOG, WITHOUT A FIGHT. THE HERO, LION, THUS IS
ABLE TO GET HIS POSSESSION BACK.
```

The present section describes three ways of specifying and using schemas in a story generation system. The first two of them are top down specifications, in the sense that the system starts with an abstract description of a complete story and proceeds towards its most specific, concrete local elements. The third schema-based approach is a bottom up specification in the sense that the system starts with short local patterns of story events, which it combines and instantiates until the whole story emerges. This section starts with the two top down approaches: explicit story schemas similar to scripts and story grammars.

The most used top down approach to schema-based story generation (e.g., [45, 46, 48, 111]) consists of explicitly representing the story schema as a sequence of mandatory and optional components, each of which corresponding to a part of the story. Some schema components may be repeated so that different stories may include them in different positions. Additionally to the sequence of story components, it is necessary to represent relations and constraints on those components (e.g., components with potential for being a good story ending, and causality and other dependency relations). In this approach, the first task of a schema-based story generation system is to traverse the given schema and select the parts to include in the story being generated. The major decisions in this stage (include a schema component in the story, and finish the story) depend on whether or not the considered component is mandatory, on dependency relations among components, on whether or not a similar component has already been included, and on whether or not the process has reached a component with potential to end a story. Different systems use different heuristics to make these decisions.

The second top down way of specifying a story schema (e.g., [112, 113]) is by means of a grammar that implicitly defines a space of possible stories, that is, a space of acceptable sequences of story components. As before, each schema component is often an abstract specification of the actual elements of the story.

Unfortunately, these two top-down approaches to schema-based story generation are not general, in the sense that they specify domain dependent story elements. For instance, the components of the Russian folktales identified by Propp [111] include, but are not limited to, villainy, hero reaction, and acquisition of magical agent, which do not apply to many domains. The components of the Eskimo story schemas identified by Colby [112] are also domain specific, including food lacking, spouse lacking, and villainy.

Either story schemas or story grammars are used to create a high-level specification of the most abstract components of the story being generated (e.g., villainy, hero departure, villainy punishment, hero return). In general the high-level story specification, produced by following the story schema or by applying the story grammar rules, is more abstract than the final elements that will be part of the story. Thus, in addition to the specified story schema or the story grammar and the process that uses the schema or the grammar to generate a specific high-level description, these two top-down approaches to story generation also include a procedure for instantiating each of the selected high-level story components with concrete, specific elements of the considered story world. To this end, systems use knowledge about the concrete elements of the considered story world (e.g., characters) and relations among them; and constraints regarding the concrete elements that may be used to instantiate the higher level story components.

The instantiation of the abstract components of a high-level story specification with concrete story elements, mostly actions, must satisfy constraints that have possibly been defined. For instance, when an abstract component is to be instantiated with a concrete action, it is necessary that the action preconditions (if any) are true in the state of the world in which the instantiation is being performed. On the other hand, if the action is actually used to instantiate a high-level specification, its effects will be reflected in the resulting world state.

The creative process that revolves about top down schemas does not exhaust itself in selecting and applying the appropriate schema or grammar rule for the current content and rhetoric goals. It can also involve the development and modification of schemas. Chambers and Jurafsky [53], McIntyre and Lapata [33], Li and colleagues [44, 114, 115] present approaches for the automatic learning of a story schema from a story corpus (section 8).

The systems representing any of the three discussed schema-based approaches exhibit exploratory creativity. In any case, their creative process explores the space of possible stories implicitly defined by the used story schemas (macro top down and micro schemas) or the story grammar, plus the knowledge used during instantiation. It may be argued that the systems that learn a story schema and then use it to generate specific stories (e.g., [44, 114, 115]) exhibit transformational creativity because, before exploring it, they define the space of possible stories.

In the introductory text of this article, we have mentioned the simulation-based or emergent approach to story generation, an approach in which the story emerges of the interactions of its autonomous characters. All the systems described in this section are monolithic systems, in which the generated story does not emerge of the autonomous interactions between its characters. On the contrary, INES (Interactive Narrative Emotional Storyteller, [48, 116]) is a distributed system consisting of a constellation of micro services, comprising Story Director, Plot Weaver, Plot Generator, Episode Generator, Emotional Tension Filter, Draft Reflector, Discourse Planner, and Sentence Planner and Linguistic Realization. In spite of being a distributed system, INES does also not represent the emergent or simulation-based approach to story generation because its component services do not play the roles of story characters. Each of them implements specific narrative tasks, such as plot generation and plot instantiation.

The following paragraphs present a story generation system based on the explicit story schema specified by Vladimir Propp in his Morphology of the Russian folktales [111], discuss the expressive potential of Propp's schema and its limitations, presents a story generation system that uses a different schema, and finally another schema-based system that learns the used schema.

Gervás [45] proposes a system – *Propper* – that generates stories from the narrative schema identified by Vladimir Propp for the Russian folktales [111]. The narrative schema is a sequence of mandatory and optional narrative functions, which are high-level specifications of the story events. Some narrative functions appear more than once in the story schema. Each narrative function in the story specification must be instantiated with a specific action before the story is generated. In addition to the sequence of narrative functions, the *Propper* system represents the set of narrative functions that may constitute adequate story endings and a set of dependency relations between narrative functions.

The *Propper* algorithm traverses the story schema and decides whether or not to include each narrative function in the story to be generated, according to a set of heuristics. Different sets of heuristics have been essayed. The decision regarding the inclusion of a narrative function takes into account if the narrative function is mandatory, if it is not coherent with other narrative functions previously included, and if the same narrative function has already been included before. Incoherent and duplicated narrative functions are not included. Mandatory narrative functions are included. The system randomly decides whether to include optional narrative functions.

The decision to end the story takes into account if the schema traversal process has already reached a narrative function that is believed to give rise to interesting story endings, and if there are still pending issues that must be solved before the story ends. Pending issues may appear due to dependency relations with already included narrative functions. The *Propper* system used four different heuristics: ($H_1$) end the story as soon as a good ending point has been reached, not considering pending issues; ($H_2$) if a good ending point has been reached, decide to look for another good ending point irrespective of pending issues; ($H_3$) end the story as soon as a good ending point is reached and all pending issues have been solved; and ($H_4$) if a good ending point has been reached and all pending issues have been solved, decide to look for a new good ending point.

The four heuristics to end the high-level story generation were evaluated with respect to three criteria: conformance with the original story schema, valid ending, and satisfaction of pending issues. All heuristics scored 100% on the conformance with the reference sequence. Heuristic $H_1$ was the best with respect to good story endings, but it scores very low on the satisfying pending issues. Heuristic $H_2$ performs better than $H_1$ with respect to the pending issues, but the length of the stories increases and the score on satisfactory endings decreases. Heuristic $H_3$ ensures a high value (96.8%) for the pending issues metric and a reasonable but not so good score on the potential for good endings. Finally, heuristic $H_4$ is worse than $H_3$ because it skips possible good endings.

Christopher Booker [117] proposes that there are seven basic plots such that all possible stories can be seen as instantiations of these. These plots are not mutually exclusive. Any given narrative may combine several of them into its overall structure, with some of these subplots possibly focusing on different characters.

Ronald Tobias [118] proposes the existence of 20 master plots, which belong to two different categories: plots of the body, involving mainly action; and plots of the mind, involving psychological development of the characters.

Georges Polti [119] proposed 36 dramatic situations that can be combined in the same story.

Gervás, León and Méndez [46] discuss the possibility of using Propp's narrative functions [111] to represent the proposals by Booker [117], Tobias [118] and Polti [119]. Gervás, León and Méndez present descriptions of Booker's and of Tobias' story plots in terms of Propp's narrative functions, although the authors have noticed that it was not possible to provide such descriptions for some story plots without changing the way the *Propper* system works. Sometimes the story plots identified by Booker or by Tobias are more general than can be described using Propp's narrative functions. Sometimes the appropriate description of Booker's or Tobias' story plots in terms of Propp's narrative functions would require that the roles of the used narrative functions be played by different characters. Finally, Propp's narrative functions are focused on what Tobias considers the plots of the body; it is not possible to directly use Propp's narrative functions to describe what Tobias considers the plots of the mind.

The dramatic situations proposed by Polti are closely related to single or to groups of Proppian narrative functions. However, some dramatic situations have no direct match to Proppian narrative functions. For these, it may be worth considering the introduction of new narrative functions. The set of Propp's narrative functions was extended with two new functions, but others would be necessary: *repentance* (e.g., to regret to have done some wrong deed) and *repentance rewarded*.

Gervás, León and Méndez [46] conclude that Propp's morphology, as formalized, is too rigid to capture appropriately the broad range of narrative schemas that have been reviewed. Propp's insistence that the narrative functions in his schema need be considered in a specific order introduces a restriction that reduces the expressive power that it might otherwise have had.

Concepción, Gervás and Méndez [48] developed a schema-based story generation ecosystem (INES, Interactive Narrative Emotional Storyteller) that does not use the story schema proposed by Propp [111]. Instead, it uses a subset of the cinematographic basic plots compiled by Balló and Pérez [120] (referred in [48]). INES ecosystem consists of a collection of micro services that implements the Afanasyev framework [121, 122]: Story Director, Plot Generator, Episode Generator, Filter Manager (Emotional Tension Filter, in the described case), Draft Reflector, and the discourse generation services (Discourse Planner, Sentence Planner and Linguistic Realization).

The Episode Generator and the Emotional Tension Filter are adaptations of the Charade system [123] to the Afanasyev framework.

INES possesses two global resources, available to all its micro services: Draft Repository and the Knowledge Base. The Draft Repository is a database that stores the ongoing story drafts. Each draft consists of a plot (the sequence of story events) and a space (story characters, living beings, objects and locations). The Knowledge Base contains all the knowledge about stories, including the story schemas (e.g., Destructive Outsider), the concepts used by the schemas (e.g., "community", "outsider", "arrival" and "hostile actions"), and a set of rules about the possible instantiations of high-level schema components with concrete elements of the story world (e.g., "When the community is a family, then the outsider can be a new partner, an unknown relative or a new lover". "When the outsider is a new partner, then the arrival can be a marriage". "Hostile actions can be the acts insult or kill"). These rules and the used concepts are represented in a graph.

The Plot Generator chooses one of the available story schemas. Then, using the relationships between the schema concepts, it selects all possible instantiations for the selected schema specified events. The Story Director sends the output of the Plot Generator to the Episode Generator.

The Episode Generator selects among the possible actions by using knowledge about the affinities among characters and by checking their pre-conditions against the story space. When it adds an action to the story, it computes the action effects and adds them to the story space.

The Episode Generator considers four types of affinities: foe (no affinity), indifferent (slight affinity), friend (medium affinity) and mate (high affinity). The affinity relation is not symmetrical, except for mates. That is, character x may think character y is a friend, while character y may think character x is indifferent. Character actions are associated with affinity types (e.g., some actions are associated with friendship, while others are associated with indifference and so on). The association of actions and corresponding affinities is used to constrain the actions each character may perform, involving another character. The actions performed by a character involving another may also change the affinity between them.

Affinity may change as a result of lack of interaction, in which case, it changes towards indifferent, or as a result of the interactions between characters. For instance, if a character interacts with another in a way that is inconsistent with their affinity, the affinity level, or even the affinity class may change.

The Emotional Tension Filter processes the output of the Episode Generator, removing all events that do not meet predefined emotional criteria. If the Emotional Tension Filter removes an event, the Story Director calls the Episode Generator again. When the current draft satisfies the emotional tension criteria, it is processed by the Draft Reflector.

The Draft Reflector checks whether all events have been instantiated as specified by the constraints represented in the knowledge base. The Story Director sends complete stories to the text generation micro services. In its current stage of development, the text generator does not try to generate a literary output. It just produces a human readable version of the story events contained in the story.

More recently, Concepción, Gervás and Méndez [116] extended INES with the goal of generating stories with multiple plotlines (which was also the goal of the planning-based story generation system developed by Porteous, Charles and Cavazza [42]; see section 6 of the current article). This new version of INES includes a new micro service, called Plot Weaver, which is responsible for interleaving the episodes of two plotlines. It receives a master plot and a secondary plot and produces a new plot resulting of interleaving events of the two received plots.

In its new version, INES's Story Director chooses a template and asks the Plot Generator to create the first plot, according to the selected template and a theme also selected by the Director. This first plot will be used as the master plotline. When the first plotline is ready, the Story Director asks the Plot Generator to generate a secondary plot using a template that shares character roles with those of the master plotline, and a different theme. This time, the choice of the template is the responsibility of the Plot Generator. To allow this operation the story templates have a set of metadata describing the template, namely the roles of its characters.

After the two plots have been generated, the Story Director calls the new micro service, Plot Waver, to merge the two plots. Although there would be different ways of merging two plotlines, INES's Plot Weaver takes an event from one plot, another event from the other plot and tries to concatenate them. This is possible only if the post conditions of the first event are consistent with the preconditions of the second event. If the two events cannot be concatenated, the Plot Weaver skips the considered event of the master plot and tries the same process with the next event. If the two plotlines cannot be merged, the Plot Weaver informs the Director, which, in turn, provides for the generation of two different plotlines as before and repeats the same process.

In the described process, the Plot Weaver always takes events from the secondary plot and merges them into the master plot. When no more events can be merged, the remaining events are appended in the end.

Li and colleagues (e.g., [44, 54, 114, 115]) developed *Scheherazade* (and later, *Scheherazade-if*), a schema-based story generation system, that first learns the story schema (i.e., a plot-graph) from a crowd sourced collection of stories and then generates the story from the acquired schema. The process by which *Scheherazade* learns the story schema is detailed in section 8. The schema-based story generation process (i.e., the generation of a story from the learned plot graph) is described next.

A plot graph is a set of nodes representing story events and a set of arcs representing order and mutual exclusion relationships between events. An order relation is represented as a collection of facts $B(e_1, e_2)$ meaning that $e_1$ must occur before $e_2$. The defined order relation is a partial order because there may be pairs of events, for which no ordering constraint has been learned. If a pair of events $e_1, e_2$ is not ordered, they may both occur in any order in the same story. Mutual exclusion relationships, $M(e_1, e_2)$, define events that cannot occur in the same story.

Given the above, the plot graph defines a space of linear event sequences constituting valid stories on the given topic. The possible stories implicitly represented in a plot graph result of the existence of unordered and/or mutually exclusive events. If events $e_1, e_2$ are not ordered, it is possible to generate stories in which $e_1$ occurs before $e_2$, and stories in which $e_2$ occurs before $e_1$. On the other hand, if $e_1$ and $e_2$ are mutually exclusive, it is possible to generate stories that include $e_1$ and stories that include $e_2$ (but not both). Thus, in *Scheherazade*, story generation consists of traversing the learned plot graph and choosing the events to include in the story being generated.

Story generation can rely on standard search techniques such as A*. Each arc of the search graph represents a story event connecting one node to the next. Each node implicitly represents the sequence of events from the initial node. At each step in the search process, *Scheherazade* selects the node with the best evaluation and computes the set of alternative events that can be added to the story from the selected node. An event can be included when all of its direct predecessors have been included, except those excluded by mutual exclusion relationships. The criteria used to evaluate the nodes and to select the best one relies on event probability information learned during the schema learning process (e.g., the probability that an event occurs in a given story, the probability that two events occur, and the probability that one event occurs after the other given the probability that they both occur). The story ends when one ending events is reached.

Li and his coworkers suggest that different evaluation criteria may be used for story evaluation, which would produce different stories. For instance, prototypical stories can be generated from a given plot graph, preferring the stories whose event sequence is the most likely. Other heuristic story generation approaches can be used to generate more interesting stories about the same situation, for instance by finding a legal sequence with an unlikely event, with likely events that occur in an unlikely ordering, with non-adjacent events that typically occur adjacently, with pairs of events that have low co-occurrence probability, or excluding an event that frequently co-occurs with another one already included in the story.

The above description finishes the presentation of schema-based approaches to automatic story generation for which the story schemas consist of an explicit sequence of mandatory, optional and sometimes mutually exclusive events. This section proceeds with the presentation of another top down schema-based approach, in which, the valid sequences of events are implicitly defined by a set of story grammar rules.

Benjamin Colby [112] proposed a grammar that implicitly describes the schema of all known Eskimo folktales, but apparently he did not build a system for story generation.

An Eskimo folktale is a sequence of *eidons*, which are classes of narrative actions somehow similar to Propp's narrative functions. There are two kinds of *eidons* in Eskimo Folktales: primary and secondary *eidons*. Primary *eidons* represent the story main events. Secondary *eidons* modify or accompany the primary *eidons*. They act as dramatic devices or vehicles for the plot and, in special cases, substitute for primary *eidons*.

Primary *eidons* are hierarchically organized in top level categories (e.g., motivation and engagement), which may in turn be organized into subcategories. For instance, the motivation category has two subcategories: value motivation and immediate motivation. Each category or subcategory includes several *eidons*. For instance, the motivation category includes the food lacking *eidon*, the spouse lacking *eidon*, and the villainy *eidon*, among others. The engagement category includes the encounter *eidon*, the hospitality *eidon* and others. Finally, each *eidon* has several *eidon* varieties. For instance, the hospitality *eidon* has varieties such as "The protagonist is invited to a house where hospitality (genuine or feigned) is given", and "The protagonist is offered or given sexual hospitality by someone he did not previously know".

The valid sequences of primary *eidons* in a narrative are governed by grammar rules (e.g., "A *move* is a motivational *eidon* followed by n responses", and "A response consists of m engagement *eidons* followed by a resolution *eidon*"). The locations where secondary *eidons* may appear are also governed by rules.

Besides primary and secondary *eidons*, Eskimo Narratives may include other minor elements: comment by the narrator addressed specifically to the listener (often preceding a new episode) and a summarizing or explanatory terminal statement.

Raymond Lang [113] developed a grammar defining the sequence of events that constitute a class of simple stories. The following is an example of a grammar rule specifying that a story has a setting and a list of episodes, so that the temporal interval associated with the setting meets the one associated with the episode list.

```
story(story(Setting, Ep_list)) --->
  setting(Setting, S_time),
  episodes(Ep_list, Ep_time),
  {meets(S_time, Ep_time)}.
```

There are other grammar rules that define other concepts such as a setting and a list of episodes. For instance, a list of episodes is a non-empty list of components of the form ep(Ev, Er, A, O), in which Ev is the event initiating the episode, Er is the emotional response of the protagonist to the event, A is the action executed by the protagonist in response to the event, and O is the state that results after the conclusion of the episode.

The proposed model was implemented and tested in a system called *Joseph*. *Joseph* comprises an interconnected set of five components: the story grammar (as described above), a grammar interpreter, a world model, temporal predicates, and a natural language generator. The set of temporal predicates defines a variety of relations and operations among time intervals. The natural language generator produces the natural language outputs from the formal data structures produced by the remaining components.

The grammar interpreter receives requests to generate stories. It implements the search algorithm of the generation process, which uses depth-first, iterative deepening search plus random choice to find a sequence of grammar rule rewrites which defines a valid story. The story grammar produces abstract representations of stories but does not specify the bindings of variables contained in those abstract representations. The potential instantiations of such variables are drawn from the characters existing in the world, the actions they perform and the world states in which those actions are performed and the results of their execution.

The world model specifies the protagonist of the story, the facts that may be added to a story, the events that may initiate an episode (e.g., a quarrel between the protagonist and his/her spouse), singular and compound actions, their preconditions and effects, interdependencies between world states (e.g., if X is married to Y and Y is in a pit, then X lives alone), the emotions felt by the protagonist in response to events, the actions and goals that result of feeling an emotion, and the beliefs about the actions that may be used to achieve each goal.

The third class of schema-based story generation, namely the bottom up approach, is represented by the *Scéalextric* system, which was proposed and developed by Veale and colleagues (e.g., [47, 104, 124]). Contrarily to other described schema-based story generation systems, their approach does not start with a high-level comprehensive schema representing a set of complete stories. It also does not learn script like descriptions of full stories. And finally, it also does not use a set of grammar rules representing top down descriptions of full stories. Instead, their approach is based on a large knowledge base containing about 3000 very specific and local schemas representing small event sequences that may possibly be included in a story. Each schema is a sequence of three events, involving two abstract characters, *A* and *B*, in which *A* is the agonist and *B* is the antagonist, for instance:

> A read about B; B impress A; A flatter B
> A flatter B; B promote A; A disappoint B
> A disappoint B; B humiliate A; A attack B

In spite of the apparent resemblance of this knowledge base of event schemas and the case-based approach to story generation, the approach of Veale and colleagues does not use any form of analogy reasoning. Therefore, it is not a case-based approach.

The main idea underlying the proposed story generation approach is that such event schemas (the so called action 3 grams) may be combined in any possible way that satisfies the restriction that the last event of one schema must be the same as the first event of the next one. For instance, the above three schemas may be combined in the following plot skeleton:

A read about B; B impress A; A flatter B; B promote A; A disappoint B; B humiliate A; A attack B.

Different articles by Veale and colleagues constitute different approaches regarding diverse aspects of the proposed general story generation process. For instance, Veale [47] describes two ways to instantiate the character placeholders *A* and *B*. The first one consists of just selecting pairs of human like animals similar to those appearing in Aesop's fables. The other consists of using a large handcrafted knowledge base, called NOC (Non Official Characterization list [125]) containing very detailed descriptions of some 1000 iconic real and virtual characters, such as Bill and Hillary. If the NOC database is used to instantiate characters in the plot, the knowledge contained in the knowledge base about the chosen characters may also be used to constrain the way existing event schemas are combined to form the story. Veale and Valitutti [124] describe a way the structure of the story can be conditioned by the knowledge about its characters, which is contained in the NOC knowledge base.

Veale [126] proposes the Flux Capacitor system, which generates transformative character arcs, that is, it generates the start and end points of a character trajectory in a story. Given the two extremes of the character trajectory, the described general story generation process is used to fill in the events that connect the two points.

As a final example, Riegl and Veale [104] present an approach to create larger stories by connecting several short stories (or episodes) generated by the general approach. This process uses a large database of story continuations, which constitute possible links that enable to connect one episode to the next one. Two additional contributions proposed by Riegl and Veale [104] are the use of information transfer actions (i.e., actions that allow the reader to believe that information was shared by the characters involved in the action) and affective beliefs, which improve the story believability.

# 4 Analogy-based approaches to story generation

Reasoning by analogy involves the establishment of a mapping between two problems or two problem domains. The mentioned mapping relates concepts in one of the problems (or problem domains) with corresponding concepts in the other problem (or problem domain). This relation identifies the concepts of one problem that play the same roles as those played by the concepts in the other problem. Two concepts *A* and *B* are analogous if they play the same role in two different problems or problem domains. Hence the name reasoning by analogy. In the simplest case, if a solution to the analogous problem is known, the solution to the original problem can be generated by replacing the concepts of the solution to the analogous problem with the corresponding concepts in the original problem. Thus, even this simplest form of reasoning by analogy combines solutions to the analogous problem with concepts from the original problem to create solutions to the original problem. According to Boden [11], generating creative artefacts this way is a form of combinational creativity. A more compelling argument can be made about the presence of combinational creativity in analogy-based system when the solution to the original problem results of the combination of the solutions to several analogous problems. Thus, in general, analogy-based story generation systems exhibit combinational creativity.

The majority of the work on using analogy to automatically generate stories relies on case-based reasoning (CBR, [127]). Those systems possess a database of previous stories (e.g., MINSTREL [23], MEXICA [37] and *Riu* [49]), abstract story schemas similar to those described in section 3 (e.g., [32, 128]), or story fragments (e.g., [67]), which is called a case base. Case-based story generation systems use the stories, the story schemas or the story fragments stored in their case bases to create a new story or to advance the story being created. This section presents examples of these three varieties of the case-based reasoning approach to story generation. It also presents an analogy-based story generator that does not rely on a case base [40].

Analogy-based story generation systems receive a partial specification of a story to be generated, often called a query (e.g., a story with a princess, a knight, and a dragon that kidnaps the princess). Case-based reasoning systems use their case bases to generate a solution to the received story specification. In general, case-based systems retrieve from their case bases the set of the stories most similar to the received partial specification. Then, they produce a new story that satisfies the partial specification, adapting the retrieved cases (e.g., [37, 129]). Instead of retrieving from their case bases the cases most similar to the received story specification, MINSTREL [23] and the *Virtual Storyteller* [67] transform the received specification until it directly matches (part of) one of the stored cases. Then, the systems apply the reverse transformation to generate the story satisfying the original specification.

The system described by Riedl and León [40], which does not use case-based reasoning, receives a story in the source domain, the description of the source domain, and the description of the target domain. The system discovers the mapping between the source domain and the target domain. Then, it applies the discovered mapping to transform the source story into a story in the target domain.

Not all the analogy-based story generation systems receive a partial specification of the target story to be produced. Hervás and Gervás [130] created a system that uses case-based reasoning with the purpose of generating the surface textual rendering corresponding to a received formal specification of a complete story. This problem, although important, falls outside the scope of the present survey.

The majority of the reviewed analogy based system were presented as monolithic approaches to story generation. Swartjes, Vromen and Bloom [67] investigate character-driven (emergent) story generation as in the *FearNot!* [65]. In this approach, characters have the double role of being believable inhabitants of a virtual story world as well as improvisational actors that help to create an entertaining experience. To consider both believable character goals and narrative goals, the system uses a case-based approach in which the cases represent example pieces of stories with believable character behaviors.

MEXICA [37, 105] generates stories, using a model of creative story writing proposed by Sharples [17], called the engagement-reflection model of writing (see section 2.4). According to the engagement-reflection model, story writing involves two different processes that occur in sequence: engagement and reflection. MEXICA repeats engagement-reflection sequences until some stopping condition holds (e.g., it generates a good enough story). In engagement stages, MEXICA is focused on the production of story events avoiding the use of explicit goals or story structure information, guided only by the emotion links and tensions between characters and by active content production constrains. Engagement stops when MEXICA is not capable of producing more content (which is called an impasse) or when some user specified condition is met. In reflection stages, MEXICA solves impasses, adding a new event to the story, it ensures that the preconditions of all story events are satisfied, which may require the addition of new events, it evaluates the story being generated in terms of story progress, novelty and interestingness, and it sets or modifies content production constraints for the next engagement stage (e.g., if the story is not interesting enough, the new content production constraints must avoid the inclusion of non-interesting events during the next engagement period).

MEXICA is not a pure case-based reasoning story generation system. The two different stages of the story generation process (engagement and reflection) correspond to very different technological approaches. During engagement, MEXICA works much similarly to a rule-based story generation system. Case-based reasoning is used only during reflection.

MEXICA uses four knowledge sources that support the engagement-reflection loop. The first knowledge source to be given to the system is the list of possible actions. Each action in the list has preconditions and effects, both of which pertaining concrete story world properties, emotion links and tensions between characters, and the tension conveyed to the reader. The actions' effects are used both during engagement and reflection. The actions' preconditions are used only during reflection.

One of the most important knowledge sources used in MEXICA is the set of knowledge structures similar to rules, each of which associates a pattern of emotion links and tensions between non instantiated characters with a set of alternative events that may be added to the story. This knowledge source is important because it drives the engagement process.

The other most important knowledge source is the case base representing examples of good stories. Story representation, both the story being generated and those stored in the case base, includes its concrete representation (the sequence of the concrete story events), the emotional links and tensions between characters, which change from one state to the next, and the tensional representation, which is a numeric vector representing the value of the tension to the reader along the story progression. The importance of this knowledge source stems from its support to the reflection stages of the creative process, and because the knowledge source supporting the engagement stages is generated from the case base of previous stories.

Finally, MEXICA uses a set of rules that infer the consequences of certain states of affairs. The inferred consequences are those that could not have been computed at the level of the actions' effects because they depend on a combination of factors. These rules are used both during engagement and reflection. The rule conditions and consequences pertain the world concrete properties, emotion links and tensions between characters, and the tension to the reader.

MEXICA's creative process is initiated with the first story event, which is provided to the system by the user. The effects of the initial action and its inferred consequences are computed (through the list of possible actions and the set of consequence inferring rules) giving rise to the first state of the story being generated, which allows the first engagement stage to start.

During engagement, MEXICA uses the set of rule like knowledge structures relating patterns of emotion-links and tensions with alternative story events. When the pattern of a rule matches the story current state, MEXICA eliminates all possible actions suggested by the rule that do not satisfy the set of currently active content production constraints. If more than one action can be selected, MEXICA randomly chooses one of them. The selected action is added to the story without considering its pre-conditions. The action post conditions (as specified in a list of the available actions) and other consequences (inferred by the set of consequence inferring rules) are added to the story. The engagement process continues until MEXICA has performed a predefined number of transformations or until it cannot find an action to add to the story, in which case an impasse is declared.

In the reflection mode, MEXICA solves impasses by adding new actions, ensures that the preconditions of all story actions are satisfied, and evaluates the story in progress regarding story flow, novelty and interestingness. If the previous engagement stage could not find an action to add to the story (declared impasse), MEXICA uses the case base of story examples to determine all alternative actions that, in story examples, have followed the deed that triggered the impasse. MEXICA adds one of them to the end of the story in progress, modifies the concrete properties of current state of the story world, modifies the emotion links and tensions between characters, and modifies the tension to the reader. The action to be added must have its preconditions satisfied in the current state of the story world.

During reflection, MEXICA also checks if all actions in the story have their preconditions satisfied. If it finds an action whose preconditions are not satisfied, MEXICA adds a new action to the story whose effects satisfy the unsatisfied preconditions.

Still during reflection, after ensuring the coherence of the story actions, MEXICA uses the stories in its case base to evaluate the story being generated regarding interestingness and novelty. Interestingness is equated with degradation improvement processes. MEXICA determines if the tensional representation of the story in progress includes degradation improvement processes, by comparing it with previous stories. If the tensional representation of the story being generated is not similar to that of existing stories, MEXICA activates constraints that avoid the inclusion of non-interesting events, in the next engagement stage.

The novelty of the story being generated is assessed through its similarity with the stories stored in MEXICA's case base. If the story being generated is too similar to those in the case base, MEXICA activates a constraint that prevents the selection of events whose action has been employed often in the past.

After reflection stops, a new engagement stage starts. The sequence of engagement and reflection stages ends when the story being generated meets the specified conditions for a good story or when stopping conditions hold (e.g., the number of repetitions of engagement-reflection sequences).

When the engagement-reflection loop ends, MEXICA performs a last step called the Final Analysis. This process consists of inserting events that create explicit goals to be achieved by the characters in the tale. The Final Analysis helps to improve the coherence of the whole story.

The system's output is a story framework rather than a story, because MEXICA is not capable of producing full natural language.

*Riu* [63] is an interactive case-based reasoning story generation system. However, the approach can also be used in a non-interactive context. *Riu* has a repository of complete stories, each one being a sequence of story phases. The story generation process starts with the initial fragment of the desired story (target story), that can be provided by the user. Using the SAM algorithm (Story Analogies through Mapping), *Riu* identifies the complete story (the source story), from its case base, whose initial phase is the most analogous to the provided initial fragment of the target story. During this process, *Riu* discovers a mapping between the initial fragment of the target story and the initial phase of the retrieved case. *Riu* completes the target story with the following phases of the retrieved case, applying them the discovered mapping between the target and the source stories. In *Riu* [49], each story phase is represented in two interlinked forms: a computer understandable description (CUD) and a human understandable description (HUD). The CUD is a frame-based representation of the phase, whereas the HUD consists of a collection of pre-authored natural language phrases and sentences used for natural language generation. Each CUD representation is annotated with more abstract concepts of the force dynamics theory [131] (e.g., agonist, antagonist, exert force, resist force). This more abstract representation makes it possible to find deeper analogies between scenes of two stories. Two stories match, whenever they share similar force dynamics descriptions.

Originally, the story case base used by *Riu* [49, 63] was manually encoded. More recently though, the stories in *Riu*'s case base are semi automatically extracted from a collection of Russian folktales by *Voz* [132], a system that extracts narrative knowledge from text [55-57].

The approach described Valls-Vargas, Zhu and Ontañón [132] extends *Voz* to automatically generate the knowledge representation used in *Riu*'s case base (excluding the force dynamic annotations) from the knowledge extracted by *Voz* from a collection of stories (e.g., [55-57]). Section 8 describes the process by which *Voz* [55-57] extracts, from a collection of stories, the structured narrative knowledge required to generate the CUD and the HUD representations and the links between them [132]. The next paragraphs describe the extension of the *Voz* system to enable it to generate the representations used in *Riu*'s case base.

The CUD representation of each story phase includes the characters and entities of the corresponding phase, their properties and relations; and the sequence of story events of the phase (which the authors call expressions). HUD is an annotated natural language description of the story events in the phase. Annotations link the human readable description to the computer readable description.

To create *Riu*'s story case base from the output of *Voz*, it was necessary to generate the dual representation of story phases used by *Riu* from the output produced by *Voz*. This involves the segmentation of the story text into the several story phases, the identification and classification of the story entities (character vs. non character), the identification of the semantic role of each character, and the identification of the narrative function of the text segment corresponding to each phase.

The authors manually segment each story in five phases [132], each comprising the sequence of sentences corresponding to a Proppian narrative function [111] (e.g., introductory and setup functions, villainy/misfortune, mediation).

Each sentence in a phase is represented as a triplet called an event consisting of the verb, the subject and the object. Each event triplet extracted by *Voz* (as in [55, 56], and described in section 8) originates an expression in the CUD and an annotation over the span of text where the event verb and its arguments appear in the HUD. The roles (e.g., villain) of the characters (e.g., dragon), extracted by *Voz* (see [55, 56] and section 8), are also represented as expressions in the CUD (e.g., (role-Villain dragon)), which are added to the HUD following the template <Character> " is the " <Role> (e.g., (Dragon "the dragon") " is the " (Villain "villain")). The text realization component of the system uses these character role expressions, when producing the final the story text. Finally, each extracted sentence (i.e., each event triplet) is associated to the narrative function to which it belongs. Narrative function information is included both in the CUD and in the HUD. Valls-Vargas, Zhu and Ontañón [57] proposed an approach for the identification of the narrative function of each text segment (see section 8).

Rather than retrieving, from the case base, the case most similar to the received story specification and adapt it, MINSTREL [23] and the *Virtual Storyteller* [67] apply transformations to the received specification so that it directly matches (parts of) stored cases. Then, they apply the inverse transformation to the retrieved solution to produce the solution to the received specification.

MINSTREL [23] and the *Virtual Storyteller* [67] use a set of creativity heuristics to transform the original partial story specification into a new similar specification, including relaxation heuristics, generalization heuristics, heuristics for the substitution of a similar subpart, and planning knowledge. Concrete creativity heuristics include *replace some event that happens unintentionally with the same event happening intentionally*; and *replace an element in the story with a generalization of that element*. Generalizable elements include story world objects, character roles, actions and their actors. Generalization relies on a domain ontology.

MINSTREL [23] assesses the generated stories according to several criteria, for example interestingness. If a generated story satisfies all assessment criteria, it is output as a solution to the original problem. If the current creativity heuristic fails to generate a story or if the generated story does not satisfy all assessment criteria, MINSTREL selects a new heuristic.

The *Virtual Storyteller* [67] does not evaluate the generated stories. The authors believe that, at the time of their writing (2007), there was not a set of accepted well understood criteria for story evaluation. Instead, they claim that since the system's cases represent good quality stories, it is likely that the generated result is also a good quality story.

The cases of several CBR story generation systems represent abstract schemas of stories (or story fragments). In the systems described in [128] and in [32], each of those abstract schemas is expressed as a sequence of narrative functions, as proposed by Propp [111]. The cases in *Riu* (e.g., [49, 63]) and in MEXICA (e.g., [37, 105]) include the concrete story events and more abstract representations, which because of their high-level match different sequences of concrete events.

*Riu* [63] represents both the concrete events of previous stories and their abstract descriptions, which are expressed in terms of the force dynamics theory [131]. MEXICA (e.g., [37, 105]) uses a similar strategy. Each story in MEXICA's case base is represented as a sequence of concrete story events, emotion links and tensions between characters, and the temporal evolution of the tension conveyed to the reader. Both the clusters of emotion links and tensions between characters, and the evolution of the tension to the reader are more abstract representations than the story concrete events.

Propp's narrative functions (e.g., departure, villainy, struggle), MEXICA's emotion links (e.g., amorous love link of intensity +3 between the princess and the knight) and tensions (e.g., love competition tension between the knight and the prince), *Riu*'s force dynamics (e.g., the agonist resists the psychological force exerted by the antagonist), and MEXICA's tension to the reader (tension = 150) are increasingly abstract story representations, in the sense that the same description applies to an increasingly larger set of concrete situations. A villainy describes a large set of situations but less than a negative brotherhood love link between character roles. And a love competition between character roles applies to less situations than two characters exerting some force on each other.

Using abstract descriptions of specific stories facilitates the process of finding deeper analogies between a target story and an existing one [63].

When any of the systems described in [32] or in [128] receives the specification of the desired story, expressed in the same way as the cases in the system's case base, it retrieves the most similar case to the received description. Both systems use a domain ontology to compute the similarity between the received specification and the cases in memory. *Riu* [63] retrieves the case with similar force dynamics description.

As in schema-based systems (section 3), when a story schema is retrieved from the system's case base to generate the desired story, the abstract specification must be instantiated with concrete domain specific actions. In this instantiation process, an action replaces an abstract specification only if the action preconditions match the current state of the story world. The action effects are reflected in the story next state.

Case adaptation is possibly the most demanding problem of case-based system. The simplest form of case adaptation consists of replacing non-adequate terms of the case most similar to the current problem with the analogical terms of the current problem. More difficult solutions involve selecting several cases (instead of a single one) and producing a solution that results of the adaptation of the several selected cases. In the system proposed by Gervás, Hervás and León [32], if the retrieved case is not enough to satisfy the query (e.g., a story partial specification), a new query containing only the non-satisfied parts of the original query is presented to the system. The newly provided solution is combined with the solution generated so far. Ontañón, Zhu and Plaza [9, 129] use case amalgamation to merge the cases most similar to the target story partial specification. An amalgam of two terms *A* and *B* is a new term resulting of combining parts of *A* with parts of *B* [90].

While Ontañón, Zhu and Plaza combine several cases, using amalgamation, other solutions to combining several cases into a single solution are possible, especially conceptual blending [69] and compositional adaptation [91].

Permar and Magerko [50] propose a conceptual blending solution to the generation of a new story script from two input story scripts: the source script and the target script. Conceptual blending modifies the target script, with elements from the source script, in order to generate the new script, that is, the script blend. Their approach represents story scripts as direct acyclic graphs connecting events/actions.

The algorithm proposed by Permar and Magerko [50] consists of three stages: counterpart mapping, mapping selection, and mapping application. The algorithm identifies, for each input script, all possible paths between the start and the end nodes. Then, it tries to identify similar events in a pair of paths, one from each of the two input scripts, relying on their structural similarity, which retains the temporal ordering of the events. If node $N_i$ in the source script and node $N_j$ in the target script are found similar, a candidate counterpart mapping is proposed. Once this possible mapping is identified, further mappings may only be found on nodes occurring after $N_i$ (in the source input) and $N_j$ (in the target input). The described algorithm uses different criteria for the identification of similar nodes [50].

After the identification of all sets of possible mappings, the algorithm selects one of them, using a combination of criteria. To avoid losing iconicity, the algorithm discards mappings involving iconic entities from paths of the target script. Then it orders the remaining mapping sets, preferring mappings involving iconic entities in paths from the source script and mapping sets with more mappings. Finally, the algorithm generates the new script by modifying the target script with the selected mapping set.

Akimoto [133] proposes a general framework for blending two stories, which could also be used to create a new story from two stories previously retrieved from a story case base. The general framework proposed by Akimoto [133] corresponds to a significant extension of the most usual approach to story blending as proposed by Permar and Magerko [50]. In addition to the combinational creativity exhibited by the proposal of Permar and Magerko, the proposal of Akimoto is capable of transformational creativity. Akimoto [133] proposes the concept of abstraction as an operation that takes a story as input and returns a simpler view of the received story consisting only of the most important information contained in the story. The specific information extracted from the received story, by the abstraction operation, depends on the considered type of abstraction. The enhanced power of Akimoto's general framework stems from the fact that, before two stories are blended into a new one, they are subject to the specific abstraction operation chosen by the system. The story content extracted, from the received story, by one of the types of abstraction proposed by Akimoto, called *story-line abstraction*, closely corresponds to the script-like story representation used in the proposal by Permar and Magerko. However, the general framework proposed by Akimoto [133] considers different types of abstraction, for instance *story-line abstraction* (extracts the relational structure of events), *story-world abstraction* (extracts the relational structure of entities), *character perspective abstraction* (extracts events and entities that are relevant to a specific character), and *temporal or spatial setting* (extracts times or places from a story). Different types of abstraction give rise to different generated story blends. According to the Akimoto's story blending framework [133], the system dynamically selects the type of abstraction that better satisfies the current criteria.

Given the subjacent goal of creative story generation, independent of a particular environment, the blending of two stories, in the proposed framework [133], is directed only by *story-centric criteria* (difference and similarity to the input or pre-existing stories, and unity or coherence of the generated story). Difference is generally accepted as an essential condition of novelty whereas similarity is necessary for organizing or anchoring new information relative to pre-existing information. The choice of the abstraction operation to be used depends on the desired similarities and differences, for instance a similar aspect between the two input stories, a different aspect between inputs, or lack of information in the blend relative to the inputs. After the application of the selected type of abstraction to the two input stories, the information structures extracted from the two input stories are combined to generate the blended story. The system evaluates the blend and decides if it is already complete. If the current story is not complete yet, the system should repeat the process, possibly using different abstraction operations. When the story is considered to be complete, the system adjusts it considering the story coherence.

The described general story blending framework [133] enables the dynamic choice of the criteria to direct the blending process. Thus the system may choose the region of the space of possible stories to consider in each circumstances, which yields transformational creativity.

When the combination of several cases generates more than one possible stories, the system may evaluate them to choose the one to be output. In [9], stories are evaluated according to their preservation degree (degree to which they include the target story) and compactness (computed from the number of characters and the number of actions). More compact stories with higher preservation degree are considered to be better. The story evaluation stage in [129] determines if the story is under the attack of defined arguments. Arguments represent general knowledge about the world (e.g., the relatives of a human are also humans), story writing (e.g., the location of a story should not change from phase to phase), and specific common sense knowledge of concrete domains (e.g., a boat cannot be driven on land). Given that arguments represent soft constraints, stories may violate arguments. The quality of the results depends on the arguments that are provided to the system.

Riedl and León [40] proposed an approach that generates novel stories by analogy without using case-based reasoning. The system receives a story, Ss, in a certain domain (e.g., the American old west), the description of that domain (i.e., the possible actions and their preconditions and effects, and a set of propositions describing the initial state of the world), Ds, and the description of a new domain (the target domain), $D_t$. The system generates a new story, $S_t$, in the target domain, by analogy. In the first stage, the system uses the connectionist analogy builder algorithm (CAB, [134]) to produce a transfer function, f, that maps actions and ground symbols in $D_s$ to actions and symbols in $D_t$. In the second stage, the system uses the generated mapping function, f, to transform the source story, $S_s$, into a skeleton of a story in the target domain, $D_t$. Possible failures to find or apply mappings can leave gaps in the target story. A partial order planner, operating in the target domain, fills in the gaps, generating the final target story, $S_t$.

# 5 Rule-based approaches

Rule-based systems are computer systems with two distinguishing components: the knowledge base and the inference engine. The knowledge base of a rule-based system is a set of rules and facts. The inference engine is a program, usually included in the used rule-based tool, that uses the rules and facts of the knowledge base to solve problems (e.g., question answering and control). There are two fundamental types of rules. The rules of the first group represent such declarative knowledge as *if condition then conclusion*. When the inference engine of the rule-based system determines that the condition of a rule is true, it infers the conclusion specified in the rule. Logic programming [135] is the best known approach to represent this type of knowledge. The rules of the second group represent such procedural knowledge as *if condition then action*. When the inference engine determines that the condition of a rule is true, it adds the specified action to the set of alternative actions that may be executed. Latter, it selects one of these and executes it. The rules of this later type are called production rules [136]. Rule-based story generation systems may use rules both to control the behavior of the story characters and to represent narrative knowledge (e.g., for determining the required degree of tension, in the current state of the story being generated).

*Novel Writer* [51], *StellA* [52] and the *Virtual Storyteller* [25] are well-known examples of story generation systems using rule-based approaches. The engagement stage of MEXICA's creative process (see section 4) also works similarly to a rule-based system.

*Novel Writer* [51] generates murder stories within the context of a weekend house party, arising from possible motives of greed, anger, jealousy or fear. The specific motives arise as a function of events during the course of the generation of the story. The particular murderer and victim may vary randomly or according to the character traits specified prior to the generation.

*Novel Writer* comprises two major components: a story simulation component, which is used to generate the story plot, and a linguistic generative component, which is used to map the representations generated by the story simulation component into natural language stories. Both of these components are rule-based systems. Given that the generation of natural language is outside the scope of this article, *Novel Writer*'s natural language component is not described.

The simulation is governed by a set of stochastic rules with configurable degrees of randomness that determine the behavior of the individual characters in the universe. Each state of the story universe is represented as a semantic network whose nodes are objects and whose arcs are relationships between them. With time, the stochastic simulation rules change the semantic network representing the state of the universe, which is used as input to the discourse generation rules comprising the natural language generation component. The story flow is derived from reports on the successive changes of the universe.

Given that the *Novel Writer*'s rules can change themselves, the exploration of the space of possible stories also changes. This means that the *Novel Writer* may exhibit transformational creativity.

León and Gervás [52] created a version of the *StellA* story generation system that uses stochastic simulation to generate possible stories, and narrative knowledge to filter out possible branches of the story generation process and to select others. The story simulation process is governed by a set of stochastic rules that determine the possible story events (actions performed by story characters and other events, such as raining) that can take place in a given state of the world. The system applies all possible simulation rules to a given world state generating a set of alternative successor states associated with different probabilities. Then, *StellA* uses narrative knowledge to evaluate and select among the several alternative plots.

In *StellA*, there are five types of actions. *Deus ex actions* are performed by the system (not by the story characters) without any causal requirement; they represent events that are too serendipitous to be the result of a detailed model (e.g., raining). *Character perception actions* result in the acquisition of information about the state of the story world, that is, they create perceptions about the internal and external state of the character. *Character desire actions* create desires (e.g., the desire to eat or the desire to escape). Desires have priorities so that, in the next step, the character may choose to follow the one with the greatest priority. *Character intended actions*, if successfully executed, fulfil a character desire. *Physical world actions* result of the physical laws of the story universe (e.g., an object that falls if it stops being hold, or an object that moves if it is pushed).

All actions are associated with probabilities the system uses to decide what to try first.

There are sets of rules for each class of actions. Each rule establishes the probability of the corresponding action. The rules that create alternative sets of intentions (intended actions) of each character at each step reflect the principles of BDI architectures (beliefs, desires and intentions, [137, 138]). The generation of intentions depends on the perceptions (beliefs) and desires of the considered character.

Finally, when a character tries to perform an intended action, its effect is determined by the physical world. For example, if the character pushes an object with a certain energy, a physical action of the world determines whether and how much the object will move.

The narrative knowledge is also represented as a set of rules describing narrative goals, constraints and objective curves. Narrative goals define the criteria that must be satisfied in finished stories (e.g., the story is finished only if there are no humans in the dungeon). Constraints define the criteria that must be satisfied in all acceptable stories, whether or not they are finished (e.g., if the distance between any two humans is less than a threshold then the story is acceptable). *StellA* uses constraints to discard alternative story plots that do not meet the defined constraints. Objective curves express the desired evolution of certain properties of the story (e.g., the richness of a story must increase monotonously). *StellA* uses objective curves both to filter out alternative story plots that do not satisfy the defined criteria and to prefer among different alternative courses of action. Each possible narrative branch is compared with the objective curves. If the difference between the narrative branch and the objective curve is larger than the defined threshold, the narrative branch is discarded. When more than one narrative branch satisfy the defined objective curves, *StellA* prefers (i.e., tries first) the one that best satisfies it.

In addition to the story simulation rules and the narrative knowledge, *StellA* possess a set of rules used in text generation (e.g., if the kind of entity is *knight* then print "the knight").

Contrarily to the *Novel Writer*, the rules used by the *StellA* story generation system (i.e., character control rules, narrative knowledge rules, and text rendering rules) do not change themselves. Thus, the exploration of the space of possible stories does not change during the creative process. This means that *StellA* manifests exploratory creativity.

The *Virtual Storyteller* [25] is a multiagent story generation system. Each story character is an independent agent implemented in JADE (JAVA Agent Development Framework) and JESS (JAVA Expert Systems Shell). In addition to the story characters, the *Virtual Storyteller* also comprises a special agent, the director. The behavior of both the story characters and the director is governed by rules, which lead the agents to achieve their goals.

The director possesses narrative knowledge about global plot structure in the fairy tale domain. The director may control the behavior of the story characters by changing the environment (e.g., adding new objects and characters), giving the characters the goals they should pursue, and disallowing a character to perform an action (before acting, the characters ask the director permission to perform an action). The director cannot force a character to perform an action.

The story plot consists of the events generated by the story characters and controlled by the director. The story plot is then converted into a narrative by the narrator, using textual templates with open slots for words or phrases that express variable information. The generation of natural language output is outside the scope of this article.

When the Story Director creates a new story object or character, or when it gives new goals to existing characters, the emerging interactions between story characters and their actions in the physical story world change. Given that, in character centric or simulation-based story generation systems, the generated stories are the result of the autonomous interactions of their characters, if the space of possible autonomous interactions change, the space of possible stories also change. Thus the *Virtual Storyteller* exhibits transformational creativity.

# 6 Planning-based approaches

Planning-based approaches to automatic story generation consist of modifications of usual AI planning approaches. The basic idea consists of using a planning system to create the sequence of events that make up the story content. However, there is an important difference between planning in general and planning-based story generation.

Planning in general seeks to find the best possible sequence of steps that transforms a given state of the world into a state that satisfies the specified goals, whereas planning-based story generation must cope with the possibly conflicting goals of the story characters and the goals the author intends for the narrative (often called author goals). Thus planning-based story generation systems must be capable of achieving the author's goals while generating plausible, coherent character behaviors at the service of their own goals.

There are different classes of approaches to planning-based story generation, several of them concerning the integration of author goals with the goals of the individual characters. This section presents examples of state-space planning algorithms (i.e., total order planning) and plan space planning algorithms (i.e., partial order planning), the two most representative groups of planning-based approaches used for automatic story generation. This section also presents two hybrid approaches combining planning with other technologies used in story generation. One of them integrates state-space planning and implicit knowledge about the general organization of multi-episode stories (as in the schema-based story generation). The second integrates plan space planning and story fragments (as, for instance, in case-based story generation). The distributed approach to planning-based story generation is also exemplified. Finally, the section ends with a proposal regarding the automatic definition of new domain actions that may then be input to the planning algorithm, and a proposal concerned with stories with several storylines. These two proposals are presented at the end of this section because they are independent of the kind of used planning approach (state-space or plan-space; and monolithic or distributed).

All planning-based approaches require the specification of the actions that can be used to define the story events, the initial state of the story world, represented as a set of propositions, and the goal to be achieved, represented as a proposition. In planning-based approaches, the algorithm tries to find a sequence of actions that can lead from the given initial state of the world to a state in which the specified goal is satisfied. The two fundamental classes of AI planning algorithms are state space planning algorithms and plan space planning algorithms. In state-space planning, each of the search states represents a state of the world; and the arcs connecting two search states represent actions performed in the world. The search process can be made in a forward fashion (i.e., from the initial state to a goal state) or in a backward fashion (i.e., from a goal state to the initial state) or combining both. One of the most successful state space planners is the FF (Fast Forward) algorithm [139].

In plan-space planning, the algorithm starts with an empty plan containing only the initial step, which has an empty precondition and the initial state as its effects, and the final step, which has the set of goals as preconditions, and an empty effect. Thus, implicitly, the empty plan represents the initial and goal states. The algorithm then performs a set of changes in the plan until all planning problems are solved. In the simplest case, planning problems are open preconditions, that is, preconditions of actions included in the plan that are not yet satisfied. The planning problems of the empty plan are exactly the specified goals. In plan-space planning, each search state is a plan. The arcs connecting two search states (two plans) represent plan modification operations that originate a different plan by fixing a problem in the received plan. One of the best-known plan space planners is the partial order planner called UCPOP (Partial Order Planner with Conditional effects and Universal quantification) [140].

In terms of the generated plans, state space planning generates totally ordered plans, whereas plan space planning may generate partial order plans, that is, plans that may contain unordered steps.

In general, a planning-based story generation system explores the space of possible stories thus manifesting exploratory creativity. However, as it will be seen later, there are exceptions. Some of the reviewed planning-based story generation systems exhibit transformational creativity (e.g., [35, 36]).

Some story generation systems (e.g. [29, 31, 42, 141]) are based on state space planning, that is, total order planning. Others are based on partial order planning (e.g., [26-28, 30, 35]).

Porteous and Cavazza [29] developed a state-space approach to automatic story generation coping with the specification of both author goals and character goals. The system enables the user to specify their goals through a set of desired facts that must be true in some state of the story to be generated and a set of temporal constraints among them. The system reasons with the specified temporal relationships to produce the temporal sequence of the facts specified by the user, here called temporal goals. Then, the system uses the FF planning algorithm [139] to generate the sequence of story character actions that transforms the initial state into a new state ($S_1$) in which the earliest temporal goal is satisfied. Then, it calls the FF planner again to generate the sequence of story events leading from state $S_1$ to a new state $S_2$ satisfying the second temporal goal. This process continues until the last temporal goal is satisfied. The plot is the concatenation of the several action sequences produced by the FF planning algorithm for all the specified temporal goals. In the described approach, the author goals are the temporal goals specified by the user. Each intermediate action sequence reflects the coherent behavior of the story characters.

Although the intermediate action sequence, in the system described in [29], is planned using the actions of the story characters, the goals they achieve may conflict with those of the characters. Therefore, this approach generates plausible character behaviors only if the user carefully thinks of their temporal goals so that they do not entail implausible character behaviors.

*Glaive* [22] is a story generation system using a state-space search algorithm that performs intentional planning and also supports conflict. Intentional planning represents an important advance in simultaneously coping with author goals and character goals. The planning algorithm seeks to achieve the specified author goals. However, each action that it includes in the final plan must be performed by one of the story characters, at the service of their intentions, and must be consented by all characters specified in the action representation. For this, the algorithm builds intentional paths of story characters. An intentional path for character *C* and goal *G* is a sequence of actions performed by *C*, which achieve *G*. Each action in an intentional path, but the last one, creates at least one precondition for the following action. *C* must intend *G* since before the first action of the intentional path until the last one. *Glaive* algorithm does not output a plan until each of its actions is justified by an intentional path. An intentional path may justify any action in the final plan even if not all the actions in the intentional path are executed in the final plan. The non-executed actions of an intentional path, which the algorithm used to justify an action in the final plan, are included in the final plan but marked as non-executed. This way, the generated stories may use even non-executed steps to better justify all story steps with the intentions of some character.

A plan containing the actions of an intentional path of some character may be thwarted, for instance, when it does not include all the author goals and, the planning algorithm finds a different plan to achieve them. When this is the case, the non-executed actions of the intentional path reflect conflict between characters given that they were not executed because, in the chosen branch of the search space, the actions of some other character were. Thus, in addition to justifying steps that were executed in the story, the non-executed actions may be used in the story to explicitly show conflict.

In *Glaive* [22], the compromise between author and character goals does not manifest itself only because character intentions are used to justify all steps in the story. When choosing the best course of action to achieve the author's goals during the search process, *Glaive* evaluates each state using the larger of two heuristic scores: the estimated distance to a state that satisfies all author's goals, and estimated cost of having all of the plan's steps justified by character goals.

In *Glaive* [22], actions may fail not because they are actually tried. An action is not executed because, although it belongs to an intentional path of a story character, the algorithm chooses a different course of action thus the action is not even attempted. However, it may be desirable for some narratives that characters may try to execute actions that fail. Authors may intentionally design failed action execution attempts for example to build tension, to prolong goal achievement efforts, or to highlight to the reader the disparities of knowledge and ability among characters. Thorne and Young [43] proposed HEADSPACE, a story generation system using a forward chaining state space planning algorithm that generates plans in which when a character actually attempts to execute an action, it may fail because the character erroneously think the action preconditions are true in the world, whereas, in effect, they are not.

HEADSPACE [43] represents the author's goals and the goals of all story characters. The representation of each world state includes the set of propositions actually describing the world and the belief sets of all story characters, which may be correct or incorrect. Story characters may be uncertain about the truth value of any number of propositions. Each ground action includes the character *C* that executes it, the actual and the believed (by *C*) sets of preconditions, and the actual and believed (by *C*) sets of effects. Character *C* may execute an action, in the world state w, if the set of propositions *C* believes to be preconditions of the action are included in the beliefs *C* holds in world state w. If the action's actual preconditions hold in the state in which the action is attempted, the action succeeds. If the action's actual preconditions do not all hold, the action fails. When *C* successfully executes an action, the actual conditions of the world, the beliefs of *C* and the beliefs of other characters change accordingly. When *C* fails their attempt to execute an action, the action's effects do not take place. However, the beliefs of *C* change. *C* becomes uncertain about the truth-value of the propositions *C* believes to be preconditions of the action.

As can be seen later in this section, about distributed planning approaches, the IMPRACTICAL system [142] also uses defective models of the beliefs of individual story characters. However, instead of exploring the possibility of actions that fail, the proposed algorithm uses this possibility to allow a character *A* to induce wrong beliefs in another character *B*, leading *B* to act in ways desirable to *A*.

Often, a narrative ought to include a protagonist with their own goals and capabilities and an antagonist that tries to thwart the plans of the protagonist. Intent based planning algorithms (e.g., [22]) may generate stories with the roles of protagonist and antagonist, provided that the goals and capabilities of the characters playing such roles are given to the system, in addition to the goals of the author, which favor the protagonist or the antagonist as desired. PROVANT (Protagonist vs. Antagonist) [31] constitutes a different approach in which, not only the behavior but also the goals of the antagonist are automatically generated by the system.

PROVANT [31] constitutes a hybrid approach integrating planning and general schema-based narrative knowledge, which implicitly defines the author's goal. According to Porteous and her colleagues [31, 42], certain narratives consist of an initial phase of exposition (presentation of the characters, their goals and the initial world state), followed by a series of chains of antagonist interference followed by protagonist actions to recover from the interference. This series terminates when the protagonist eventually achieves their goals. By using this schema, the system chooses a type of narrative in which the protagonist eventually succeeds. The top level procedure of the PROVANT system implicitly uses this knowledge about the structure of the envisioned kind of narratives. The top-level PROVANT algorithm starts with the initial state. While the protagonist has not achieved their goals, PROVANT generates the set of all possible interference episodes, selects one of them, updates the current story state with the results of executing the action sequence of the selected episode, and updates the story with the action sequence of the selected interference episode. When the protagonist achieves their goals, PROVANT returns the generated story. In this process, each interference episode is a sequence of actions resulting of the protagonist planning to achieve their goals and the antagonist counter planning to prevent that from happening.

To generate an interference episode, PROVANT creates the plan for the protagonist, in which some actions are visible to the antagonist. Then, using a goal recognition algorithm [143] with an initial (increasingly longer) subsequence of the visible actions of the protagonist's plan and the set of their possible goals, PROVANT identifies the likely actual goal of the protagonist and corresponding landmarks. The set of landmarks for a certain goal *G* is the set of propositions that must be true in some state of each possible plan that achieves *G*. Thus, the landmarks of a certain goal *G* constitute good candidate points the antagonist may consider attacking if they want to prevent the protagonist from achieving *G*. Thus, an interference is any plan that results in removing preconditions of actions that achieve landmarks of the assumed protagonist's goal. PROVANT uses the algorithm described in [144] to generate the set of all possible interferences for the assumed protagonist's goal. An interference episode is a sequence of steps merging the initial sub plan of the protagonist and the considered interference plan of the antagonist.

Whenever PROVANT generates more than one interference episodes, it selects the one that leaves the protagonist as far as possible from their goal but from which the protagonist may still recover.

With the aim of improving efficiency and story novelty, Farrell and Ware [141] use a general search heuristic, called *Novelty Pruning*, that can be used in any kind of state space planning approach to story generation. Farrell and Ware [141] described a set of experiments designed to evaluate the efficiency of the story generation process and the diversity of the generated stories, which confirm that *Novelty Pruning* improves planning efficiency and moderately improves story novelty.

*Novelty Pruning*, which was taken directly from the *Iterated Width* algorithm [145], consists of discarding all world states of the state space that are not sufficiently different relative to the others. The novelty of a certain world state depends on whether or not the considered state contains propositions that are not contained in any other already existing world state.

The experimental tests compared the results of four planning algorithms, in terms of planning efficiency and story novelty: breadth first search, breadth first search with *Novelty*

*Pruning*, the *Glaive* algorithm [22], and the *Glaive* algorithm with *Novelty Pruning*. The evaluation general results strongly confirmed that *Novelty Pruning* significantly reduces planning times and the number of visited world states.

For assessing the novelty of the generated stories, Farrell and Ware [141] used a domain-independent distance metric to compare two story plans. The more different a story is from the others, the more novel it is. The used metric of the distance between two story plans, proposed in [146], compares the summaries of the two stories using the Jaccard index [147]. A story summary consists of two components: the set of the story important steps (i.e., those with highest causal dependency and impact) and the summaries of the intentional frames of the story plan. The causality of a step is the number of its satisfied preconditions plus the number of its effects that have been used by subsequent steps. An intentional frame is a triplet containing a character intention, the event that creates the intention and the sequence of events that achieve it. Finally, an intentional frame summary consists only of the intention, the step that creates it, and the step that achieves the intention.

The experimental results provide moderate evidence that the *Novelty Pruning* heuristic generates more innovative stories.

Up to this point, this section has been presenting story generation work based on state space planning algorithms. Now, we move to plan space planning approaches to story generation.

Several story generation systems extended partial order planning (i.e., the planning style of plan-space planning) in different ways. For instance, Riedl and Young [28, 41] defined and used the algorithm IPOCL (intent-driven partial order causal link) that extends partial order planning with the aim that the reader perceives the behavior of the characters controlled by the algorithm as intentional. According to the authors, intentionality is one of the more important factors of believability. In addition to causality (which is the focus of traditional planning algorithms), IPOCL is concerned with the intentionality of the planned actions. In a valid plan, all actions must belong to intentional action sequences performed at the service of one of the goals of the character. Intentional action sequences must start with an action that creates the intention of achieving one of the goals of the character and must end with an action that satisfies the selected goal. At each step, IPOCL selects a causality related flaw or an intentionality related flaw and tries to fix it by changing the plan.

Ware and Young [30] defined a similar intent-driven extension of partial order planning. Additionally, the proposed approach enables the generated story to explicitly represent conflict between characters. The presented CPOCL algorithm (Conflict Partial Order Causal Link) extends intent based planning to explicitly capture how characters can thwart one another in pursuit of their goals. A conflict exists when an intended step of one character threatens an intended step of another character and one of these steps is not executed. To allow the story to use information about conflict, CPOCL marks certain steps in a plan as non-executed (the threatening or the threatened steps). This way, CPOCL preserves the conflicting sub plans of all characters without damaging the causal soundness of the overall story.

Bahamón, Barot and Young [26, 27] presented yet another extension of the traditional partial order planning algorithm, called CB-POCL (Choice based Partial Order Causal Link). The subjacent idea is that the reader will infer the personality of the character from the observations of the choices they make along the narrated sequence of events. Whenever a character has a choice of possible actions to achieve a certain goal or sub goal, the proposed algorithm evaluates the alternative sequences of actions that start at that choice point, rates each of them according to the degree to which it reflects the intended personality of the character, and chooses the most consistent one. If there is enough contrast among the possible courses of action, the algorithm presents the most contrasting alternatives to the reader, facilitating their interpretation of the character behavior as consistent with the personality the author intended for the character.

*Suspenser* [148] is a partial order planning-based story generation system. It receives a partial order plan representing a complete story. The purpose of *Suspenser* is to generate a new version of the received story, comprising all of its relevant events, but triggering the specified kind of suspense (high vs. low) from the reader, at the specified point in the story. *Suspenser* evaluates all events in the received story, according to their importance. The importance of an event reflects its causal relationships with the goals of the protagonist and its qualitative importance. To estimate the causal relationship of an event with the goals of the protagonist, *Suspenser* counts the number of earlier events that set preconditions for the considered event and the number of effects of the event that set preconditions used by future events in the story. The qualitative importance of an event depends on whether it is the opening act of the story, its closing act, or an event that achieves at least one of the goals of the protagonist. After evaluating all events of the original story, *Suspenser* creates a complete partial order plan containing only the most important events that achieve all the protagonist's goals. *Suspenser* then changes this complete plan to achieve the desired kind of suspense (high or low) at the specified point in the story: it adds events from the original story that help raise (or lower) the reader's suspense level before the specified point, and postpones the presentation of events with low (or high) suspense potential until after that point. In this process, the suspense potential of an event increases with the number of its effects that negate one of the protagonist's goals, decreases with the number of its effects that unify with one of the protagonist's goals, and varies inversely with the estimated distance between the considered effect and the considered goal.

Riedl and Young [35] proposed the *Fabulist* system, which is yet another approach that extends partial order planning. *Fabulist* [35] may overcome constraints defined by the initial state of the world or by the preconditions of the actions available to the story characters. The main idea is twofold. First, some of the propositions describing the initial world have unknown truth values. Second, some of the action preconditions are interpreted as soft constraints, that is, constraints that may be violated. When necessary, *Fabulist* assumes that propositions with unspecified truth value are true or false as convenient. If necessary, the system may also decide to disregard soft preconditions. Section 9.3 shows in more detail that these two possibilities extend the space of possible stories therefore, *Fabulist* [35] exhibits transformational creativity.

In the following paragraphs, this section addresses hybrid story generation systems integrating a planning-based approach with the use of good quality story fragments, as it is the case in case-based story generation.

Riedl, León and Sugandh [39, 64] proposed VB-POCL (*Vignette* Based Partial Order Causal Link), another extension of the original partial order planning algorithm. VB-POCL extends partial order planning with plot fragments, called *vignettes*. Each *vignette* is also a partial plan representing a good plot fragment. When VB-POCL selects an open condition to fix, it can add a temporal constraint between actions already in the plan, it can add a new action to the plan, or it can retrieve a *vignette* from a case base of plot fragments and use it in the current plan. Both the plans being generated and the plot fragments are partial order plans. When a plot fragment is used by the planner to fix an open condition flaw, the plot fragment is not sequenced in the plan being generated. Instead, the two partial order plans are merged, giving rise to a new partial order plan. In the merging process, it may be necessary to create and fix new open condition flaws. The process of incrementally merging the plot fragment and the current plan being generated avoids unnecessary repetitions of events and enables the earlier detection of problems that prevent the plot fragment from being used.

VB-POCL plot fragments do not reference specific characters, objects, or entities so that a planner can fit the *vignette* into new story contexts by making appropriate assignments. For this to be possible, VB-POCL plot fragments specify a set of constraints on the characters and objects with which it may be instantiated.

Some systems that use planning algorithms for automatically generating the story plot (e.g., [21, 24, 39, 64]) share an idea with a class of case-based reasoning systems whose cases represent plot fragments assumed to represent components of good stories (e.g., [67]). The use of plot fragments representing good plot event sequences is an attempt to generate good quality stories in spite of the absence of reliable computational models of good stories.

The planning algorithm of the TALE-SPIN system [24] uses plot fragments dynamically generated by planboxes. A planbox is a procedure that, when executed, generates the sequence of actions to be performed by a story character in the circumstances in which it is used. In TALE-SPIN, a planbox is associated with a goal, a set of pre-conditions and a set of post conditions. When a planbox is successfully used, it achieves the goal to which it is associated. If the planbox's post conditions are true after the execution, it means that the desired effect was successfully achieved. Otherwise (e.g., the story character was wrong about the truth-value of a pre-condition leading the planbox not to produce the desired effect), the system must re plan to repair the failure. When the pre-conditions of a planbox are not totally satisfied, the planner may have to resort to other planboxes to achieve some or all the preconditions of the initial planbox that are not true.

TALE-SPIN is the only reviewed system whose plot fragments are dynamically generated by the execution of procedures (planboxes). In all other systems that use plot fragments, the plot fragments are sequences of events specified *a priori*.

UNIVERSE [21] also uses planning with plot fragments. In addition to being pre specified sequences of events, the plot fragments used in UNIVERSE represent sequences of story events that may be played by different characters, whereas the plot fragments dynamically generated in TALE-SPIN represent sequences of actions of a single character. In the case of plot fragments comprising the actions of several characters, the goal to be achieved by the plot fragment is an author goal, not the goal of any of the story characters. A plot fragment used in UNIVERSE is associated with the goals it was designed to achieve, a list of character roles, a list of pre-conditions and a list of effects.

In TALE-SPIN, the planner generates the story-plot by sequencing plot fragments. In UNIVERSE, in addition to sequencing plot-fragments, the planner performs a kind of hierarchical planning because a plot fragment may contain both low-level actions and other goals that also require planning.

Planning approaches to automatic story generation may also be organized according to a different criterion: monolithic approaches and distributed approaches. In monolithic approaches, a single planning algorithm creates plans that control all the story characters. In distributed approaches, each story character is associated with an individual instance of a planning algorithm. The majority of work in planning-based story generation has adopted a monolithic approach (e.g., [22, 26, 28-31]). Consequently, the current section on planning-based story generation has also focused on monolithic approaches. Next, we present a description of distributed planning-based story generation.

IMPRACTICAL (Intentional Multi-agent Planning with Relevant ACTions) [142, 149] is a distributed planning-based story generation system comprising a set of local planners, one for each story character, and a global narrative planner, responsible for actually creating the story. When, in its search process, the global narrative planner considers a world state to continue the story, it asks all cooperating local character planners for all the actions they find adequate, given their intentions. Cooperating characters are those that share the intention considered by the global narrative planner at each iteration of the process. Each character planner reasons about the received state and the selected intention, and returns the actions of all its plans that might achieve the intention. Given that the global narrative planner must pass the considered world state to the local character planners, the narrative planner must be a state space planning algorithm. Moreover, the global narrative planner must apply the actions returned by the local character planners to the considered world state, thus it must be a forward search algorithm. Given that the actions returned by a local character planner to the global narrative planner must belong to the intentional plans of the character, the local character planners must perform intent-based planning.

The global narrative planner uses three different strategies to request the set of relevant actions from the local character planners. In the first strategy, the basic strategy, for each open intention of the global narrative planner (in the considered state), it requests the relevant actions from local character planners that share the selected intention. The global narrative planner uses only those actions whose preconditions are satisfied in the considered world state.

The main idea underlying the second strategy consists of taking into account the effects of actions that will be performed in the future by cooperating characters. To account for these potential effects, the global narrative planner repeats the basic strategy until it does not return any action. In each iteration, the global narrative planner computes an updated virtual state containing the add effects of the basic actions returned in the previous iteration. In the next iteration, instead of passing the current world state to the individual planners, it passes them the updated virtual current state so that the individual planners may return new relevant actions taking into account the effects of future actions of the other characters.

Finally, the third strategy relies on persuasion (intention induction). For each intention, it starts by requesting the set of relevant actions from the character planners with that intention (cooperating agents), as in the basic strategy. If the local character planners do not return a set of relevant actions, the global narrative planner asks the cooperating agents for actions that can create the intention in other characters that do not already share it – the intention inducing actions (or persuasion actions). Next, it includes persuaded agents in the set of cooperating agents and repeats the process until a non-empty set of relevant actions is returned.

In the system described in [149], all character planners have perfect knowledge of the world at all times. Later, Teutenberg and Porteous [142] extended IMPRACTICAL by allowing each character to have their own possibly defective set of beliefs about the world. Given that story characters may have wrong beliefs about the world, it is possible to create a story in which a character may perform actions because of their wrong beliefs. This makes it possible to create stories with deception, which might be important to improve the story interestingness. Deception arises, for instance, if a story character *A* performs an (deceptive) action whose effects create wrong beliefs in another character *B*, so that *B* will perform an action, allowed by their wrong beliefs, whose effects satisfy *A*'s intentions.

The new version of IMPRACTICAL [142] uses a set of rules that determine the belief changes of the story characters, for example, "*If an action changes an object's location, all agents at its previous and new locations perceive predicates relating to this object*", "*The actor and all participants of an action perceive its preconditions and effects*", and "*If an action is public, all agents perceive its preconditions and effects*". The belief change rules, as the above ones, require to augment the usual action representation with the set of participant characters, the specification of whether the action is private or public and the location in which it occurs (if any).

While the other approaches presented in this chapter are complete planning algorithms, the next one is concerned with a specific part of a planning-based approach, namely the definition of new domain actions. Additionally, the following proposal may be used, in principle, in any planning-based story generation system.

With the purpose of increasing the number of the stories that can possibly be generated by the same planning-based system, Porteous and colleagues [36] present a system, named ANTON, that takes a planning domain model, identifies missing actions, and generates the new domain model, which augments the original one with the new previously missing actions. ANTON identifies two kinds of missing actions, both of which produce effects that are required by other actions but no other action produces: actions that produce effects contrary to those produced by other existing actions (i.e., antonymic actions); and actions that produce or delete relevant conditions (i.e., property gaining and property loosing actions). In addition to identifying and defining required actions, ANTON generates human readable names for them. The modified planning domain model, augmented with the newly defined actions, may be given to both state space planners and plan space planners, or even to those hybrid planning algorithms with plan fragments that can still use individual actions (e.g., VB-POCL [39, 64]).

The definition of a new domain action requires the action name, the action arguments, the action preconditions and its effects. To generate the name of an antonymic action, ANTON uses the WordNet *synsets* associated with the name of the original action to widen the scope of the subsequent search process. Then, it searches other online linguistic resources to look for antonyms of the words in the *synsets* and to rate them. Finally, it selects the antonym with the highest total rating. For antonymic actions of multi word named actions (e.g., love spell), ANTON creates a pair for each word in the name comprised of the word itself and its antonym. Then, it generates combinations of a word in each pair, except the original naming expression, (e.g., hate-spell, love unspell, and hate unspell) and choses the most frequent combination in an online n gram language model. The parameters of antonymic actions (e.g., divorce) are the same as those of the original actions (e.g., marry). With respect to preconditions and effects, the basic ideas are as follows. The set of conditions achieved by the original action (e.g., the two characters are married) are used as the preconditions of the antonymic action. The positive effects of the antonymic action are the negative effects of the original action (e.g., the characters are single), and the negative effects of the antonymic action are the positive effects of the original action (e.g., the characters are married).

Property gaining actions for property p (e.g., beautiful) are called "become"-p (e.g., become beautiful). The name of a property loosing action for property p is "become"-antonym(p) (e.g., become ugly). Property gaining and loosing actions have a single argument: any object of the type that may exhibit the considered property (e.g., ?x of type princess). The precondition of a property gaining action is the negation of the property to be gained, applied to the action parameter (e.g., *not(beautiful(?x))*), and the effect is the property to be gained, applied to the action parameter (e.g., *beautiful(?x)*). Finally, the precondition of a property loosing action is the property to be lost, and the effect is the negated property to be lost.

Given that the possibility of using new actions augments the space of possible stories, the approach proposed by Porteous and colleagues [36] manifests transformational creativity. It transforms the space of possible stories, as implicitly defined by the initially given set of possible actions, into a new, larger space of possible stories, implicitly defined by the extended set of possible actions.

Although interleaving multiple storylines may contribute to maintain the attention of the reader, automatic story generation has not focused on that possibility. This section, including the part on distributed approaches to planning-based story generation, has also been concerned with stories with a single storyline. Even the systems described in [29] and in [31], which tackle multi-episodic stories, are not explicitly concerned with stories with multiple storylines, each one with multiple fragments. Porteous, Charles and Cavazza [42] defined a hybrid approach, using both implicit schema-based story generation and planning-based story generation, that extends previous work on multi-episode stories [29] to handle stories with multiple storylines, which in turn have multiple segments. The different individual storylines are implicitly defined by the goals the author (i.e., the system's user) provides to the system. The different segments within the same storyline are also implicitly defined by the author sub goals for that storyline.

The system described in [42] tackles the interleaving of the several storyline segments using schema like narrative knowledge. The planning algorithm has the responsibility of generating the sequence of steps of each of the storyline segments. Although the authors used a monolithic state-space forward planner for each segment, they could have used about any other planning algorithm.

Porteous, Charles and Cavazza [42] structured each storyline according to a relatively general story schema, which is used also by Porteous and Lindsay [31]. Each individual storyline is a sequence of narrative segments, which may belong to several classes of narrative segments: introduction, obstruction, resolution and exposition. A storyline requires some initial introduction to the protagonist and their goals, followed by one or multiple obstruction segments in which the protagonist tries to recover from encountered obstacles to their goals. The storyline may end with a resolution segment. Individual storylines can, at any stage, include phases of exposition providing additional information to the audience (e.g. plot re-caps or further insight into characters).

Multiple storyline narratives interleave segments from a number of individual storylines, which are not independent because characters can appear in different storylines (possibly, in different roles). Thus, the consequences of events in one storyline can impact the unfolding of the narrative in other storylines. Adjacent narrative segments in a story must belong to different storylines.

The system's top level procedure implicitly encodes the narrative knowledge captured by the described schema. In each step of its main loop, it selects the next storyline to handle, that is, the next author goal. Within the selected storyline, it selects the sub goal to satisfy, which implicitly defines the narrative segment to be generated. Then the system generates the segment, using its planning algorithm. The next storyline to work on is the one that is different from the last one, and with more narrative segments (sub goals) still to be interleaved in the global story. When the length of a given segment is not enough, according to the given specification, the system generates additional exposition segments.

# 7 Evolutionary approaches

Evolutionary story generation uses evolutionary algorithms [102] to generate stories. In its simplest form, evolutionary story generation starts with an initial population of possible stories. These initial stories can for instance be randomly generated, previously provided to the system, or extracted from a dataset of existing stories. At each step of the evolution process, the story generation system evaluates the current population of possible stories. Then, it selects a subset of the existing population, based on their evaluation, trying to preserve good stories and story diversity. The evaluation process can rely on the desired set of criteria, for instance novelty and interestingness. Finally, the evolutionary process generates new stories from the selected subset. The newly generated stories may be the result of mutating or combining previous ones. Then, the evolutionary process repeats until some specified condition holds (e.g., a good enough story is generated or a predefined number of evolutionary steps has been reached).

When, in an evolutionary step, a new story results of the combination of two previous ones we face combinational creativity. When a new story results of a modification of a previous one, the modification consists of replacing a previous story component (e.g., a sentence) with a new one. The new component may express an entirely new concept, as long as it makes sense according to commonsense knowledge. That is, mutating a previous story also involves the combination of a previous story with a different concept. Thus, a creative story generation process that generates new stories through mutation may exhibit combinational creativity.

Evolutionary approaches have been used in automatic story generation but often to address more specific sub problems of the main generation process. For instance, evolutionary algorithms have been used for the invention of new concepts [150], which could then be used in generated stories. In a more related problem, Oliveira [151] used evolutionary algorithms to generate poems, but poetry is subject to more specific and restricted formal constraints, amenable to be expressed as fitness functions, which is not the case with stories.

The work of Baydin and colleagues [34, 152] on memetic evolution (in the sense of Dawkins [153]) is worth describing here given its relevance to story generation. First, the evolved new ideas (as memes) can be used in stories as fictional ideas. Second, the proposed approach to meme evolution works with graph representations, which is also the case with the script-like knowledge represented as plot graphs. Especially this second argument shows the obvious resemblance of meme evolution and story evolution. In the concluding remarks of their paper [34], the authors acknowledge the possibility of using a similar approach to story generation.

Baydin, López de Mántaras and Ontañón [34] presented an evolutionary approach to create meaningful analogues of memes provided as input to the system. Both the input meme (e.g., *humans are animals capable of thinking that desire to learn*) and the generated analogues are represented as semantic networks. In the described approach, the evolutionary algorithm uses several variation operators defined to work with graphs: two crossover operators (*subgraph* *crossover* and *graph merging crossover*) and six mutation operators (which insert, delete and replace nodes and relations in the input graphs). All of these operators have in common an important property: the changes that they operate on the received graphs to generate the next generation of individuals are constrained by commonsense knowledge represented in two freely accessible existing commonsense knowledge bases, the WordNet [154] and the ConceptNet [155]. For instance, the mutation operators do not just add any relation between two selected concepts; they can only add a relation between two concepts if, according to the commonsense knowledge, the relation applies to the concepts. And the *subgraph crossover* operator will only replace a certain subgraph (from one of the parents) with another subgraph (from the other parent) if, according to the commonsense knowledge, it is acceptable to replace one with the other. This contributes to the meaningfulness of the generated results.

The first stage of the proposed algorithm is the automatic generation of the initial population of potential solutions. Each individual of the initial population is initially set to a single node randomly picked from the commonsense knowledge base. Then, for a node randomly picked from each network, the algorithm adds it a relationship (and the other node of the relationship) retrieved from a commonsense knowledge base. A relationship may be added only if the plausibility of the added network fragment is above a certain threshold. If more than one network fragment can be added, one of them is randomly picked. The process continues until a certain criteria (e.g., network size) is satisfied. At each stage of the evolutionary process, the fitness of each individual of the population is computed.

The fitness of an individual is the degree to which it is structurally analogous to the user provided input network. The structural similarity of two graphs is computed by the SME algorithm (Structure Mapping Engine) by Falkenhainer and colleagues [156], adapted for graphs (originally, SME works with predicate logic representations).

To select the individuals of the population to be used in the next evolution step, the described approach employs tournament selection [157], because it contributes to preserve population diversity (given that less fit individuals have some chance of being selected), which is beneficial to overcome local maxima.

Although the proposed approach has been designed to generate analogues of graphs representing declarative information about a set of entities (whereas the nature of stories is mainly procedural), the approach could be used to generate declarative descriptions of the story entities and their relations; and it could also be adapted to operate with graphs representing procedural information about story events.

McIntyre and Lapata [33] proposed an approach in which, an evolutionary algorithm, using both mutation and single point crossover, takes an initial population of stories, represented as a graph (plot graph) and generates a new story. The fitness function scores the individual stories according to coherence criteria combined with additional constraints potentially specified by the user.

In the system described in [33], before the system can be used to generate stories, it processes a large corpus of existing stories and builds a set of so called entity narrative schemas, for all entities existing in the corpus. An entity narrative schema is a partially ordered graph representing the events participated by the entity (and possibly by other entities that also participated in the same events). The extraction of the narrative schemas of all entities in the available story corpus is described in detail in section 8. This section focuses on the evolutionary process of story generation.

When the user provides a sentence from which to build a new story (e.g., *the princess loves the prince*), the system selects and merges the entity schemas of all the entities referred in the sentence (e.g., princess and prince) that satisfy the constraints potentially provided by the user (e.g., maximum number of events). The plot graph, consisting of the sub graph of the merged entity narrative schemas that starts with the user provided sentence, represents a set of possible stories, which constitute the initial population of the evolutionary algorithm.

The evolutionary algorithm performs both single point crossover and mutation. Single point crossover consists of generating a story containing the sentences of one of the parents until a certain point and the sentences of the other parent from the selected point onwards.

Mutation consists of changing the order of sentences or replacing an expression of a selected part of speech with another one with the same part of speech. Since verbs have structural importance, when replacing a verb, the system must also pay attention to its arguments. If the verb to be mutated is a matrix verb or the verb of a simple clause (a clause that does not contain subordinate clauses), the whole sentence is replaced with a new one (e.g., the simple clause "*The prince married the princess in the castle*" is replaced with "*The prince holds the princess*"). If the verb to be mutated is the head of a sub clause, the entire sub clause is replaced with a new sentence headed by a verb that has been found to co-occur with the corresponding matrix verb (e.g., in the sentence "*The prince knows the princess loves children*", the sub clause "*The princess loves children*" is replaced with "*The princess escaped the dragon*") because the verb to escape was found to frequently occur with the matrix verb to know.

The evolutionary algorithm uses a fitness proportionate strategy (also called *roulette-wheel* strategy) as the selection approach. This decision, selecting individuals with a probability proportional to their fitness, ensures that even less fit individuals have a certain (although small) chance of being selected and thus increasing diversity, which may enable the algorithm to overcome local maxima.

Andrés Gómez de Silva Garza, Erik Cambria and Rafael Pérez y Pérez [158, 159] describe a hybrid approach to story generation integrating an evolutionary algorithm and a theory based approach. In the proposed approach, the initial population of the evolutionary algorithm is retrieved from the system's case base.

In the hybrid approach, any story (either from the initial population or a newly generated one) is evaluated against a general domain theory, general commonsense knowledge (e.g., stories should not contain repeated events), and general principles regarding good stories (e.g., the tension should increase until a certain point near the end). Instead of rejecting stories that do not satisfy the applicable criteria, such stories are assigned low fitness values. This way, although less likely, stories that do not meet the desired criteria still have a chance of being selected to generate the new population, increasing the population diversity, which may lead to the generation of better quality final results.

# 8 Learning based approaches

Two major problems affect the knowledge based approaches to story generation: they require a considerable amount of manual intervention to create the required knowledge, and the knowledge based system is limited to the engineered knowledge. Learning based approaches appear as a promise to overcome these two problems.

Learning-based approaches to story generation are organized into two different classes. In one of the classes, a learning algorithm is used to learn the explicit formal representation of narrative knowledge (e.g., story schemas or story cases) from a set of stories of a given domain (e.g., [33, 53, 55-57, 115, 132]). The learned narrative knowledge representations may then be used to generate stories, using schemas [115], story case bases [132] or evolutionary algorithms [33].

In the second class of learning-based story generation approaches, the learning algorithm learns implicit knowledge about possible event sequences (e.g., [58, 59, 62]), for instance, in the form of a probability distribution of the occurrence of a story event given the previous occurrence of other events. In this class of approach, after the learning algorithm has been trained, it is given the first story event. After that, story generation consists of using the algorithm to generate the next event to include in the story, feeding the next event again into the algorithm and, again, using it to generate the next one, and so on, until, some defined criterion is satisfied (e.g., a fixed number of events have been included in the story).

Most of the described learning based approaches use publically available knowledge bases with linguistic knowledge, such as WordNet [154] and ConceptNet [155] and natural language processing software such as the Stanford CoreNLP Suite [60] and the Apache openNLP.

Chambers and Jurafsky [53] propose an approach to learn narrative event chains from a textual corpus of raw news. In the proposed approach, a narrative event chain is a partially ordered set of events participated by the same protagonist. The authors used the Apache openNLP co reference resolution engine to solve co referencing problems. To learn a narrative event chain, the approach uses an unsupervised learning algorithm that identifies the events that belong to the narrative event chain. At each step, the training corpus event that maximizes the total mutual information between itself and the events already in a chain is included in the narrative event chain. The total mutual information between a set of events and a target event is the sum of the pointwise mutual information between each event in the chain and the target event.

Once the set of events that belong to a specific narrative event chain has been built, the approach proposed by Chambers and Jurafsky [53] identifies temporal order relations among the events. The process of identifying the temporal order between two events comprises two stages. First, the proposed approach uses supervised learning to assign temporal related labels to the two events (e.g., tense, grammatical aspect, and aspectual class) using a set of grammatical features (e.g., neighboring part of speech tags, neighboring auxiliaries and modals, and WordNet *synsets*). Second, the assigned labels together with additional features (e.g., syntactic dominance relations between the two events, and whether the events occur in the same or different sentences) are used to train a supervised learning algorithm that learns to classify the temporal order of two events. In both stages of the temporal order classification, the authors used SVMs (Support Vector Machines, [160]).

The confidence of the temporal order assigned to two generic events *A* and *B* results of comparing the number of times that the assigned order was found with the number of times that the inverse temporal relation was observed.

To generate script like structures, Chambers and Jurafsky [53] first clustered the events in the corpus, using mutual information scores. Than they applied the ordering relations to produce a directed graph.

McIntyre and Lapata [33] proposed an approach that generates story plots from plot graphs automatically extracted from a training corpus of existing stories. To learn a plot graph, the system builds a knowledge structure, called an entity schema, for each entity in the training corpus. The entity schema represents all events participated by the entity, called the focal entity of the schema, together with other entities possibly involved in the same events. As it was the case with Chambers and Jurafsky [53], the algorithm uses the Apache OpenNLP co reference resolution engine to solve co referencing problems.

The algorithm proposed by McIntyre and Lapata [33] incrementally builds entity schemas for each entity in the corpus, a story at a time. Each event in the first story, participated by a certain entity, is a node in the entity schema of that entity. The entity schema does not contain events not related with the schema's focal entity. An arc from a node to the next represents the temporal ordering of the two events, and their mutual information. When the algorithm processes another story, it adds new nodes to each entity schema or merges events from the current story with nodes already contained in the entity schema. If two events share the same verb with the focal entity playing the same role, and the remaining corresponding arguments, if any, are semantically similar, the algorithm merges the two events in the same node. Semantic similarity of two entities is computed from the WordNet and considering the length of the shortest path between the entities.

When the system starts generating a story with a set of entities (e.g., prince and princess), it merges the entity schemas of all the involved entities. The resulting knowledge structure is a plot graph, which represents all possible stories with the selected set of entities. As described in section 7, the system proposed by McIntyre and Lapata [33] uses a genetic algorithm over the population of possible stories in a plot graph until one story is generated that fulfills the specified criterion (e.g., the number of events).

Boyang Li and his colleagues (e.g., [44, 54, 114, 115]) developed an approach for story generation that generates stories from script like knowledge structures dynamically learned from a set of crowd sourced stories. The following description focuses on the script learning approach proposed by the authors. Section 3 describes the process of story generation from the learned scripts.

The script learning approach used by Boyang Li and his colleagues starts with a crowd sourcing process carefully designed to facilitate the subsequent automated learning process. Crowd source workers were asked to write temporally linear stories about several social situations. The authors instructed the workers to use proper names for all the characters in the task (character names are provided for all possible roles, such as the cashier in a restaurant), to segment the narrative into single event sentences, to use simple natural language such as only one verb per sentence, and to avoid complexities such as conditionals. These instructions largely facilitate the automated stage of script learning because it avoids some difficulties such as the identification of omitted characters, co referencing solving, and decomposing compound sentences into several events.

The automated script learning algorithm includes three major stages: event identification, learning of event precedence relationships, and learning of event mutual exclusion relationships.

Event identification is the result of an event clustering process based on semantic similarity of the events. Each of the resulting clusters of similar events is taken to represent a single core event. Li and colleagues use the k-Medoids clustering algorithm [161] to identify event clusters, with the constraint that events that belong to the same story do not belong to the same cluster. The algorithm must know *a priori* the number of clusters, k, thus the proposed approach experiments with different values of k, starting with the average story length, with the goal of minimizing intra cluster variance while maximizing the extra-cluster distance. The decision to include a certain event in a cluster depends on the semantics of the events, which requires a method to compare the meanings of two events.

First, Li and colleagues preprocess the narratives to extract the main verb, the main actor, and the most salient non-actor noun, if any. Second, they compute the semantic similarity of each event and other events, using semantic gloss information from WordNet. Finally, they cluster steps in order to identify the core set of events. The first step uses the Stanford parser of the Stanford CoreNLP suite to extract the main verb, the main actor, and the most salient non-actor noun, and a rule-based approach only to help identify the most salient non-actor noun.

The semantic similarity of two events is the weighted sum of the semantic similarity of the verbs, the semantic similarity of the nouns, and difference of the event locations. The similarity of two corresponding words is the cosine distance of the vectors representing the most adequate *synsets* of the two words in the WordNet, in which, the most appropriate *synset* of each word was selected using a word sense disambiguation technique. Event location is the percentage of the way through a narrative, in which the event occurs. Event location helps disambiguate semantically similar steps that happen at different times.

After the system has identified the core events, it learns precedence constraints between events, but always avoiding loops. Given that the instructions provided to the crowdsource workers were very strict, the stories they wrote are temporally linear event sequences. Thus, identification of precedence relations between events may rely on the order by which the events appear in the story. However, considering that crowd workers generate noisy story examples (e.g., they omit events), this step must be resilient to noise. For all pairs of plot events $e_1$ and $e_2$, the system selects between the two orderings $e_1 \rightarrow e_2$ or $e_2 \rightarrow e_1$ based on statistical frequency. The system only creates precedence relations with statistically significant frequency.

Finally, using a measure of mutual information, the system identifies mutually exclusive events, that is, events that can never co-occur in the same narrative. Li and colleagues consider that two events are mutually exclusive when, from a statistic point of view, the presence of one event predicts the absence of the other, and when their mutual information is greater than a certain threshold, indicating strong interdependence of the two events.

*Scheherazade*, the story generation system created by Li and his colleagues (e.g., [44, 54, 114, 115]), generates stories from the script like knowledge (i.e., the story schema) learned from crowd sourcing. The story generation process is described in detail in section 3.

Unlike previous approaches aimed at learning story schemas (e.g., plot graphs), Valls-Vargas, Zhu and Ontañón [132] propose an approach to case-based story generation in which the system's case base is automatically created by a system that automatically extracts narrative knowledge from a collection of Russian folktales. The proposed approach [132] integrates and extends *Voz* (e.g., [55-57]), a narrative knowledge extraction system, with *Riu* [49], a case-based story generation system. Section 4 fully describes *Riu* [49] and the extension required to *Voz* [132] to enable it to generate *Riu*'s case base. The following paragraphs describe the narrative knowledge extraction approach of *Voz*, which uses a combination of learning algorithms and knowledge based methods (e.g., [55-57]).

*Voz* identifies story events using the Sanford CoreNLP suite. Each story event is represented as a triplet containing the verb lemma, the subject and the object. *Voz* represents the subject and the object of an event by the corresponding co reference groups of mentions in the text. Mentions (expressions referring story entities) are noun phrases identified by the Sanford CoreNLP suite and are encoded, in *Voz*, as feature vectors containing the encoding of the parse tree of the sentence containing the mention, the sub parse tree corresponding to the mention, the actual words of the mention with corresponding part of speech tags, and the list of dependences with references to nodes in the subtree.

*Voz* extracts a large diversity of textual information from a given story. It classifies the story entities as character or non-character, using a case-based approach that associates the mentions in every story of the available corpus with the corresponding class (character / non character) [56]. It assigns roles (e.g., villain, hero) to character-referring mentions using two distinct methods: a case base approach that associates character roles to all character-referring mentions in the available story corpus [56]; and a hill-climbing algorithm whose initial state is a random assignment of character roles to character-referring mentions [55]. At each step of the search process, the algorithm chooses the best assignment and generates its successors by changing, in each successor, the role assignment of a single character. *Voz* considers that the best assignment after 100,000 iterations of the algorithm is the correct role assignment. The evaluation of each role assignment, during the search process, is based on the comparison between the actions actually performed by the concrete story characters and those expected to be performed by characters playing the assigned role in the example stories comprising the training corpus. This requires the relation of all character roles and the actions associated to them in the training corpus. This information is provided as input to *Voz*. The similarity between verbs is computed from the depths of each verb and of their least common subsumer in WordNet.

Finally, *Voz* assigns narrative functions (e.g., departure, return) to the text segments of the story, using a beam search algorithm [57]. At step i of the search process, the algorithm preserves the N best nodes, each of which representing the assignment of narrative functions to the first i text segments. Then, it computes the successor nodes by adding all possible assignments to text segment i+1. When the algorithm reaches the last story segment, it returns the best narrative function assignment.

*Voz* uses the combination of three methods to evaluate each narrative function assignment, except when it evaluates the assignment of the last story segment, in which case, it also uses a fourth method. One method uses k-NN (k Nearest Neighbors, [162]) supervised learning algorithm to predict the assignment of each text segment. The algorithm is trained with a dataset comprised of all pairs of text segments of all stories and their narrative functions (manually provided). Given that the k-NN algorithm works with numeric vectors (representing points in a multidimensional space), the text segments are automatically converted by *Voz* in feature vectors, representing features automatically extracted by *Voz*: the position of the segment relative to the beginning of the story, the character roles mentioned in the segment, and the 10 most relevant verbs of the segment, which are chosen using feature selection algorithms. The second method is a Markov model [100] that predicts the occurrence of a narrative function given the previous occurrence of another narrative function. The third method is a set of rules that constrain the sequence of narrative functions (e.g., the return function may happen only after the departure function, and the first narrative function must be one of the introductory functions, a villainy or a lack).

Narrative function assignments including the last text segment are additionally evaluated by a fourth method which predicts the likelihood of each complete sequence of narrative functions.

The two last learning algorithms are trained with the set of all sequences of text segments in all stories in the training corpus.

The reminder of this section describes learning based approaches to story generation in which the narrative knowledge used in the generation is not explicitly represented. Instead, in the following approaches, the required narrative knowledge is only implicitly learned. All the following approaches use recurrent neural networks (RNN, e.g., [163]), which are widely used to learn sequential patterns. According to Roemmele [164], recurrent neural networks are a promising machine learning framework for language generation tasks. In natural language processing tasks, RNNs are trained on sequences of text to model the likelihood of the next sequence unit (often a character or word) given the sequence up to that point.

In the following description, we have assumed that the basic unit of an input sequence is the word. However, the explanations would be similar if, instead of words, the basic input and output units were, for instance, characters.

The simplest RNN architecture has an input layer, a hidden layer, and an output layer connected to a Softmax function [165]. The input layer encodes the current word of the input sequence, the hidden layer encodes the current underlying state of the sequence, and the output of the Softmax function to which the output layer is connected might be assumed to encode the probability distribution for the next unit in the sequence.

Each word in the input layer is represented as a vector with the same dimensionality as the number of nodes in the input layer, which is the same as the number of words in the lexicon. All elements of this vector are equal to 0, except the position that encodes the considered word, whose value is 1. Such vectors are called one hot vectors.

After the network has been trained, the result of applying a Softmax function to the output layer can be interpreted as the probability distribution of the next word over all possible output words. This means that the output layer must also contain as many nodes as possible output words (often the same as the input layer).

According to Harrison, Purdy and Riedl [62], although recurrent neural networks have been successfully used in other natural language generation tasks, they have shown some difficulty maintaining a coherent story progression for more than a few events. Thus, a great deal of the research work using RNNs for story generation focuses on finding ways of improving the coherence of the generated stories.

One of the reasons for the lack of coherence of stories generated by RNNs is the fact that each sentence of the story comprises several words but the network generates a word at a time. In fact, it would be almost impossible to learn sequences of sentences because the same sentence appears only very few times even in very large corpora. Therefore, if the average sentence length is five words, generating a story with 10 sentences involves generating a coherent sequence of 50 words in average. Several authors (e.g., Roemmele [164] and Martin and colleagues [59, 166]) go around this problem, using RNNs in interactive settings in which a person and the RNN take turns generating sentences. This way, the human user will, if desired, ensure the overall coherence.

Roemmele [164] explored the use of RNNs as a tool for story auto-completion to help writers with possible continuations for their stories. The amount of editing performed by the writer on the story continuation suggested by the network is used by the system as feedback regarding the quality of the generated output.

Another attempt to mitigate the lack of coherence of generated stories is to adopt more abstract representations of the story sentences, which reduces sparsity (both during training and test). If the sentences are represented by more abstract representations, it is possible that several sentences originate the same representation. This increases the number of times the same event occurs in the training corpus and hence improves learning. Pichotta and Mooney [58] and Martin and colleagues [166] propose two similar event representations. For Pichotta and Mooney [58], an event is represented as a 5 tuple (v, $e_s$, $e_o$, $e_p$, p), where v is a verb lemma, $e_s$, $e_o$, and $e_p$ are nominal arguments representing the subject, direct object, and prepositional complement, and p is the preposition relating v and $e_p$. Pichotta and Mooney [58] propose two variants of the same representations regarding the information actually used in the event nominal arguments (subject, direct object, and the noun phrase of the prepositional complement). In their work, nouns may be represented by their corresponding noun lemmas or by entity identifiers, which are integers identifying the corresponding argument's entity according to a co reference resolution engine. Experimental results show that noun information improves the system performance.

Martin and colleagues [166] represent events as 4 tuples of the form ⟨*subject*, *verb*, *object*, *modifier*⟩. In some experiments, instead of a 4 tuple, a 5 tuple was used, containing genre, which is a number representing the genre cluster of the event. Named entities were replaced with the tag <NE>n, where n indicates the nth named entity in the sentence. Other named entities were labeled as their named-entity recognition category (e.g. location, organization, etc.). The rest of the nouns were replaced by the WordNet hypernym two levels up in the hierarchy (e.g. self-propelled vehicle:n:01 vs the original word "car" (car:n:01)). Finally, verbs were replaced by VerbNet frames [167] (e.g. "arrived" becomes "escape 51:1", "transferring" becomes "contribute 13:2 2"). This representation was used in other research (e.g., Harrison, Purdy and Riedl [62], and Martin and colleagues [166]). Martin and colleagues [166] tested their approach with different generalization degrees, such as using the original sentences, using event representations with the original words, and using fully abstract event representations. Experimental results indicate that generalization improves the generation of the next event. However, with more abstract representations, it is more difficult to obtain natural language sentences as the system's output.

Martin and colleagues [166] propose a two stage machine translation approach to generate the output natural language sentences. Their approach consists of generating abstract sentences from the events generated by the recurrent neural network and filling in the slots of the generated abstract sentences. In the first stage, they use a sequence-to-sentence neural network, which generates an abstract sentence for each event, in the sense that the sentences contain slots for the character and, otherwise, the subject, the object and the modifier are represented as WordNet *synsets*. This decreases translation errors by pushing the grounding of specific names and nouns to a later stage.

During the slot filling stage, the named entities and nouns in generalized sentences produced by sequence to sequence network are filled back in by searching the agent long term memory graph for the most salient entities. In this approach, agents have a long term memory, which consists of a graph in which nodes represent events and entities. When a slot cannot be filled using long term memory, it is handled by inventing new names by pulling them from the US Census or picking new nouns from WordNet hierarchies.

Another problem affecting RNN solutions is known as the exploding or vanishing gradient problem. During backpropagation, the network weights are updated as a function of the derivative of the error function with respect to the considered weight. When these derivatives may reach very large or very small values, the weight updating process rapidly explodes or vanishes. This problem may be tackled by using networks with Long Short-Term Memory nodes (LSTM, [168]). All the reviewed RNN proposals described in this section use LSTM nodes.

With the purpose of generating stories that satisfy defined criteria, Martin, Harrison and Riedl [59] proposed a deep learning approach combining LSTM recurrent neural networks with reinforcement learning, whose rewarding function encodes the desired criteria. Combining the expectation based strategy of recurrent neural networks with the value based strategy of reinforcement learning, it is possible to include both expected and non-expected events in the generated stories.

Harrison, Purdy and Riedl [62] propose yet a different approach with the goal of generating stories satisfying a given pre-defined criterion. In their proposal, the recurrent neural network is not used to generate the next event in the sequence of story events. Instead, the LSTM recurrent network is used to evaluate generated stories. Harrison, Purdy and Riedl [62] propose a Markov chain Monte Carlo process [169] of story generation. This approach incrementally learns to generate stories that satisfy an *a priori* unknown probability distribution reflecting desired criteria. Each time a new story is generated by the current probability distribution, it is evaluated against the previously generated story, according to the defined criteria, and the system decides whether or not to accept it.

If the generated story is accepted, the probability distribution is updated to reflect the new story, and the newly accepted story becomes the previous story for the next cycle. If the generated story is not accepted, it is discarded.

According to the criteria proposed by Harrison, Purdy and Riedl [62], a story is accepted if its events occur in a sequence consistent with the event sequencing pattern subjacent to the training story corpus; and if the story contains a previously specified sequence of possibly non-adjacent events. This second criterion is used to allow the user to specify stories containing certain desired events.

In this process, the Markov chain Monte Carlo (MCMC) initial distribution is a Markov chain trained with the available corpus of stories. This distribution is used to generate the words of the first story, one word at a time. As generated stories are accepted, this distribution is updated. In the end, after training, the system will be capable of generating stories adhering to the defined criteria.

One of the criterion for story acceptance, namely that the story must contain a specified sequence of possibly non-adjacent events, is implemented by a Skipping Recurrent Neural Network (S-RNN, [170]) trained, on the same corpus as the MCMC initial distribution, to extract the sequence of the *k* most important events in the story. S-RNNs learn to select a sequence of *k* elements in a sequence that minimizes the amount of information lost by removing all other items, without the constraint that those elements are adjacent in the original story. After training, this S-RNN network is used to extract the summary events of the generated story, allowing to check if they match the specified event sequence. The stories that satisfy this criterion are scored against the other criterion.

The other criterion, namely that the generated story events must reflect the event sequence subjacent to the story corpus, is the responsibility of a sequence to sequence encoder decoder LSTM recurrent network trained with the same data as the initial MCMC distribution and the S-RNN. After training, this sequence to sequence network produces the probability that a certain event follows another event. During the process being described, each sequence of the words comprising two adjacent story events are scored with the probability generated by the sequence to sequence network. The story is scored by summing the scores of each sequence of two events. Given that story events are represented as in [166], each pair of events is a sequence of eight words.

The experiments performed by Harrison, Purdy and Riedl [62] show that, after training, the proposed approach generates stories adhering to the specified criteria 85% of the time.

# 9   Contributions to the art of good stories

No one knows for sure what a good story is; otherwise a perfect story generation system would have been created already. Some say that, to be interesting, stories must have some form of suspense [148]. Others focus on tension or conflict. Tension should increase until near the end, and the main character should learn something [37, 52, 105]. Others have other kinds of beliefs, such as that a story must revolve about demonstrating some point or moral (e.g., Orhan Pamuk [171]) or, on the contrary, that it is enough to have good and rich characters to create a good story (e.g., Ruth Aylett [172], and Andrew Stern and colleagues [173]). Many agree that character believability is important for good stories (e.g., Mark Riedl and Michael Young [35, 41]). In spite of the lack of a widely accepted notion of what good stories are, this section presents approaches claimed by the involved researchers to contribute to improve the quality of the generated stories.

Possibly, the most widely spread attempt to improve the quality of the generated stories consists of creating systems whose generation process is somehow conditioned by the goals the author intended for the narrative. However, most of the reviewed story generated systems talk indistinctly about author and user goals. At least implicitly, this means that the researchers that created those systems do not actually take them as authors, which is a bit disappointing, at least from the standpoint that the ultimate purpose of computational creativity is the development of computer systems that could be considered the authors of the generated artefacts (e.g., stories).

Indeed, several researchers (e.g., Guckelsberger, Salge and Colton [13], Linkola, Kantosalo, Männistö and Toivonen [80], Ventura [14, 82] and Bay, Bodily and Ventura [15]) have emphasized the requirement of intentional creativity, which is tightly related to the mentioned problem of the system with goals, that is, the view of the system as an author. However, nothing is said about how the system's high-level goals can unfold into its intended objectives for the story to be generated.

Maria Teresa Llano and colleagues (e.g., [174, 175]) developed mechanisms for the generation of fictional ideas, which is a valuable contribution. However, to the best of our knowledge, no one provided the links necessary for a creative system to adopt the generated fictional ideas and commit to the generation of stories at the service of such fictional ideas.

The generality of the story generation systems provide some form of consideration for the goals of the authors. For instance, in the system proposed by Porteous and Cavazza [29], the author's goals specify a partially ordered sequence of world states the story must satisfy. In this case, they condition the general outline of the story. If carefully thought, they may contribute to the generation of good stories. In intent based planning (e.g., [22, 28, 30, 142, 149]), the author's goal specifies the goal state for the planning algorithm. In these cases, the author's goals determine the way the story ends. The story characters determine the way the story may achieve the final state.

The author's goals may assume different forms in different kinds of systems. In case-based systems (e.g., [37, 132]), they usually specify some of the actions of the story to be generated (e.g., *the princess went to the market*, *the princess bought a nice blanket*, *the princess met jaguar knight*). In schema-based systems (e.g., [45, 46, 48]) or in case-based systems where a case consists of a schema (e.g., [32, 128]), the author's goals are specified as a sequence of high-level narrative functions, for example "villainy, departure, villain punished, return". In analogy-based story generators, other than CBR (e.g., [40]), the author's goals may specify a story to be transformed, its domain and the target domain, which condition the story to be generated as output. The same kind of specification is provided to the system based on evolutionary algorithms proposed by Baydin, López de Mántaras and Ontañón [34]. In the system developed by McIntyre and Lapata [33], the user provides a sentence describing the initial state of the story (e.g., the princess loves the prince) and may also specify the number of episodes desired in the generated story. TALE-SPIN [24] allows the specification of the moral the story will have to convey.

Good characters is certainly one of the most important and pervasive prerequisites for the generation of good stories, especially if a significant number of story generation decisions depend on the richness of the story characters. Although most reviewed work is not particularly focused on story characters, this section starts with a brief description of the properties of the story characters in story generation systems (subsection 9.1).

The purpose of narrative knowledge is the generation of good stories. Subsection 9.1 of this section describes the way narrative knowledge is used in each general class of the reviewed story generation approaches, and the narrative knowledge additionally used by specific systems to improve the quality of the generated stories.

Maybe good stories have a better chance of being generated if the space of possible stories is larger or if the mechanisms used to explore it have access to more of its regions. Thus we describe several approaches that increase the number of the stories that the system may generate, with the hope that better stories are produced. The attempts to improve story quality through the extension of the space of possible stories is described in subsection 9.3.

Finally, in subsection 9.4, we spot the possible lack of narrative knowledge in some of the reviewed approaches, assuming that randomly made decisions may signal such lack of knowledge. We briefly identify the types of decision situations that are solved by chance in some systems and discuss the knowledge that could be missing in such systems.

## 9.1 Story characters

The characterization of the story characters, in many story generation systems, is one of the most determinant aspects of the story quality because the selection of the actions to be included as story events strongly depends on the story characters. Action preconditions and effects often depend on and change several character properties (e.g., beliefs about the world). BDI-like models of characters (e.g., [137, 138]) are often used to describe and control characters. Planning-based story generation systems are amongst those that rely on BDI-like models. Character beliefs appear in action preconditions of virtually all planning-based systems (e.g., [22, 28-31, 35, 39, 43, 142, 149]). Character goals, which may be transformed into intentions during planning, are the fundamental piece of intentional frames in intent-based planners such as IPOCL [28], CPOCL [30], Glaive [22] and Impractical [142].

In Impractical [142] and in HeadSpace [43], each story character has their own beliefs, which may be wrong. While the character beliefs in HeadSpace change as a result of the action effects, Impractical has a set of rules defining a general policy regarding the way the character beliefs change as a consequence of an action being executed, for example, "If an action changes an object's location, all agents at its previous and new locations perceive predicates relating to this object", "The actor and all participants of an action perceive its preconditions and effects", and "If an action is public, all agents perceive its preconditions and effects".

BDI-like models are not used only in planning-based systems. For instance, StellA [52] is a rule-based system (section 5) that uses a BDI-inspired approach. StellA has rules determining perception actions, which are responsible for the creation of beliefs, rules determining the creation of desires, and rules determining the creation of intentions from beliefs and desires.

Several story generation systems use character models with different or additional concepts. The system proposed by Permar and Magerko [50] uses a measure of iconicity of the story characters. Since iconicity is highly dependent on the context (e.g., an entity may be iconic in a context but not in another one), that knowledge must be dependent of the context.

INES [48] uses a relation regarding the class and intensity of affinities between characters, and a set of rules that determine the way the affinity between two characters changes as a consequence of interaction (or the lack of it).

Joseph [113] uses knowledge regarding the emotions the protagonist feels as a consequence of story events, and about the actions and goals that result of the felt emotions.

CB-POCL ([26, 27]) is a planning-based story generation system in which, the protagonist chooses the course of action most consistent with their personality. Whenever several alternative courses of action are possible for the same character, CB-POCL evaluates the degree to which each specific course of action is consistent with the assumed personality of the story character. CB-POCL uses a set of rules that compute the consistency of a personality trait and an action effect.

Several story generation systems use significantly rich character models (e.g., [24, 39, 125, 176]), including such properties as personality traits (e.g., kindness), emotional state (e.g., fear), physical traits, beliefs (e.g., the princess is in the room next door), general needs (e.g., hunger, thirst, sex, rest) or goals (e.g., kill the villain) and interpersonal relationships (e.g., feeling indebted towards some other character). This allows the stories to be both more believable and more interesting.

NOC (Non Official Characterization list) [125] is a publically available database of more than 1000 iconic real and fictional characters used in story generation systems (e.g., Scéalextric [47] and Scéalextric Simulator [104]). The NOC characters are abundantly described with much and diverse information as gender, locale (e.g. New York, Tatooine), style of dress, spouses or lovers, known enemies, apt vehicles, trademark weapons, relevant domains (e.g., arts, science, politics), semantic types (e.g., politician, playboy), fictive status (fictional or real), genres (e.g., science fiction), creators and screen actors (if fictive), typical activities (e.g., building casinos, running political campaigns), political leanings (left, right or moderate) and group affiliations (e.g. Tony Stark belongs to The Avengers, Eliot Ness belongs to The Untouchables, Darth Vader belongs to the Dark Side, and Donald Trump to the Republican Party).

Each NOC character possesses a set of affective beliefs toward past actions known to the character, as well as the intensity and type of the character's feeling toward it. Intensity, an integer, ranges from 1 to 9, while type can be proud, guilt, admiration or shock. Characters feel pride or guilt for their own actions and admiration or shock for the actions of others.

TALE-SPIN's characters [24] are capable of sophisticated and varied kinds of inference. For instance, they can infer that someone that stays in the water for a long time will drown, and that if someone drowns then that someone will die. Inferences may also take place in the domain of feelings towards others. For instance, a character A may become devoted to another character B, if B rescues A from dying.

In addition to the mentioned character properties, the character model used in UNIVERSE [176] also includes offspring and a list of historic events. In UNIVERSE, characters are partially described through the use of stereotypes and additional individual properties that are not covered by the used stereotypes.

UNIVERSE knowledge base contains a set of definitions of stereotypes describing both stereotypical classes of people (e.g., doctor, socialite, big eater and junkie) and stereotypical interpersonal relationships (e.g., doctor-patient and divorced-mother). The description of a particular character relies on the specification of stereotypical information potentially complemented with and/or superseded by individual information.

In UNIVERSE, each stereotype used to describe a person consists of a set of attribute/value pairs. The set of attributes includes but is not limited to wealth, promiscuity, intelligence, age and sex. The stereotypes used to describe an interpersonal binary relationship consists of the name of the relationship, a set of constraints on the two related roles (e.g., the possible stereotypes of each role) and the evaluation of each participant regarding the relationship with the other participant, along four dimensions (i.e., positive/negative, intimate/distant, dominant/submissive, and the degree of attraction).

Of all the reviewed approaches, UNIVERSE [21, 176] is the only one for which there is a description of the process of automatic character creation [176]. The director of the Virtual Storyteller [25] may add new characters to the plot being generated but nothing is said regarding the process of character creation (if any).

Due to author goals, it is often necessary to create a story character with a specified set of characteristics. However, it is not enough for a character to have a specified set of personality and physical traits. Characters must have history and interpersonal relationships. When UNIVERSE creates a new character, it steps through their life, creating spouses and children, and a set of past events, which in turn give rise to interpersonal relationships. UNIVERSE uses simplified versions of previously created stories as the source of consistent past lives for the new characters.

When the simulation of the past life of new character past life reaches the present, UNIVERSE chooses an occupation for the character corresponding to the occupational stereotype that includes the largest number of the required characteristics. If not all the required traits are covered by the selected occupational stereotype, UNIVERSE pics other stereotypes that bring the new character closer to the specification, without conflicting with already existing traits. Finally, it might be necessary to add individual traits to the character. As a result of this process, the new character will have all needed traits plus all the traits inherited from the assigned stereotypes, which is important for believability. It would be possible to simply adding the specified set of traits to the new character, but the result would most likely not be believable. In addition to improving believability, the described process may lead to interesting but coherent characters, such as a videogame-addicted warden, with intricate but consistent past lives and interpersonal relationships.

During the process of creating a new character with a past, it might be necessary to create other characters. For instance, if the new character has children, it might be necessary to create them. Hence, new characters may be created to satisfy author goals or as a necessity arising of the process of creating and rehearsing other characters.

## 9.2 Narrative knowledge

Narrative knowledge is procedural knowledge used to generate good stories or declarative knowledge about the properties of good stories. The latter can be used both for generation (e.g., in grammar based story generation systems as *Joseph* [113]) or, more frequently, in the evaluation stage of the generation process (e.g., [129, 177]). In spite of its apparent simplicity and conciseness, this characterization of narrative knowledge is not always enough to distinguishing narrative knowledge from other kinds of knowledge in a story generation system. To illustrate the ambiguity, we briefly consider the knowledge used in story generation systems to control the actions of the story characters.

For example, in planning-based systems, the control of the actions of the story characters is exerted essentially by the planning algorithm. Action representation includes a set of preconditions and a set of effects. For example, one of the prerequisites for a character to move from place *A* to place *B* is that it is located at *A*; and the effects are that it is located at *B* and no longer located at *A*. It is easy to accept that this knowledge (the preconditions and effects of the action) is not about good stories. Rather, this is knowledge about the properties of world in which the story unfolds. However, the distinction becomes blurred when the personality of the characters, their affective state or their intentions are considered when choosing their next actions, with the goal that the generated behaviors are perceived by the reader as believable. Character believability is certainly one of the most important aspects of narrative knowledge (e.g., [26-28, 35, 41]). The difficulty to tell narrative knowledge from other kinds of knowledge increases, when the algorithm that chooses the actions of the story characters, besides considering the preconditions and effects of the possible actions, considers also the goals the author intended for the narrative (e.g., [22, 30]). In such cases, controlling the characters relies on the narrative knowledge, according to which, good stories must be shaped by the appropriate balance between character and author goals (e.g., [178]).

Given the discussed difficulty, we chose to present cases of the deliberate efforts of diverse approaches to improve the quality of the generated stories, trying not to focus on the control knowledge exclusively related with the physical conditions required by the considered actions.

We start by identifying, in the general case, where the narrative knowledge (the knowledge about good stories) operates in each general class of the technological approaches reviewed in this article. Then we describe more specific efforts to improve story quality.

Given that character believability is presented by many researchers, as one of the most important aspects of story quality (e.g., [26-28, 35, 41]), we discuss the ways several systems improve character believability.

Before we move to the description of the narrative knowledge used in the reviewed classes of approaches and in specific systems, we present two distinctions regarding narrative knowledge: implicit vs. explicit narrative knowledge, and narrative knowledge that is used or specified separately from vs. together with other knowledge used in the system. These distinctions are illustrated during the discussions to follow.

Some examples of story generation systems explicitly represent narrative knowledge whereas others implicitly use narrative knowledge reflected in the way their algorithms work. The degree to which different aspects of narrative knowledge are explicitly represented may vary along a wide range of possibilities. It is possible to implicitly integrate the need for coherence in a system's planning algorithm and explicitly represent a rule stating the facts that should occur in some states of good stories. The story generation systems that implicitly use narrative knowledge in their algorithms have the advantage of taking care of important narrative decisions without requiring the explicit specification of the required knowledge. However, in such systems, it may be impossible to freely specify the desired narrative knowledge or the conditions under which it should be used.

Any story generation system may represent explicit narrative knowledge separately from or together with other kinds of knowledge used by the system. For example, *StellA* [52] keeps explicitly represented narrative knowledge totally separated from character control knowledge whereas the rules used to control the story characters, in the *Novel Writer* [51], may also make decisions reflecting narrative knowledge. Arguably, a story generation system is more flexible if the same rule can control a story character and express narrative knowledge because designers may control whatever aspect of the story generation process they want. However, if the story generation system does not separate the explicitly represented narrative knowledge, it might be impossible to use the same narrative knowledge in different story worlds with different characters.

This survey of non-interactive story generation reviewed systems based on story schemas/templates, analogy, rules, planning, evolutionary algorithms, explicit knowledge learning, and implicit knowledge learning. This subsection describes the way narrative knowledge generally operates in these classes of systems and the specific efforts made in some of the concrete systems to improve story quality.

We start by considering all systems that include an automatic evaluation stage in the generation process, which constitutes an enormous subset of the reviewed ones. Independently of the involved technological approach, the evaluation of a complete or an incomplete story always involves some form of implicit (e.g., [179]) or explicit (e.g., [129, 177]) narrative knowledge. For instance, at each reflection stage of the engagement-reflection loop of the MEXICA system [37], MEXICA checks if the story events produced so far contribute to the degradation improvement process thought to be essential in all good stories. If not, MEXICA creates new constraints that will condition the next engagement stage to use more adequate events.

A significant amount of narrative knowledge in schema-based systems (e.g., [45, 46, 48]), including the case-based reasoning systems whose cases represent abstract story templates (e.g., [32, 63, 128]) or those whose story schemas are previously learned from a corpus of stories (e.g., [33, 44]), is captured in the story schemas. In effect, a story schema specifies the several ways a good story may unfold. The basic idea is that each possibility is a good story; it is not necessary to make special efforts to conduct the story in an adequate direction with the aim of eventually producing a good one. For each possible story, the schema also implicitly specifies the author's goals because it specifies the way each story ends.

This does not preclude the possibility of adding further narrative knowledge to schema-based systems, in the attempt of generating stories with even better quality. For instance, Li, Lee-Urban, Appling and Riedl [44] propose that different heuristics, capturing narrative knowledge, may be used, in schema-based story generation, to generate stories with different general properties, such as, comprising an highly unlikely event in the midst of a sequence of likely events, with the purpose of triggering the feeling of surprise.

The cases of some case-based story generation systems (e.g., [23, 37, 49]), including those that are automatically extracted from a set of available stories using learning algorithms [132], are supposed to represent examples of good quality stories. Thus the narrative knowledge of such systems is implicitly captured in the stories in their case bases. However, as it was the case with schema-based systems, it is possible to use additional narrative knowledge to improve the generated stories.

Rule-based systems (e.g., [25, 51, 52]) are not enough normalized to enable us to describe the general way narrative knowledge operates. The use of narrative knowledge in rule-based story generation systems is described later in this section.

The narrative knowledge generally used in planning-based story generation systems (e.g., [22, 27-31, 41, 42, 149]) addresses the compromise between character believability and author goals. On one hand, several planning algorithms were modified to create plans that emphasize the believability of the story characters, revealing their personality [27], their emotions [180], their intentions [28, 41], or their conflicts [22, 30, 31].

On the other hand, all mentioned planning-based approaches allow the author's goals to be achieved without substantially endangering the believability of the story characters. This is achieved basically in two ways. In one (e.g., [29, 42]), the author's goals are used for establishing a sequence of world states that must be achieved during the course of the story, whereas the planning algorithm is repeatedly used to achieve each of those states at a time, considering only the behavioral capabilities of the story characters. In the other (e.g., [22, 27, 28, 30, 41, 149]), the planning algorithm generating the sequence of story events only selects actions that, while contributing to the author's goals, can be perceived by the reader as believable from the point of view of the intervening characters.

The narrative knowledge that conditions the operation of story generation systems based on evolutionary algorithms (e.g., [33, 34]) may intervene in the generation of the initial population, in the crossover and mutation operators and in the fitness function. In principle, the narrative knowledge used in evolutionary approaches could address a multitude of concerns (e.g., author's goals, the story coherence and interestingness, common sense compliance). However, the reviewed approaches focused on only a few. The initial population may be chosen to comply with common sense knowledge [34] or with the goals of the author [33]. The crossover and mutation operators may be designed to perform changes that comply with common sense knowledge [33, 34]. And finally, the fitness function may rely on the story coherence or on the author's goals.

Systems based on implicit knowledge learning (e.g., [58, 59, 61, 62, 164, 166]) acquire narrative knowledge during training. In general, these approaches use recurrent neural networks to learn sequencing narrative knowledge from very large training corpus of stories. However, it is not feasible to directly learn the sequence of story events because, even in very large training corpus, any given sentence (representing an event) only appears a very limited number of times. Thus the systems learn the sequence of words in a story, not good event sequences. However, specific implicit knowledge learning systems rely on more elaborate approaches that learn narrative knowledge.

In addition to the narrative knowledge generally found in the different reviewed technological approaches, specific systems make specific efforts to improve story quality. Of the reviewed literature, we emphasize the use of narrative knowledge aimed at improving character believability (e.g., [22, 26-28, 30, 37, 41, 105, 142, 149]), suspense [148], dramatic tension (e.g., [37, 52, 105]), the order by which or the conditions under which certain story events will happen (e.g., [21, 51]), the relation between the story initial conditions and morals to be conveyed by the story [24], among other concerns.

MEXICA (e.g., [37, 105]) proposes two ways of improving character believability: selecting the story events during the engagement stage of the engagement-reflection loop, taking into account the emotion links and tensions between characters; and, in the final step (after the engagement-reflection loop has stopped), creating the character goals that would explain their exhibited behavior. Although MEXICA selects actions, during engagement, considering character emotions and tensions, the story does not explicitly tell the reader about those states.

In intent-based planning (e.g., [22, 28, 30, 41, 142, 149]) the planning algorithm generating the story selects actions that contribute to achieve the author's goals. However, all selected actions must belong to an action sequence to achieve the intentions of a character. This way the events in the story will be better accepted by the reader because they were executed at the service of some character intention. Additionally, the conflict based planning algorithms described in [22, 30] emphasize the conflicts between story characters. The generated story may use actions intended to be performed by a character but that were thwarted due to the actions of another character.

In addition to improving the perception of intentionality regarding the character behavior, the last reviewed version of the IMPRACTICAL system [142] adopts a model of possibly wrong character beliefs that allows deception. In IMPRACTICAL, a character may plan to create a wrong belief in another character. This way the deceiving character is capable of making the deceived character act at the service of the goals of the deceiver. This allows a character to act believably even when not at their best interest. Besides, this possibility reflects the narrative knowledge according to which, deception may contribute to story quality.

Bahamón, Barot and Young [26, 27] extended the partial order planning algorithm to make better use of the choices each character faces when selecting a given course of action among several possible alternatives. In addition to using rich characters as those used in UNIVERSE [176] or in NOC (Non Official Characterization list, [125]), Bahamón, Barot and Young [26, 27] created a planning algorithm that generates plans in such a way that the reader will interpret the behavior of the character as revealing their personality. The subjacent idea is that the reader will infer the personality of the character from the observations of the choices they make along the narrated sequence of events. Whenever a character has a choice of possible actions to achieve a certain goal or sub goal, the proposed algorithm evaluates the alternative sequences of actions that start at that choice point and rates each of them according to the degree to which it reflects the intended personality of the character. If there is enough contrast among the possible courses of action, the algorithm presents the most contrasting alternatives to the reader, facilitating their interpretation of the character behavior as consistent with their personality. According to the authors, this process significantly contributes to enhance character believability and story quality.

According to Brewer and Lichtenstein [181], and to Alwitt [182], suspense contributes significantly to the enjoyment of a narrative by its readers. Recognizing such an importance, Cheong and Young [148] developed the *Suspenser* system, which reorganizes the presentation of the events of a received story with the aim of triggering the specified type of suspense from the reader (high vs. low) at the specified moment. In the first stage of its processing, *Suspenser* creates a minimum version of the received story, containing only the most important events that are enough to tell the received story without losing information. The importance of a certain event e was operationalized as the number of previous events that established preconditions of e plus the number of effects of e that are used by future events. Additionally, opening events, closing events and events that establish one of the planning goals are more important. In the second stage, *Suspenser* creates the mechanisms responsible for triggering the desired type of suspense (high vs. low) from the reader. *Suspenser* adds events, from the original story, with high (vs. low) level of suspense to the minimum version of the story. These events are presented before the moment the reader should feel the desired suspense. Finally, *Suspenser* delays other events in the original story, with a low (vs. high) level of suspense, until after the specified moment. For Cheong and Young [148], the level of suspense of an event increases with the number of its effects that negate one of the protagonist's goals, decreases with the number of its effects that unify with one of the protagonist's goals, and varies inversely with the estimated distance between the considered effect and the considered goal.

The dramatic tension of the story was addressed in MEXICA [37, 105] and in *StellA* [52] relying on separated narrative knowledge explicitly represented in the systems.

The design of the MEXICA system [37, 105] relies on the belief that interesting stories must have degradation improvement processes. During degradation, the tension to the reader should increase until it reaches a maximum. Then, during improvement, the tension decreases. In MEXICA, the choice of actions to insert in the story in progress depends on the tensions between characters and also on the tension perceived by the reader. On one hand, during the engagement process, actions are selected considering the tensions existing between the story characters. On the other hand, whenever the story being generated does not involve adequate patterns of tension variation, MEXICA generates constraints that disallow it to choose actions, in the next engagement process, that do not match the necessary tension dynamics. The generation of the tension based constraints that shape the action selection process is totally separated from the character control knowledge. However the description of each action includes narrative knowledge properties (emotion and tension) as well as properties more related with the usual character control knowledge (e.g., the knight is injured, the enemy kidnaped the princess).

Possibly, the story generation system with most sophisticated means for narrative knowledge representation is *StellA* [52]. In *StellA*, narrative knowledge may be expressed in the form of objectives, constraints and objective curves. Objectives are declarative rules that specify the properties of a finished story (e.g., in a finished story, there are no humans in the dungeon). Constraints are declarative rules that specify the properties of stories either finished or unfinished (e.g., in any acceptable story – finished or not –, the distance between any two humans is less than a predefined value). Constraints are used by the system to discard unpromising alternative courses of action (e.g., if the distance between two humans is not acceptable in a given course of action, that alternative is discarded). Objective curves specify the desired evolution of certain story properties (e.g., in any story – finished or not –, the danger must increase). *StellA* computes the degree to which all alternative courses of action satisfy the objective curve. Alternative courses of action whose degree of satisfaction of the objective curve is below the defined threshold are abandoned. Of the acceptable courses of action, *StellA* prefers the one that best satisfies the objective curves. Objective curves are a generalization of tension curves, which were previously proposed by Pérez y Pérez and Sharples [37, 105].

In *StellA*, it is enough to specify the rules that define objectives, constraints and objective curves. The system automatically uses the represented knowledge in the story generation process. It checks whether a finished story satisfies all objectives, and it maintains only those alternative plot lines that satisfy the constraints and the objective curves, discarding the others.

HEADSPACE [43] also contributes to the dramatic tension of the generated story by allowing the story characters to execute actions that fail. As in [142], HEADSPACE also models the possibly wrong beliefs of the story characters. However, instead of using the wrong beliefs for deception, HEADSPACE uses them to allow characters to attempt to execute actions that fail because the beliefs of the character about the satisfaction of the action preconditions happen to be incorrect.

The importance of dramatic tension is the focus of the PROVANT system [31]. Relying on the narrative knowledge that good stories may depend on the existence of a protagonist whose goals are attacked by an antagonist, PROVANT, in addition to creating plans intended to satisfy the goals of the protagonist (given to the system, as input), automatically generates the antagonist and their behavior. The algorithm identifies the protagonist's goals and its landmarks (facts that must be true somewhere in any plan that achieves a certain goal). Then it automatically plans the behavior of the antagonist with the goal of deleting the preconditions of protagonist's actions whose effects establish the landmarks of their goals. In the generated stories, tension increases until the moment in which the antagonist attacks the goals of the protagonist, and decreases as the protagonist recovers from the attack.

Finally, the already described possibility of generating stories with deception [142] also contributes to dramatic tension because deception is a possible ingredient of tension.

The *Novel Writer* [51] and the UNIVERSE [21] use narrative knowledge that allows them to set or to change the order by which or the conditions under which certain story events will happen. In the *Novel Writer* [51], each group of character control rules is associated with an evaluation frequency, which can be dynamically changed. The system evaluates each group of rules according to its evaluation frequency. In addition, a defined evaluation sequence determines the order in which the system evaluates the different groups of rules. The evaluation sequence can also change dynamically. The evaluation frequency and the evaluation sequence constitute mechanisms through which the system can exert narrative control over the story. The same rules used in the *Novel Writer* to control the characters can also change the evaluation frequency and the evaluation sequence. That is, the *Novel Writer* does not force the user to separate character control knowledge from narrative knowledge.

The planning algorithm of the UNIVERSE system [21] also allows the user to define the conditions under which a given goal may be pursued by the system. Instead of allowing the user to define the evaluation frequency or the evaluation sequence of the character control rules, UNIVERSE allows the user to associate a set of pre-conditions with each goal. The planner can only choose to try to achieve a goal whose pre-conditions are satisfied. This mechanism allows conditioning the unfolding of a story according to criteria that are thought to hold in good stories.

Even TALE-SPIN [24], which has often been pointed as not using narrative knowledge, uses in fact narrative knowledge to set the initial situation of the story world (e.g., the story characters and their goals) that allow the generation of a story that conveys the specified moral (a story point). For this, TALE-SPIN uses knowledge linking general abstract morals (or story points) to the concrete things and events in the story domain.

The *Virtual Storyteller* [25] offers significant flexibility, with respect to narrative knowledge representation, because it allows the user to specify any kind of narrative knowledge (not only objectives, constraints and objective curves). This increased flexibility is exactly the reason that prevents the system from automating the use of represented narrative knowledge. Hence, contrarily to what happens in *StellA*, all narrative knowledge can be used during all stages of the generation process. As a consequence of this flexibility, if a certain rule is not supposed to apply in a given stage of the process, the rule must explicitly include the necessary conditions to prevent its usage in that stage.

Narrative knowledge is represented, as a set of rules, in the knowledge base of the Director of the *Virtual Storyteller*. The Director uses its narrative knowledge when the story characters ask it permission to execute an action, when it generates the goals it assigns to the story characters, and when it changes the environment in which the story takes place, for example, adding new objects or characters.

The narrative knowledge used in the *Virtual Storyteller* [25] is explicitly represented and it is separated from the character control knowledge.

Knowledge about good stories is sometimes used in story evaluation processes with the purpose of determining if the story can be considered the system's output, if the system should continue its creative process with the story being generated, or if the story should be discarded. For instance, the arguments used in the system presented in [129] to determine if the generated stories are good stories represent declarative narrative knowledge about the properties of good stories or merely commonsense knowledge about the story world (e.g., the relatives of a human are also humans).

Wang, Chen and Li [179] studied several alternative general approaches that can arguably be used by any system to automatically evaluate the quality of the generated stories. First, they built a classifier (using different classification algorithms and different sets of features) to distinguish a story from a non-story. Experiments show that the Random Forest algorithm [183] with 500 trees operating with bag of words frequency [184] yielded the best classification results.

Next, the authors conceived and studied three deep neural network models to predict the story quality [179]. They have operationalized the story quality as the predicted number of up votes the story would receive from users in a social media website.

All the three models organized each story text in word chunks (regions), each of which was processed by three convolutional network layers [185] followed by a max pooling layer [186] and a fully connected layer to reduce dimensionality. In the first model, the outputs of the regional processing were linked to two fully connected layers. In the second model, the outputs of each regional processing, before being sent to the final two fully connected layers, were linked to a recurrent neural network (with or without long Short-Term memory nodes) capturing the interdependence between adjacent regions of text. The third model is similar to the second one but it captures interdependences between all regions; not only between adjacent ones.

All three models operated on a set of features for example word embeddings, the number of words representing objects (which are evidence of prompting mental imagery and thus story quality), words with positive or negative charge, the number of times a story was viewed and its logarithm, the number of words, images and grammatical mistakes.

Experiments show that the third model with long Short-Term memories is the best model, which represents a significant improvement relative to the random forest baseline, which was the best of the classical models. The paper also concluded that using the story text is better than using only the selected linguistic features.

With a similar purpose of that of Wang, Chen and Li [179], namely to create an automatism capable of evaluating generated stories in the place of one or more human evaluators, Purdy, Wang, He and Riedl [187] proposed a set of features they assumed to correlate with global story judgments: grammaticality, temporal ordering, local contextuality, and narrative productivity (reading ease and lexical complexity). Experimental results with human evaluators show that these features positively correlate with global story judgments (i.e., story quality and story enjoyment). The paper proposes a set of machine learning based algorithms that automatically compute measures of the proposed features.

To automatically assess grammaticality, the paper proposed to use ridge regression [188] from the GUG dataset (Grammatical versus Ungrammatical) using the number and proportion of misspelled words in a sentence, and the max-log and min-log probabilities of the count of n grams in the sentence, according to English Gigaword.

Given the difficulty of devising a single measure for narrative productivity, the paper proposes to measure reading ease using the Flesch Reading Ease score [189] and the SMOG Index (Simple Measure of Gobbledygook, [190]), and lexical complexity using the type token ratio and corrected type/token ratio [191].

Local contextuality measures the semantic relevance of sentences in the context of their neighboring sentences. The semantic similarity between two input sentences was determined using the cosine between their embeddings, obtained from the CMU Plot Summary training corpus. The local contextuality score of a story is the average of the semantic similarities of every pair of adjacent sentences in the story.

Temporal ordering assesses the plausibility that one sentence should follow another in a story, even if the two are not adjacent. Given that the two sentences do not have to be adjacent, the paper trained a Skipping Recurrent Neural Network (S-RNN, [170]), using the CMU Plot Summary Corpus, to build a directed graph representing temporal relationships between the verbs of story sentences. To mitigate the sparsity problem associated to the very low frequency of occurrence of the same sentence in a story corpus, the paper represents each sentence as a tuple (called an event) ⟨subject, verb, object, and modifier⟩ (as proposed in [166]; see section 8 of the present article). Given any two sentences in a story, they are considered to be in the correct temporal order if their verbs are in the right order in the referred graph (considering that temporal orderings are transitive). The temporal ordering score for the story is the number of properly ordered pairs divided by the total number of pairs for which both verbs appear in the graph.

Finally, the paper reports experimental results that show that the automatically extracted features positively correlate with human judgements of the same features, although apparently to a lesser degree than that of the correlation between the proposed human evaluated features and the global story judgements. Unfortunately, the paper does not present the correlation between the automatically measured features and the human global story evaluations.

Pérez y Pérez [177] proposes an automated process capable of evaluating if the output of a story generation system is creative. The proposed process can be used both in MEXICA [37] to evaluate the stories it generates, and to evaluate the stories generated by other systems. According to Pérez y Pérez, a computer model of evaluation must consider if the evaluator/reader, as a result of the assessment/reading process, incorporates new knowledge structures into its knowledge base (e.g., new points of view). This corresponds to the novelty criterion advanced by Margaret Boden (e.g., [11]) and endorsed by many researchers. The computer model of plot evaluation must be able to recognize if the events that comprise a narrative increment and decrement the dramatic tension, which is possible only if the evaluator (which might be a computer program) has affective responses. Given that dramatic tension may improve story enjoyability, this criterion matches the interestingness criterion often mentioned about creative outcomes. The evaluator (a computer program) must also be capable of determining if the sequence of actions that comprise a story satisfies common sense knowledge. This proposal by Pérez y Pérez may be seen as corresponding to the coherence requirement usually adopted by story generation researchers.

To determine if the generated story is creative, the proposed system evaluates a set of core characteristics, according to novelty, interestingness and coherence: opening, closure, climax, reintroducing complications, satisfaction of preconditions, repetition of sequences of actions and novel knowledge structures. Pérez y Pérez defined the criteria used to evaluate each core characteristic (e.g., a story has a correct opening when, at the beginning, there are no active dramatic tensions in the tale and then the tension starts to grow). Finally, the proposed system uses a set of enhancers and debasers, which may increase or decrease the evaluation of the considered story. For example, if there are too many repeated sequences, the story evaluation is decreased.

The evaluator generates a report to explain the criteria employed during the process of evaluation. The report is divided in four sections: section one includes a general comment about the whole narrative; section two provides observations about the story's coherence; section three incorporates notes about the story's interestingness; and section four offers comments about the narrative's novelty.

In [192], Pérez y Pérez describes two agents (implementing the MEXICA model of story writing, but using different story case bases) that collaborate to automatically generating stories. In addition to the criteria defined in [177], for a story to be considered the result of a collaborative creative process, neither of the collaborative agents alone would be capable of generating it. After the two agents have generated the stories, the evaluation procedure determines if any of the agents could have generated them alone, the quantity of new knowledge incorporated into the agents, and if the agents use the acquired new knowledge for generating new stories.

## 9.3 Increasing the space of possible stories

With the goal of improving the creativity of their systems, by allowing them to consider a larger spectrum of possibilities, several researchers have presented approaches with the goal of extending the space of possible stories. This section considers three classes of approaches, namely relaxing world descriptions and/or softening constraints, creating new search operations, and transforming the initial creative problem in another one.

We start with approaches that widen the space of possible stories using different types of relaxation of the world description and / or softening of constraints. These include approaches relying on world properties with unknown value or on discarding a certain type of action preconditions, approaches with incomplete world descriptions, and approaches with characters with defective belief sets. In *Fabulist* [35] and in *StellA* [52], the systems interpret unknown world properties and a certain type of action preconditions in such a way that they can generate a story that otherwise they would not be capable of. In their approach, the story characters can perform more actions than they would if they observed all required conditions.

Riedl and Young [35] developed two proposals, initial state revision and recommendation planning, aimed at increasing the space of possible stories to be explored by the IPOCL (Intent-based Partial Order Causal Link) planner, contributing to improve the creativity of the system while preserving the believability of their characters.

The generation process is constrained by the description of the initial state of the world and by the preconditions of the actions the story characters can perform. Each of the two proposals tries to find sensible ways to avoid unnecessarily constraining the story generation process.

In initial state revision planning, the initial state of the story world is represented as a set of propositions assumed to be true and by a set of propositions whose truth-value is undetermined. When the planner needs to know the truth-value of undetermined propositions, it assumes whatever truth-value is convenient for enabling required actions to take place. Consistency requirements may however lead the planner to backtrack and assume different truth-values for the undetermined propositions. Given that the system may choose different truth values for a set of propositions, the space of possible stories is larger than it would be if these choices were not available.

The preconditions associated with the set of possible actions may refer to static and fluent propositions. Static propositions, for example those describing the personality of story characters, do not change as the result of the execution of the actions the story characters perform. Fluent propositions, for example those representing the location of the story characters, may change as the result of character actions. Recommendation planning consists of interpreting action preconditions involving static propositions as mere recommendations instead of hard constraints. For instance, instead of assuming that, for killing somebody, a character must be violent, the planner assumes that it is recommended that the character is violent (i.e., it is a good idea but not mandatory). When a character violates a recommendation (e.g., when a non-violent character executes some violent action), their behavior will most likely be perceived as out of character. For this reason, the proposed algorithm always tries to satisfy recommendations as if they were hard constraints; it disregards them only if necessary to proceed the story. Given that the system may choose to disregard some preconditions, it has more options (the space of possible stories is larger) than it would have if the preconditions were always satisfied. However, these new possibilities, which create more room for creativity, come at the expense of the possible lack of believability.

*StellA* [52] also uses the value *unknown* for certain world properties. Story characters also choose their actions assuming that unknown world properties have in fact the value required by the selected action precondition.

MEXICA [105] extends the space of possible stories because, in certain conditions, the story characters select their actions considering only incomplete world descriptions. Thus, their behavior becomes less constrained. MEXICA creative process involves a repetition of a two stages with different characteristics: engagement and reflection. During the engagement stage of the creative process, MEXICA adds events to the story in progress considering only emotion links and tensions between characters. The other properties of the story world state (e.g., the location of the characters or their physical condition) are not considered when the system selects the events to be add to the story. Considering only emotions and tensions, it is likely that many more actions can be added to the story in progress than if the remaining action pre-conditions were also considered. This approach increases the set of possible stories that may develop from any given state.

HEADSPACE [43] and IMPRACTICAL [142] allow the story characters to have incomplete and even wrong beliefs about the world.

The use of defective belief models of individual story characters also increases the number of possible generated stories because a character with a wrong belief may act in ways that they would not act if they had only correct beliefs. This happens because, in a given state of the world, there are more actions that the character may be willing to execute. Thorne and Young [43] propose a planning-based story generation system that can generate plans with actions that fail. Teutenberg and Porteous [142] propose a planning-based story generation system in which the story characters may be persuaded (by wrong beliefs) to acquire goals and to act on their behalf giving rise to stories that would not arise if the beliefs of all characters were all correct and fully described the totality of the relevant aspects of the story world.

Rather than relaxing the world description or the conditions for action, Julie Porteous and colleagues [36] present ANTON, which is a system that extends the original model of the world with the definition of new actions whose need it identifies. ANTON creates two classes of missing actions: antonymic actions, which change the story world in opposed ways to the changes operated by already existing actions; and property gaining (e.g., become-beautiful) and property loosing (e.g., become ugly) actions. Given that the number of possible stories depends on the actions the story characters may execute, this process extends the space of possible stories and hence the potential creativity of the system.

MINSTREL [23] and the *Virtual Storyteller* [67] also expand the space of possible stories. When they receive a story specification ($S$) they transform the original specification $S$ into an analogous specification $S'$ (using a set of transformation heuristics) to which they try to generate a story ($P'$). If they successfully find a story $P'$ that satisfies $S'$, they use the reverse transformation to generate a story $P$ that satisfies $S$. Considering different types of transformations, the creative process searches a larger space of possible stories.

In addition to or instead of expanding the space of possible stories, some systems try to direct their search process to more creative regions of the space. We briefly mention two such cases. During reflection, MEXICA activates constraints that push the search process into regions of potentially more interesting stories. For example, such constraints may disallow the selection of actions that do not contribute to degradation improvement processes. Planning-based story generation with *Novelty Pruning* [141] (see section 6) tries to discard regions of the search space that are not enough innovative, which is thought to improve creativity.

## 9.4 Randomness in story generation: what does it mean?

There seems to be a belief among some story generation researchers that creativity arises of processes with some degree of non-determinism. For instance, León and Gervás [52] believe that non-determinism is a *sine qua non* precondition for creativity.

Non-determinism is used as an explicit component of the proposed model in story generation systems that make decisions according to a probability distribution, such as those that select the story events using recurrent neural networks coupled to the Softmax function (e.g., [58, 164]), those using evolutionary algorithms with non-deterministic selection strategies (e.g., [33, 34]), or simulation-based systems that probabilistically choose one of a possible set of story branches according to their plausibility (e.g., [52]). However, in some other situations (e.g., [28, 39, 41, 51, 64]), randomness seems to be used because the involved researchers did not address the problem of devising mechanisms that could support the considered decisions.

It may happen that those decisions made randomly, possibly just because the focus of the papers that describe them was not to find alternative decision processes, could be supported by more narrative knowledge, by more common sense knowledge, or by better models of creativity. This section highlights several of these cases with the purpose of identifying aspects of the story generation process that could be subject to further research. After all, following Todd Pickering and Anna Jordanous [1], using random components in the story generation process might be acceptable only as a means of choosing different wordings but not as important parts of the creative process. We feel that even different wordings may be chosen for a good reason rather than just by chance.

In the reviewed literature, the impact of this kind of apparently non justified randomness in the story generation process may range from very specific and local (e.g., the choice of character actions [37]) to the global (e.g., the order by which some author goals are achieved by the story generation system [29]).

Just to clear the case, we start by presenting an example of *StellA*, a system in which non-deterministic decisions do not directly prompt further research because they are explicit integral part of the proposed model. Then, we move to several examples of possibly non justified randomness comprising such local impacts as the selection of character actions (e.g., [6, 7, 37, 51]), intermediate level choices in the search algorithm (e.g., case-based systems [23, 67], planning-based systems [28, 39] and evolutionary algorithms based systems [34]), choices at the level of story schema or story grammar traversing (e.g., *Joseph* [113] and *Propper* [45]), and the choices at the level of the story global contour (e.g., [29]).

*StellA* [52], a story generation system, uses non-deterministic simulation to create alternative sequences of events and corresponding story world states. *StellA* takes the current story world state and generates a set of alternative successor states, using all applicable rules. This way, instead of a single successor state, each state has several successor states, each of which associated with a given probability. Each branch of events from the initial state until a goal state is reached is a possible story in the story world. In the next step, *StellA* expands first the successor state with the largest accumulated probability since the initial state. Through this process, *StellA* explores a larger space of possible stories than it would if each state had only a single successor. In *StellA*, non-deterministic simulation, which allows the exploration of larger conceptual spaces, is an explicit component of the creativity model.

Contrarily to *StellA* [52], several story generation systems rely on algorithms with random decisions that were possibly not designed as first principles. Grimes' system [6, 7] generates stories using a schema-based approach. To generate a story, the system traverses the story schema proposed by Propp for the Russian folktales [111], choosing the narrative functions to include in the final story. Each narrative function is associated with a set of alternative concrete actions that the system selects randomly to instantiate the chosen sequence of narrative functions with concrete story events. According to Ryan [7], each concrete action is associated with several templates for natural language generation. The system randomly chooses the used template. In Grimes's system, the randomness appears only at the level of the selection of individual actions and their associated templates for natural language generation. The *Novel Writer* [51] and MEXICA [37, 105] use random choices in a similar way with the same kind of local impact.

In *Novel Writer* [51], each rule that controls the behavior of the characters has a set of conditions in its left hand side (LHS). Contrarily to deterministic rule-based systems, a *Novel Writer* rule may fire even if the conditions in its LHS are not all satisfied. The number of satisfied conditions is expressed as the probability of the rule firing. When a rule is evaluated a random number is generated. If the number of satisfied conditions of the rule is greater than the generated random number, the rule is used. This means that certain events that would otherwise not be included in the story may be used only by chance.

In addition to randomly deciding to include some events in the story in progress, the *Novel Writer* may also randomly choose the characters that play the two main roles in the story (i.e., murderer and victim). If the character traits of the murderer and the victim are not previously specified to the system, *Novel Writer* chooses them randomly. This random decision seems to have a more important and wide spread impact in the whole story because the traits of the main characters may condition several other decisions during the creative process.

In the engagement mode, MEXICA [37, 105] adds events to the story in progress using a set of rules associating clusters of emotion links and tensions with the set of actions that may be performed in the story. MEXICA uses the rule whose pattern of emotion links and tensions matches the emotion links and tensions between characters in the current state of the story in progress. Of all the actions suggested by the used action selection rule, MEXICA removes those that do not satisfy active constraints, and randomly selects one of the remaining.

In the next examples, randomness is more pervasive. It occurs in several decisions of the used search process. Case-based reasoning (e.g., [23]), intent driven partial order planning (e.g., IPOCL [28]) and the *vignette* based partial order planning (e.g., VB-POCL [39, 64]) make several random decisions during the search process.

MINSTREL [23] and the *Virtual Storyteller* [67] use an unusual approach to case-based reasoning. When the system receives a problem to solve, instead of retrieving the most similar case from the system's case base and adapting the solution of the retrieved case to the received problem, they transform the received problem until a part of a previous case directly matches the transformed problem. MINSTREL and the *Virtual Storyteller* may use several transformations to transform the received problem (e.g., generalization). Sometimes, it is possible to apply some transformation to more than one component of the received problem. When this happens, the system randomly chooses one of them to transform. This decision conditions the case that is used by the system to solve the received problem. This means that a random choice may impact the overall solution to the received problem.

In IPOCL, when more than one flaw is detected in a partial order plan, the planner chooses non-deterministically which of them to fix first.

When an open condition is detected in VB-POCL, the algorithm can use three strategies for fixing it: adding a temporal constraint to an action already existing in the plan, adding a new action to the plan, or retrieving and using a *vignette* (section 6). VB-POCL randomly chooses one of these strategies. Deciding to continue the story in progress by merging in it a whole new story fragment, possibly consisting of several events, instead of using a new single event or trying an event already in the story is not a mere detail. Possibly this decision ought to be better supported with more or better knowledge or using a more detailed model of creative writing.

When an action can be added to several existing intentional sequences of actions, the IPOCL planner randomly decides to which ones should it belong (an action can be part of more than one intentional action sequence if it can be used in the course of achieving more than one goal). This is an important decision given that the acceptance of the story may depend on the plausibility of the behavior of their characters. In a story, presenting a certain event as being part of the intentional behavior of a character instead of belonging to another intentional may have a significant impact on its acceptance by the reader. Possibly, rather than being randomly made, this decision should be supported by narrative knowledge or maybe by the communicative goals of the author, which in autonomous creative systems should be the system.

Although the described randomness affects diverse decisions of the planning algorithm its impact may sometimes be smaller than could be expected at first sight because often, the algorithm tries different decisions. If later, the random choices lead to an impossible to fix plan, the planner backtracks and a different decision is made. Thus, some of the used randomness affects only the order by which alternatives are considered or sub problems are tackled, at least until a solution is found.

Although it is not apparent from the reviewed sources, it is also possible that the random decisions that affect the selection of cases in case-based approaches discussed in this section may be undone if the system cannot find a case that solves the problem.

In the next example though, the randomness has a much wider effect in the search process. Baydin, López de Mántaras and Ontañón [34] propose an evolutionary approach to meme evolution. Each meme is represented as a graph in which the nodes represent concepts and the arcs represent relations between them. Before the evolutionary process starts, the system creates the initial population of meme representing graphs. The algorithm that creates the initial population makes several random decisions. Each individual of the initial population is initially set to a single node randomly picked from a common sense knowledge base. At each step, the algorithm that creates the initial population randomly picks a node from each graph representing a meme (of the initial population) and adds it a relationship and the other node of the relationship. The choice of this relationship follows common sense criteria. However, if more than one can be picked, according to the used commonsense criteria, the algorithm chooses one randomly. This process continues until some specified condition is met. As studied, for instance, in [193] and in [194], the characteristics of the initial population of evolutionary algorithms (above all, diversity) have a strong impact both on the time it takes to find a good solution and also on the quality of the generated solutions. Thus, the random choices affecting the creation of the initial population might have a large impact in the creative process.

Using explicitly given knowledge and learning from a collection of Russian tales, *Voz* extracts narrative and domain knowledge from specific stories. The retrieved knowledge is used to automatically create a case base of story examples [132] that are used by *Riu* [49] for case-based story generation. Part of the knowledge extracted by *Voz* from specific stories consists of the roles played by the story characters (e.g., villain or hero). In the system by Valls-Vargas and colleagues [55], roles are assigned to story characters using a hill climbing algorithm whose initial node is a random assignment of roles to characters. Subsequent nodes generated by the search process result of changing the character role assignment of a single character from the parent node. As in the system presented in [34], Valls-Vargas, Zhu and Ontañón [55] use a search algorithm with a random initial state. However, in the latter case, the impact may be smaller. Whereas the system described in [34] randomly creates the initial population of solutions, the approach described in [55] creates only the initial roles of the story characters (not the initial stories to be evolved).

*Propper* [45] is a schema-based story generation system that uses the narrative functions proposed by Propp for the Russian folktales [111]. *Propper* traverses a story schema retrieved from its case base and, using a set of heuristics, decides whether or not to include each optional narrative function and whether or not to stop the story generated so far. If the used heuristics are not enough for each of these decisions, *Propper* decides randomly. The generation process of schema-based systems, as *Propper*, integrates two tasks. One is selecting the sequence of narrative functions to include in the story. The other is instantiating the selected abstract narrative functions with concrete events performed by concrete characters. During the first of these tasks, the mentioned random decisions are about all that must be decided. Hence, randomness has a large impact in the generated output. Nevertheless, randomness is not used unless the system's heuristics are not enough to choose only one alternative. And even so, it is used to choose only among the alternatives selected by the heuristics.

*Joseph* [113] is a grammar-based story generation system. Using a set of story grammar rules, *Joseph* chooses the abstract structure of the story. Then, it instantiates the abstract description with concrete events and characters. Thus, the output story is strongly conditioned by the choice of story grammar rules. When more than one grammar rule can be used to generate the story, *Joseph* picks one of them randomly. This means that randomness has a strong and almost ubiquitous impact in the generated stories.

Porteous and Cavazza [29] proposed a story generation system that accepts the specification of events (i.e., author's goals) that have to occur somewhere in the story and a set of temporal relationships between them. In the face of two specified events with no temporal ordering constraints among them, the algorithm randomly chooses one of them to achieve first. This random decision has a significant impact in the general layout of the generated stories. It is possible to create stories in which some events occur early and stories in which the same events only occur later. The two versions may prompt different reactions from the reader, which means that the used randomness could be replaced by a more informed mechanism.

The reviewed approaches to story generation include random decisions that possibly were not proposed on the basis of first principles. The impact of such random decisions is sometimes not clear for the system. For instance, the impacts of the choice of the part of the problem to address (first) in a case-based system [23, 67], and the choice of the flaw to fix first in a partial order planning system [28] are not clearly predictable to the system, at the moment that those choices are made. Possibly, the system will backtrack and some of these choices will be undone. We also include in this group the choice of the initial assignment of roles to characters in a search algorithm [55]. Maybe discovering more informed mechanisms to these classes of random decisions is not the most promising research to address.

The impact of some other random choices, although not straightforward, is more predictable than the previously mentioned ones. The choices of whether to include an optional narrative function or the choice of whether to finish the story, in schema-based systems (e.g., [45]), can better be reasoned about, in spite the fact that sometimes, the same narrative function occurs more than once in a schema. However, it is possible to reason about this choices because the system has access to the schema that it is using. After all, a story with some events is not the same as a story without them. If they are important, they should be included where their effect is the best. The same goes for choosing among the possible grammar rules applicable in a given situation in systems based on story grammars such as *Joseph* [113]. We suggest these are potentially good research aspects in story generation. In this group, we also include the choice of the initial population of stories to be evolved in an evolutionary algorithm [34]. However, in a more cautionary note, it might be difficult if not impossible, to relate specific communicative goals and rhetoric resources with the initial population of a story generation system based on evolutionary algorithms.

Clearly, the justification of the character actions (e.g., the choice of the intentional action sequence in which to include a certain story event with the aim of improving character believability [28]) is a fruitful research problem.

The impact of using a specific character with their personality traits in a certain role, as described in [51], may certainly be a significant and wide spread one, although it may be almost impossible to predict all the specific details of such impact. For several researchers on automatic story generation (e.g., Aylett [172], and Stern, Frank and Resner [173]), good stories are the result of the interactions of interesting and rich characters.

Definitely, the choices that directly lead to the inclusion of new events in the story (e.g., [6, 7, 37, 39, 51]) and those that shape the general outline of the story, as it is the case in [29], should be supported by more informed mechanisms. First of all, the impact of such decisions is an important one. Secondly, the decisions of this class have an almost direct effect. Finally, the mechanisms that would replace the random choices can either be based on more narrative or common sense knowledge, on better models of creativity or both, thus their importance lies much beyond solving the mentioned decision problems.

Finally, the choice of the best template to generate natural language (e.g., [6, 7]) will have a strong impact on the generated story especially if the goal is the generation of literary natural language. However, literary natural language generation was intentionally left outside the scope of this survey.

# 10 Concluding remarks

In this section, we present conclusions about the reviewed literature, in the form of a set of suggestions regarding research areas that can have a strong impact on the advancement of the state of the art on the autonomous generation of non-interactive stories: generation, adoption and enactment of the main idea to be conveyed; adequate balance between author and character goals; using different levels of abstraction; using different communicative goals; using adequate stylistic resources; overcoming unprincipled random choices; and strategies used to ensure the creativity of the generated stories.

A story generation system can address different types of concerns, including but not limited to concerns about the main idea to be conveyed (e.g., to condemn gender prejudice), content concerns (e.g., in order to convey the main idea, is it better to use an imagined or a real world situation, a profession-centered or a family centered situation), different kinds of communicative goals such as experience communicative goals (e.g., the kind of feeling the author wants to cause on the reader), and natural language concerns, for instance stylistic resources (e.g., the stylistic resources more effective to create the experience of exasperation).

It is a widely spread practice, in story generation systems, to accept the specification of the story to be generated from the user (or from the programmer). However, human writers do not always write stories revolving about given specifications. Often, human writers write stories to communicate their own ideas. Thus autonomous creative story generation systems should also be responsible for the generation, adoption and enactment of their own ideas. In the context of this section, the ideas the autonomous creative system wants to convey may result from any kind of attitude regarding story writing. Thus, these ideas may include some moral, the presentation of an innovative perspective on some subject, or even the fruition of the possibly surprising interactions among a set of rich characters.

Possibly, the major challenge in the development of creative autonomous story generation systems is contributing to the way the system generates and adopts the main idea it wants to convey. Most if not all other major concerns are dependent on the main idea to be conveyed. For instance, decisions regarding the content of the story or about the more adequate feelings to trigger in the reader, as their reading progresses, are subordinated to what the system wants to tell.

Simply generating an idea is not enough. First, the system must evaluate the generated idea and find some value in it. Secondly, the system must commit itself to generating a story that will be used to convey the generated idea. If the system does not have enough resources (e.g., the knowledge) to generate the story for the adopted idea, it must either change the idea or acquire the necessary resources. All this is strongly related with the intentionality requirement (section 2.1) and with models of creativity, in particular, models of creative writing (sections 2.3 and 2.4).

Llano and her colleagues worked on the generation of fictional ideation (e.g., [174, 175]). This certainly contributes to the problem of the generation of the main idea. However, after a sufficiently positive evaluation of the generated idea, the system must commit to communicating it in the form of a story; an intention must be formed. The work of Llano and colleagues does not address this side of the problem and neither did any of the reviewed papers. Even the reviewed work on models of human creativity has not tackled the sub problem of the adoption of the generated idea (i.e., the formation of the intention to express the idea through a story), after it has been sufficiently positively evaluated. The engagement and reflection model of creative writing, initially presented by Sharples [17], proposes that the writing needs a primary generator, that is, an idea that guides the creative process. However, Sharples [17] does also not describe the way those primary generators are created let alone the way they become intentions.

Gervás and colleagues [195] discuss recent advances in various topics of narrative generation (e.g., plot structure, character behavior, and reader-induced suspense). According to Gervás and colleagues [195], narrative generation involves a complex feedback cycle that may include all or some of the following tasks: inventing content, organizing content, interpreting content, and validating content. The authors claim that these are exactly the tasks postulated by the ICTIVS model of story generation (Invention, Composition, Transmission, Interpretation and Validation of Stories, [76]). However, according to Gervás and León [76], the ICTIVS model does not try to solve or study how each of these process is carried out from a social or psychological point of view. Thus, although recognizing the importance of the invention of the main idea to be conveyed in the story, Gervás and co-workers do not describe the mechanism responsible for the invention of such ideas and they also do not describe a mechanism responsible for the adoption of the invented idea and the commitment to convey it in the form of a story.

The design and implementation of story generation systems capable of choosing or generating the main ideas to be conveyed in the story, depending on its own feelings, beliefs and goals, depending on its personality, depending on its past experience is still a largely unsolved problem. We suggest this as possibly one of the most interesting and fruitful topics requiring more research.

If the system has the task of conveying a certain idea to the reader through a story to be generated, another problem arises: the possible conflict between the goals entailed by that idea and the goals of the story characters. The story generation procedure makes decisions and performs actions with the purpose of achieving the goals implied by the main idea to be expressed in the story. However, the story characters have their own goals, which the system must also consider, especially because the behavior of the story characters must be believable to the reader. And it is accepted that the behavior of the story characters will be believable if the reader perceives it to be coherent (i.e., in accordance with the story world) and intentional. The difficulty for the story generation procedure consists of striving to achieve the author's goals and, at the same time, to ensure that the story characters act coherently and intentionality.

Systems in which each story character acts at the service of their own goals usually ensure the desired coherence of the generated stories because its characters behave according to the laws of the world they inhabit, their capabilities, their motives and the interactions in which they participate. However, it has often been argued that the resulting stories frequently lack interest because little or no attention is paid to what constitutes a good story, in particular the tension that is thought to be the *sine qua non* ingredient of interesting stories. Natalie Dehn [178] argues that a story generation system should satisfy the goals of the author. However, story generation systems uniquely focused on the goals of the author have more difficulty ensuring the necessary coherence and intentionality of the behavior of their characters. This has led some planning researchers to devise approaches integrating both the goals of the author and those of the story characters (e.g., [22, 28-30, 142, 149]).

However, more can be researched in different ways to achieve the right compromise between the goals of the author and the goals of the story characters. One possibility is to replicate the same fruitful lessons learned in planning-based approaches to other technological approaches to story generation (e.g., schema-based systems, case-based system, rule-based systems, evolutionary systems and learning-based systems). Another possibility is to attempt a different approach. When it is impossible to reach the adequate balance, one should contemplate the possibility of changing the story characters.

Story generation systems reflect content concerns. However, for the most part, content concerns are manifest at the level of the story micro events or at the level of the story initial conditions. Examples include, for instance *Tony's father threatened Liz that if she left Tony, he would kill Neil* [21], *James Bond defuses a bomb at Fort Knox* [29], and *the knight blocked the green creature with the shield* [52].

Given the main idea to be told, TALE-SPIN [24] is capable of setting the initial conditions (e.g., the story characters, their goals and the world) that allow it to generate a story conveying the specified moral. Once more these are relatively low-level content decisions.

Some systems (e.g., schema-based story generation systems), use two abstraction levels: the higher level narrative functions (e.g., departure, return) and the concrete domain specific contents (e.g., Bond took a plane to Kingston). However, reasoning about contents could be done ate several levels of abstraction. For instance, Proppian narrative functions may be instantiated with lower level narrative functions, which in turn may be instantiated with domain specific contents.

In the reviewed story generation systems, no reference was made to other higher level decisions regarding content. However, reasoning at different levels of abstraction, as in general problem solving, would enable creative systems to handle complex problems by incrementally tackling different concerns, in a top-down, bottom up or mix fashion. For instance, it might be easier separate higher level content choices depending on the main idea to be conveyed in the story to be generated (e.g., what kind of world situation should be chosen to convey the desired point of view to the selected target readers, what kind of character arcs are most suited) from decisions regarding the concrete sequence of story events. Moreover, some higher level decisions can be made totally independent of the concrete story domain. Contrarily to current practices (e.g., [32]), the highest levels of the story organization (e.g., Should the most relevant events be presented upfront or near the end? Should the story characters be presented near the beginning, along the story, or offered as justification for certain events or states of affairs?) do not have to be dependent on the concrete story domain.

We suggest that further research is needed to enable story generation systems to make their decisions regarding content and the corresponding stylistic resources, at different levels of abstraction. In particular, such research should aim the unveiling of relation between the diverse kinds of decisions and generation tasks and the abstraction level in which they fit better.

Some systems are concerned with triggering tension, suspense or conflict feelings on the reader (e.g., MEXICA [37, 105], *Suspenser* [148], CPOCL [30], *Glaive* [22] and PROVANT [31]), but only at relatively coarse grain. In part because the generated stories are relatively short and, in part, because the production of linguistic text was not the concern of the reviewed systems, they do not tackle the problem of deciding about the more specific and also finer grained subjective impressions that the writing should cause on the reader, at each moment (e.g., astonishment or perplexity, impression of an elated character, or unpleasantness). We suggest that more research is required about new forms of experience that story events could adequately cause on the reader and better ways to fine tune the dynamics of such experiences, which will allow stories to be more interesting to the reader.

Rendering the generated stories in natural language was not considered in this review. The reviewed literature was also not concerned with this problem, especially if we were interested in the generation of literary natural language. There are, in fact, systems more or less concerned with natural language generation. However, to the best of our knowledge, they are not focused on the generation of literary quality natural language. Even the system proposed by Hervás and Gervás [130], whose main concern is text generation, is more focused on the goal of selecting what has to be told than on the way to tell it.

The reviewed literature on story generation does not relate adequate stylistic resources with particular communicative goals. Nevertheless, there is already a relevant bulk of work on some stylistic resources that may be used in generated stories, especially those related to figurative language (e.g., [196-199]). We suggest that another important research direction is the study of the most effective stylistic resources for each possible communicative goal.

In section 9.4, we identified several subtasks of the story generation process that were randomly solved, possibly in unprincipled ways. Replacing the mentioned unprincipled randomness with principled processes may lead to a better understanding of the creative process and also to better stories. In general, we emphasize that research contributing not only to the choices made during specific situations of story generation, but above all to a better understanding of the creative process involved in writing or to increase the general narrative knowledge may be fruitful research areas. Of the possibilities analyzed in section 9.4, we stress that more research is needed to better support decisions regarding the inclusion or not of story contents (generally, story events), the perception, by the reader, of the justification for the included story events, and the choice of good characters for the required roles.

Most proposed approaches to automatic story generation include some components in their creative process that are responsible for ensuring that the generated stories meet some creativity criteria. Before continuing, we would like to stress that we have used the expression "creativity criteria" in the broadest possible sense. By "creativity criteria" we want to imply the criteria used by the system to ensure that the generated stories have desired properties such as novelty and value, regardless of what the system intends for novelty and value.

The mentioned creativity ensuring processes may be designed separately from or integrated with the generative process. For instance, the reflection stages of MEXICA's creative process, which are responsible for ensuring the coherence, the novelty and the interestingness of the generated story, occur separately from the engagement stages of the process, which are responsible for content production. But, in story generation systems based on different forms of intent based planning (e.g., [22, 28, 30]), the behavioral believability of the story characters is ensured exactly by the same process that generates the story. Although several approaches to story generation recognize the importance of some process that ensures the creativity of the generated artefact, the reviewed literature does not propose the mechanisms responsible for the design of the creativity criteria to be ensured.

*Au pair* with the autonomous generation, adoption and enactment of the main idea to be expressed in the story, the conceptual design and implementation of creativity ensuring processes is the component of the story generation process that requires more research. We finish this article with two arguments supporting our claim about the urgency and importance of that research. First, following Ontañón and Zhu [200], improving the quality of the evaluation component of the creative process is a cheap strategy to improve the quality of the generated artefact. Second, ensuring the creativity of the generated stories may be the result of entirely subjective criteria. That is, the stories being generated by a specific creative system may be judged as creative according to the autonomous interests of that system. The system my find value in the story potential for popularity, in surprise, in the positive assessment of a certain group of people or other agents, or in being provocative. Thus research in this direction will contribute to a better understanding of the development of autonomous creative systems.

**Acknowledgments**: I would like to express my gratitude to my son Miguel, who helped me with paper formatting.